\documentclass[10pt,twocolumn,letterpaper]{article}
\usepackage[pagenumbers]{cvpr}              
\usepackage[dvipsnames, table]{xcolor}

\usepackage{graphicx}
\usepackage{amsmath}
\usepackage{amssymb}
\usepackage{booktabs}
\usepackage{mathtools}
\usepackage[accsupp]{axessibility}  
\usepackage{color}
\usepackage{array}
\usepackage{pdflscape}
\usepackage[longtable]{multirow}
\usepackage{longtable}
\usepackage{tabularray}
\usepackage{booktabs}
\usepackage{times}
\usepackage{epsfig}
\usepackage{amsmath}
\usepackage{bbm}
\DeclareMathAlphabet\mathbfcal{OMS}{cmsy}{b}{n}
\usepackage{makecell}
\usepackage{siunitx}
\usepackage{svg}
\usepackage{float}
\usepackage{comment}
\usepackage{lipsum}
\usepackage{stfloats}
\usepackage{multicol}
\usepackage{multirow}
\usepackage{bm}
\usepackage{etoolbox}
\usepackage{icomma}
\usepackage{array}
\usepackage{tabulary}
\usepackage{xcolor}
\usepackage{paralist}
\usepackage{booktabs}
\usepackage{adjustbox}
\usepackage{caption}
\usepackage{subcaption}
\usepackage{arydshln}
\usepackage{rotating}

\definecolor{gray}{rgb}{0.3,0.3,0.3}
\definecolor{blue}{rgb}{0,0.5,1}
\definecolor{mask_red}{rgb}{1,0,0.8}
\definecolor{green}{rgb}{0.2,1,0.2}
\definecolor{rblue}{rgb}{0,0,1}
\definecolor{lightblue}{HTML}{6495ed}
\definecolor{lightred}{HTML}{F19C99}

\definecolor{graytablerow}{gray}{0.6}

\usepackage{pifont}
\newcommand{\cmark}{\ding{51}}%
\newcommand{\xmark}{\ding{55}}%
\usepackage{tabu}
\usepackage[pro]{fontawesome5}

\usepackage{tikz}
\newcommand*\circled[1]{\tikz[baseline=(char.base)]{
\node[shape=circle,fill=gray,inner sep=0.5pt] (char) {\textcolor{white}{\small \textbf{#1}}};}}
\definecolor{cvprblue}{rgb}{0.21,0.49,0.74}
\usepackage[pagebackref,breaklinks,colorlinks]{hyperref}
\hypersetup{colorlinks, citecolor=cvprblue}
\usepackage{tikz}
\usepackage[capitalize]{cleveref}
\crefname{section}{Sec.}{Secs.}
\Crefname{section}{Section}{Sections}
\Crefname{table}{Table}{Tables}
\crefname{table}{Tab.}{Tabs.}
\def\paperID{****} 
\def\confName{CVPR}
\def\confYear{2024}
\title{
RoDLA: Benchmarking the Robustness of Document Layout Analysis Models
}
\author{Yufan Chen$^{1}$, 
~~Jiaming Zhang$^{1,2,*}$
~~Kunyu Peng$^1$,
~~Junwei Zheng$^1$,
~~Ruiping Liu$^1$,\\
~~Philip Torr$^2$,
~~Rainer Stiefelhagen$^1$\\
\normalsize
$^1$Karlsruhe Institute of Technology,
\normalsize
~$^2$University of Oxford \\
\url{https://yufanchen96.github.io/projects/RoDLA}
}

\begin{document}
\twocolumn[{
\renewcommand\twocolumn[1][]{#1}
\maketitle
\begin{center}
    \centering
    \captionsetup{type=figure}
    \centering
    \begin{subfigure}[t]{0.3\textwidth}
        \centering
        \includegraphics[width=\textwidth, trim={2.7cm 7.5cm 2.5cm 6cm}, clip]{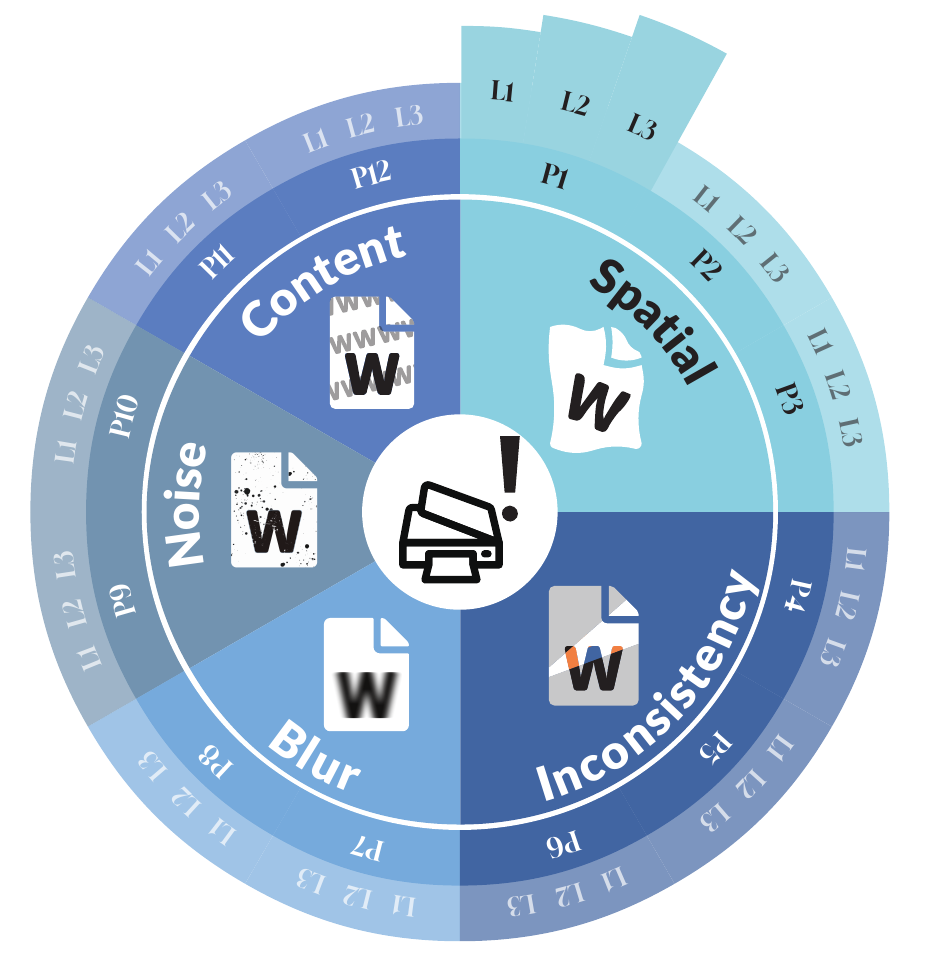}
        \caption{Taxonomy of document perturbations} \label{fig1-a}
    \end{subfigure}%
    \begin{subfigure}[t]{0.40\textwidth}
        \centering
        \includegraphics[width=\textwidth]{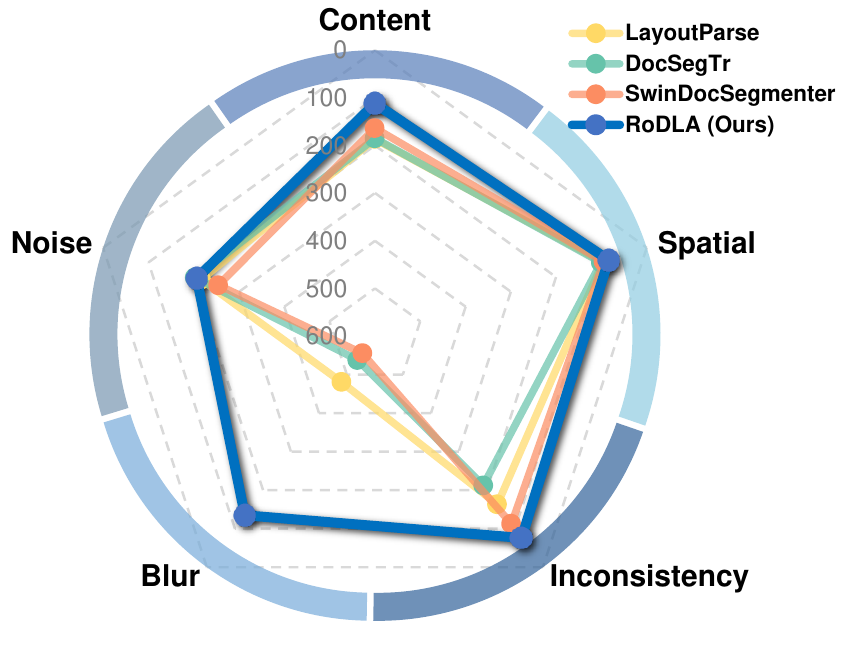}
        \caption{mRD results in five perturbation groups.
        } \label{fig1-b}
    \end{subfigure}%
     \begin{subfigure}[t]{0.3\textwidth}
        \centering
        \includegraphics[width=\textwidth,
         trim={0.1cm 0.1cm 0.1cm 0.1cm}, clip]{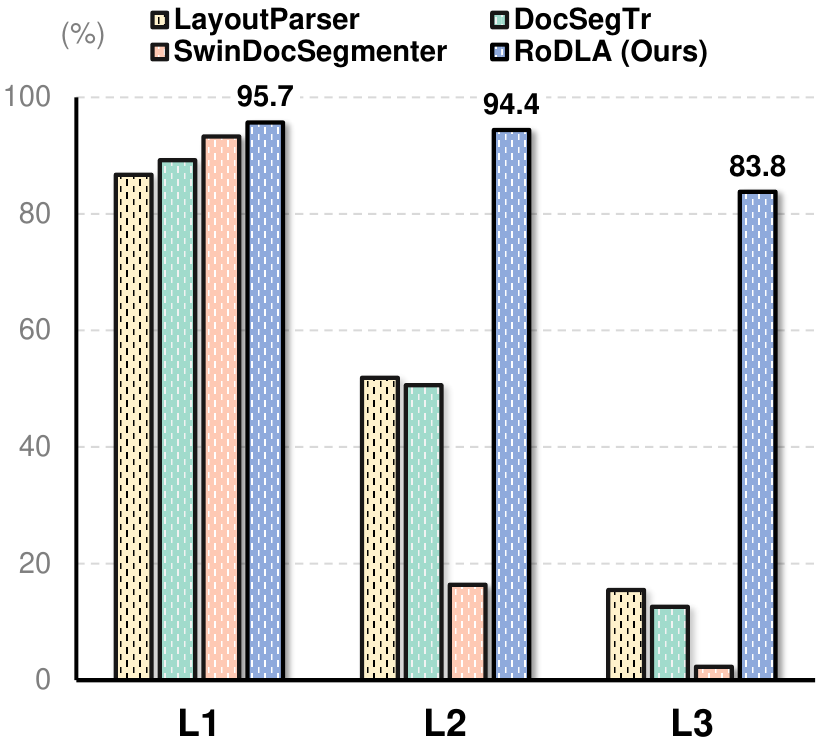}
        \caption{mAP across severity levels, \eg, \textit{defocus}. 
        } \label{fig1-c}
    \end{subfigure}%
    \setcounter{figure}{0} 
    \vskip -1ex
    \captionof{figure}{\faFileWord[regular] \textbf{Robust Document  Layout Analysis (RoDLA)} with 
    hierarchical perturbations. 
    (a) For benchmarking, we propose 5 groups (\ie, \textit{spatial transformation}, \textit{content interference}, \textit{inconsistency distortion}, \textit{blur}, and \textit{noise}) and 12 types of perturbations (P1--P12) inspired by real-world document processing, as well as 3 severity levels (L1--L3) for each perturbation. 
    Our RoDLA method obtains (b) higher mean Robust Degradation (mRD) in all 5 groups of perturbations, and (c) stable mAP scores across 3 levels of perturbation (\eg, \textit{defocus}).
    }
    \label{fig1:banner}
\end{center}
}]

{
  \renewcommand{\thefootnote}
    {\fnsymbol{footnote}}
  \footnotetext[1]{Corresponding author (e-mail: {\tt jiaming.zhang@kit.edu}).}
}
\begin{abstract}
\vskip -2ex
   Before developing a Document Layout Analysis (DLA) model in real-world applications, conducting comprehensive robustness testing is essential. However, the robustness of DLA models remains underexplored in the literature. To address this, we are the first to introduce a {robustness benchmark} for DLA models, which includes 450K document images of three datasets. To cover realistic corruptions, we propose a perturbation taxonomy with ${36}$ common document perturbations inspired by real-world document processing. Additionally, to better understand document perturbation impacts, we propose two metrics, Mean Perturbation Effect (mPE) for perturbation assessment and Mean Robustness Degradation (mRD) for robustness evaluation. Furthermore, we introduce a self-titled model, \ie, Robust Document Layout Analyzer (RoDLA), which improves attention mechanisms to boost extraction of robust features. Experiments on the proposed benchmarks (PubLayNet-P, DocLayNet-P, and M$^6$Doc-P) demonstrate that RoDLA obtains state-of-the-art mRD scores of $115.7$, $135.4$, and $150.4$, respectively. Compared to previous methods, RoDLA achieves notable improvements in mAP of $+3.8\%$, $+7.1\%$ and $+12.1\%$, respectively. 
\vskip -2ex
\end{abstract}

\section{Introduction}
\label{sec:intro}
Document Layout Analysis (DLA) is an essential component in document understanding, it indicates a fundamental comprehension of documents. As this field evolves, the shift from electronic to real-world documents presents unique challenges. These challenges are largely due to variable image quality influenced by factors like uneven \textit{illumination} and human-induced \textit{vibrations}~\cite{saifullah2022distortionstudy, jimaging2019degrated, Hegghammer2022ocrbenchmark}. These factors introduce additional complexities in DLA, as they can lead to 
distorted representations of documents, making accurate layout analysis more challenging.
According to our observation in Fig.~\ref{fig1-c}, perturbed document images raise large performance drops of previous state-of-the-art DLA models~\cite{shen2021layoutparser, biswas2022docsegtr, banerjee2023swindocsegmenter}, \eg, ${91.0\%}$ performance decrease with SwinDocSegmenter~\cite{banerjee2023swindocsegmenter} from L1 to L3, \ie, increasing perturbation severity in \textit{defocus} perturbation. It shows a notable weakness of previous models in resisting document 
perturbations. However, the robustness of DLA models remains underexplored in the literature. 
To fill the gap, we propose an extensive benchmark with almost 450,000 document images from 3 datasets for evaluating 
robustness of DLA models. The perturbation taxonomy is shown in Fig.~\ref{fig1-a}. Building on research into document image degradation~\cite{jimaging2019degrated, saifullah2022distortionstudy, Hegghammer2022ocrbenchmark, Krishnan2019handwriten, jiang2022dewarping, fronteau2023robustDocCls, feng2023doctr}, we categorize all document image perturbations into 5 high-level groups and 12 types of perturbations (P1--P12). Recognizing that perturbations not uniformly impact documents, we include 3 severity levels (L1--L3) for each perturbation. 

A suitable metric to evaluate the perturbation effects of document images and model robustness is required for the new benchmark. 
Previous metrics are constrained by the pre-selected baseline, serving as a reference for perturbation effects~\cite{hendrycks2019imagenetc, Li_2023_imagenete, yan2023RobustLidar, Kamann2020RobustnessSemantic}, which leads to metric uncertainty stemming from inherent model randomness. To address this, we design a perturbation assessment metric, \textit{Mean Perturbation Effect (mPE)}, a combination of traditional image quality assessment methods and model performance. By employing multiple methods to assess the perturbation effects, we mitigate the randomness and inconsistencies in the effect evaluation of different perturbations.
Furthermore, we propose \textit{Mean Robustness Degradation (mRD)}, a mPE-based robustness evaluation metric. The mRD results on our benchmark are shown in Fig.~\ref{fig1-b}. Our metric can minimize the impacts of model randomness and baseline selection, yielding a better measurement of model robustness.

Based on our study of robustness benchmark, we propose a novel robust DLA model, \ie, \textit{Robust Document
Layout Analyzer (RoDLA)}. It includes the channel attention to integrate the self-attention. Then, we couple it with average pooling layers to reduce the excessive focus on perturbed tokens. This crucial design enables the model to capture perturbation-insensitive features, thus significantly improving robustness. 
For instance, Fig.~\ref{fig1-c} shows the stable performance of RoDLA on the \textit{defocus} perturbation, while previous methods~\cite{shen2021layoutparser, biswas2022docsegtr, banerjee2023swindocsegmenter} have large performance drops. 
Our RoDLA model also achieves $96.0\%$ mAP on the PubLayNet dataset~\cite{zhong2019publaynet}. 
We obtain $70\%$ in mAP with a $+3.2\%$ gain and $116.0$ in mRD on the PubLayNet-P benchmark. Besides, our method reaches the state-of-the-art performance on the DocLayNet-P and M$^6$Doc-P datasets, having mAP gains of $+7.1\%$ and $+12.1\%$, respectively. 

To summarize, we present the following contributions:
\begin{compactitem}
    \item We are the first to benchmark 
    the robustness of Document Layout Analysis (DLA) models. We 
    benchmark over $10$ single- and multi-modal DLA methods by utilizing almost $450,000$ documents. 
    \item 
    We introduce a comprehensive taxonomy with common document image perturbations, which includes $5$ groups of high-level perturbations, comprising $12$ distinct types, each with $3$ levels of severity. 
    \item We design a perturbation assessment metric \textbf{Mean Perturbation Effect (mPE)} and a robustness evaluation criteria \textbf{Mean Robustness Degradation (mRD)}, which separates the impact of perturbations on images from the intrinsic perturbation robustness of the model, allowing for a more accurate robustness measurement. 
    \item 
    We propose 
    \textbf{Robust Document Layout Analyzer (RoDLA)}, which achieves state-of-the-art performance on clean datasets 
    and the 
    robustness benchmarks (PubLayNet-P, DocLayNet-P and M$^6$Doc-P).
\end{compactitem}
\section{Related Work}
\label{related_work}

\subsection{Document Layout Analysis}
Document Layout Analysis is a fundamental document understanding task, which extracts 
the structure and content layout of documents. 
Thanks to the diverse datasets and benchmarks~\cite{li2020docbank,shihab2023badlad,zhong2019publaynet,shen2020large,moured2023line}, machine learning-based approaches~\cite{diem2011text,garz2010detecting} and deep learning-based approaches~\cite{long2022towards,gemelli2022doc2graph,peng2022ernie,zhu2022towards,coquenet2023dan,yang2022transformer} have made progress. 
Both single-modal methods, \eg, Faster R-CNN~\cite{ren2017FasterRcnn}, Mask R-CNN~\cite{He_2017_MaskRcnn}, and DocSegTr~\cite{biswas2022docsegtr},
and multi-modal methods, \eg, DiT~\cite{li2022dit} and LayoutLMv3~\cite{huang2022layoutlmv3}, are well explored in DLA. Besides, text grid-based methods~\cite{zhang2021vsr, yang2017learning} deliver the combination capability of the text grid with the visual features. 
Recently, transformer models~\cite{coquenet2023dan, yang2022transformer, cheng2023m6doc, tang2023unifying, li2022dit, arroyo2021variational, wang2022lilt} have been explored in DLA. 
Self-supervised pretraining strategies~\cite{xu2020layoutlm, xu-etal-2021-layoutlmv2, li2021structext, Appalaraju_2021_DocFormer,Luo2022BiVLDocBV,huang2022layoutlmv3,luo2023geolayoutlm} have also drawn considerable attention in DLA, \eg, DocFormer~\cite{Appalaraju_2021_DocFormer} and LayoutLMv3~\cite{huang2022layoutlmv3}. However, existing works in DLA focuses on clean document data, overlooking real-world issues like noise and disturbances. In this work, we aim to fill this gap by benchmarking the robustness of DLA models.

\subsection{Robustness of Document Understanding}
As a related task, document restoration and rectification is the task of improving the image quality of documents by correcting distortions.
DocTr++~\cite{feng2023doctr} 
explores unrestricted document image rectification.
In \cite{fronteau2023robustDocCls}, robustness against adversarial attack is investigated for document image classification.
Auer~\etal~\cite{Auer2023ICDAR2C} propose a challenge for robust document layout segmentation. To address this, Zhang~\etal~\cite{Zhang2023WeLayoutWL} propose a wechat layout analysis system. 
The robustness evaluation on the RVL-CDIP dataset~\cite{harley2015RVL_CDIP} is for document classification.
Tran~\etal~\cite{Tran2017ARS} propose a robust DLA system by using multilevel homogeneity structure. However, these works are oriented towards optimizing performance on clean documents. 
Our research is the first to systematically study real-world challenges of DLA. We benchmark DLA methods with extensive perturbation types, encompassing $3$ datasets, $5$ perturbation groups, $12$ distinct types, and $3$ severity levels for each type. 

\subsection{Robust Visual Architectures}
A robust visual architecture is required to maintain 
reliable visual analysis. Some researches are established in object detection~\cite{milyaev2017towards,shen2020noise,piao2022noise,michaelis2019benchmarking,hahner2022lidar,du2022unknown,wang2022ryolo} and image classification~\cite{modas2022prime,papakipos2022augly,8889765,dong2020benchmarking}.
Modas~\etal~\cite{modas2022prime} introduce few primitives which can boost the robustness in image classification field.
R-YOLO~\cite{wang2022ryolo} propose a robust object detector under adverse weathers.
FAN~\cite{zhou2022fan} proposes fully attention networks to strengthen the robust representations.
A Token-aware Average Pooling (TAP)~\cite{guo2023tap} module is proposed to encourage the local neighborhood of tokens to participate in the self-attention mechanism.
However, directly applying existing robust methods to domain-specific tasks like DLA cannot yield optimal performance due to the unique challenges. To address this, a synergistic attention-integrated model is designed to enhance the robustness of DLA by concentrating attention on key tokens on multi-scale features and enhancing attention interrelations.  
\section{Perturbation Taxonomy}
\label{perturbations_taxonomy}
\subsection{Hierarchical Perturbations}
In the field of Document Layout Analysis (DLA), it is essential to grasp the structure and content arrangement 
This is particularly challenging with document images from scans or photos, where processing errors and 
disturbances can degrade DLA performance. Previous benchmarks~\cite{zhong2019publaynet, pfitzmann2022doclaynet, li2020docbank} have not fully addressed these challenges, as they simply include a collection of real documents images without analysis of perturbations. 
Therefore, we introduce a robustness benchmark with 450,000 document images, incorporating $12$ 
perturbations with 3 level of severity to systemically evaluate the robustness of DLA models.

Given a DLA model $g {:} X {\rightarrow} L$, trained on a digital documents dataset $\mathcal{N}$, the model in traditional benchmark is tested by the probability $P_{(x,l)\sim \mathcal{N}}(g(x) = l)$. However, real-world documents images often experience various disturbances 
~\cite{saifullah2022distortionstudy,jiang2022dewarping, Hegghammer2022ocrbenchmark}, which are not represented in a digital dataset. We consider a set of document-specific perturbation functions $O$, and approximate the perturbation distribution in real-world with $P_O(o)$. To bridge gap from digital to real, our approach evaluates the DLA model's effectiveness against common perturbations in documents by the expectation $E_{o \sim O}[P_{(x,l)\sim \mathcal{N}}(g(o(x)){=}o(l))]$. Through expectation, we can effectively measure the average model performance impact under same perturbation type, rather than merely assessing performance under worst-case perturbation.

\renewcommand\thesubfigure{P\arabic{subfigure}}
\begin{figure}[t]
 \footnotesize	
 \centering
 \subfloat[Rotation]{\frame{\includegraphics[width=0.3\columnwidth]{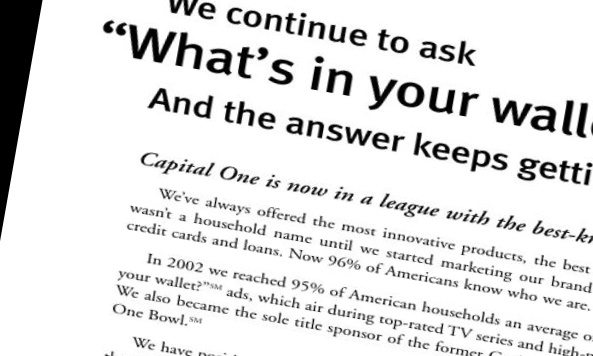}}}\hspace{5pt}
 \subfloat[Warping]{\frame{\includegraphics[width=0.3\columnwidth]{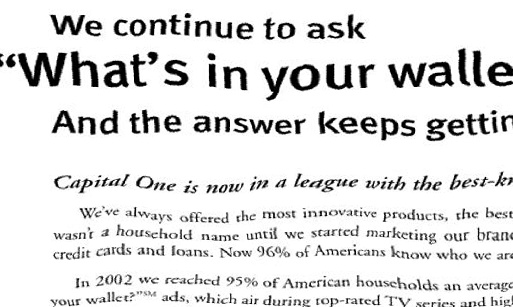}}}\hspace{5pt}
 \subfloat[Keystoning]{\frame{\includegraphics[width=0.3\columnwidth]{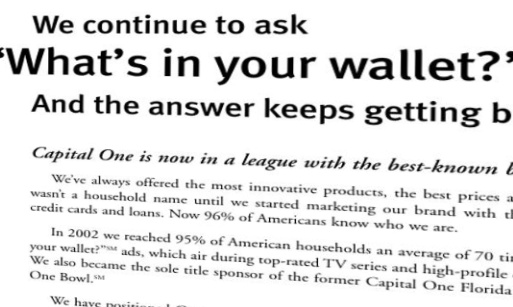}}}\hspace{5pt}
 \subfloat[Watermark]{\frame{\includegraphics[width=0.3\columnwidth]{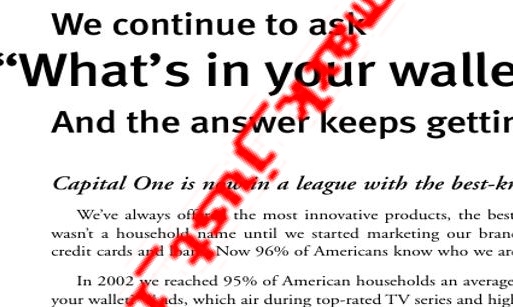}}}\hspace{5pt}
 \subfloat[Background]{\frame{\includegraphics[width=0.3\columnwidth]{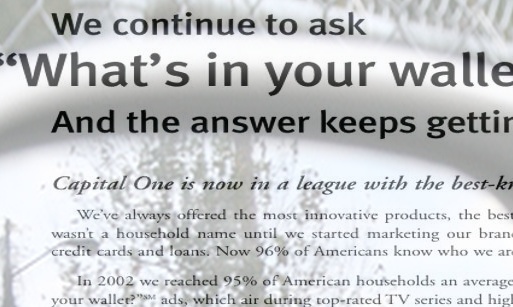}}}\hspace{5pt}
 \subfloat[Illumination]{\frame{\includegraphics[width=0.3\columnwidth]{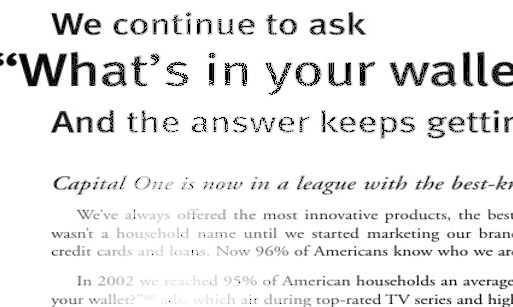}}}\hspace{5pt}
 \subfloat[Ink-bleeding]{\frame{\includegraphics[width=0.3\columnwidth]{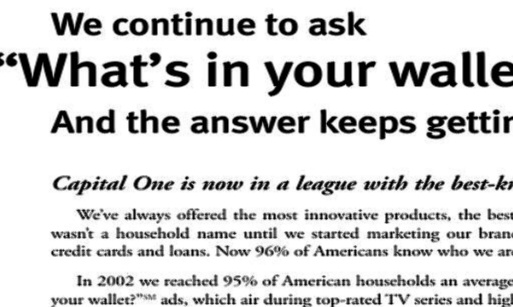}}}\hspace{5pt}
 \subfloat[Ink-holdout]{\frame{\includegraphics[width=0.3\columnwidth]{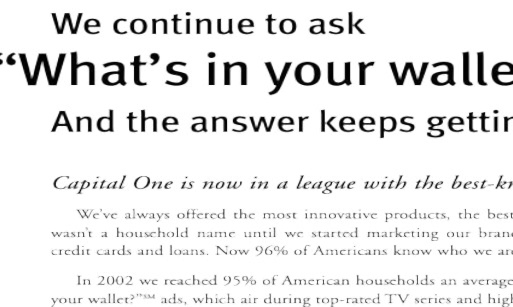}}}\hspace{5pt}
 \subfloat[Defocus]{\frame{\includegraphics[width=0.3\columnwidth]{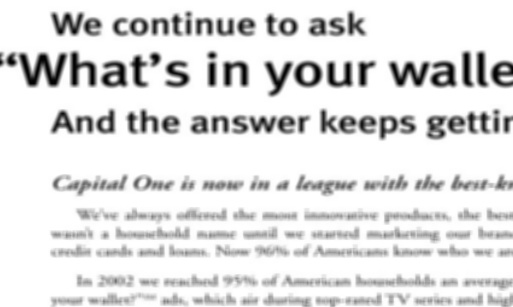}}}\hspace{5pt}
 \subfloat[Vibration]{\frame{\includegraphics[width=0.3\columnwidth]{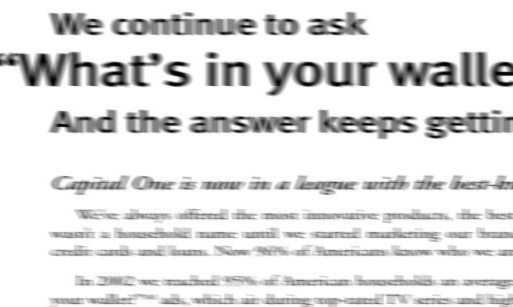}}}\hspace{5pt}
 \subfloat[Speckle]{\frame{\includegraphics[width=0.3\columnwidth]{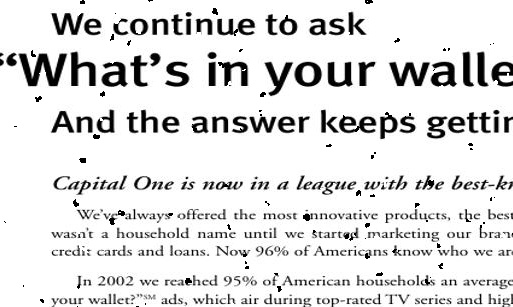}}}\hspace{5pt}
 \subfloat[Texture]{\frame{\includegraphics[width=0.3\columnwidth]{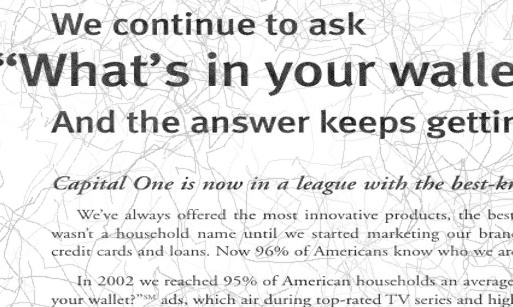}}}
 \vskip -1ex 
 \caption{\textbf{Visualization of document perturbations.}} \vskip -1em 
\label{fig:perturbation_effect}
\end{figure}
\renewcommand\thesubfigure{\alph{subfigure}}
As shown in Fig.~\ref{fig1-a}, we divide the document perturbations into 5 groups, \ie, \textit{spatial transformation}, \textit{content interference}, \textit{inconsistency distortion}, \textit{noise} and \textit{blur}. In these 5 groups, there are totally 12 types of document perturbation with 36 severity levels. The visualization of 12 perturbations is shown in Fig.~\ref{fig:perturbation_effect}. These document perturbation settings are derived from common perturbation observed in scanned documents and photographic document images. They represent the typical challenges and variations encountered in real-world scenarios, ensuring that our approach is both comprehensive and practical. The detailed settings of each perturbation are described as the following. 
\begin{table*}
\centering
\small
\caption{\textbf{\faFileWord[regular] Document perturbations} on PubLayNet-P, DocLayNet-P, and M$^6$Doc-P datasets. mPE: mean Perturbation Effect. }\vskip -1em
\label{tab:setting_p}
\setlength{\tabcolsep}{5pt}
\renewcommand{\arraystretch}{1.1}
\resizebox{\linewidth}{!}{
\begin{tabular}{l|l|c|c}
\toprule
\multicolumn{2}{c|}{\textbf{\faFileWord[regular] Document Perturbation (P)}} & \textbf{Description} & {\textbf{mPE{$\downarrow$}}} \\ 
\hline 
\multicolumn{2}{c|}{None} & The original clean data  & 0 \\ 
\hline
\multirow{3}{*}{{\circled{1} Spatial}}
&\textbf{(P1) Rotation} & Rotation simulation with $ \theta {=} [5^{\circ}, 10^{\circ}, 15^{\circ}]$   & 58.30 \\
&\textbf{(P2) Warping} & Elastic transformation simulation with $ R_{\sigma} {=} [0.2, 0.06, 0.04]$ and $ R_{\alpha} {=} [2, 0.6, 0.4]$ & 22.00 \\
&\textbf{(P3) Keystoning} & Perspective transformation simulation with $ R_k {=} [0.02, 0.06, 0.1]$ & 34.49\\
\hline
\multirow{2}{*}{{\circled{2} Content}}&\textbf{(P4) Watermark} & Text overlay with $\alpha_w {=} [51, 153, 255]$ and $ R_z {=} [2, 4, 6] $ & 09.05 \\
&\textbf{(P5) Background 
} & Random image inserting in background with $ N_i {=} [1, 3, 5] $ & 26.70 \\
\hline
\multirow{3}{*}{{\circled{3} Inconsistency}}&\textbf{(P6) Illumination 
} & Region illumination change with $V_l {=} [51, 102, 153]$ or $ V_s {=} [0.5, 0.25, 0.17]$ & 10.67 \\
&\textbf{(P7) Ink-bleeding 
} & Erosion simulation with kernel size $ K_e {=} [3, 7, 11] $ & 08.91 \\
&\textbf{(P8) Ink-holdout 
} & Dilation simulation with kernel size $ K_d {=} [3, 7, 11] $ & 15.64 \\
\hline
\multirow{2}{*}{{\circled{4} Blur}}&\textbf{(P9) Defocus 
} & Gaussian blur simulation with kernel size $ K_g {=} [1, 3, 5] $& 08.10 \\
&\textbf{(P10) Vibration 
} & Motion blur simulation with kernel size $ K_m {=} [3, 9, 15] $& 16.29 \\
\hline
\multirow{2}{*}{{\circled{5} Noise}}&\textbf{(P11) Speckle} & Blotches noise simulation with $D_b {=} [\num{1e-4},\num{3e-4},\num{5e-4} ]$ & 24.31 \\
&\textbf{(P12) Texture 
} & Fibrous noise simulation with $ N_f {=} [300, 900, 1500]$ & 40.57 \\
\hline
\multicolumn{2}{c|}{Overall} & The average of all perturbations  & 22.90 \\
\bottomrule
\end{tabular}}
\vskip-3ex
\end{table*}

\subsection{Perturbation Description}
In Table~\ref{tab:setting_p}, $5$ groups and $12$ types of perturbations 
are mathematically defined and divided into $3$ levels of severity. 

\noindent \textbf{\circled{1} Spatial Transformation.} This group involves three perturbations: (P1)~\textbf{Rotation} around the center of document without preserving the ratio. 
The severity is determined by the angle $\boldsymbol{\theta}$, which is randomly chosen from a uniform distribution within predefined ranges. 
(P2) \textbf{Warping}, which is enacted by generating two deformation fields, $(\Delta x, \Delta y){=} \alpha \cdot \mathcal{G}( U_{(x,y)}, \sigma)$.
$U_{(x,y)} $ obeys the uniform distribution, $\mathcal{G}$ denotes Gaussian filter,  $\sigma$ and $\alpha$ are the smoothness and magnitude controlled by scaling factor $\boldsymbol{R_{\sigma}}$ and  $\boldsymbol{R_{\alpha}}$, respectively. 
(P3) \textbf{Keystoning}, simulated by perspective transformation using homograph matrix $H$, maps the original coordinates to new ones. 
The corner points are adjusted by offsets drawn from a normal distribution, with standard deviations scaled by factor $\boldsymbol{R_k}$. 
Annotation transformations are adjusted accordingly to alignment.

\noindent \textbf{\circled{2}  Content Interference.} Document content interference stems from two primary sources: (P4) Text \textbf{Watermark} is commonly used as an anti-piracy measure. To simulate the text watermark, we overlay uniformly sized characters at random document positions with random rotations. The interference strength is controlled by adjusting the zoom ratio $\boldsymbol{R_z}$ and alpha channel opacity $\boldsymbol{\alpha_w}$. 
(P5) Complex \textbf{Background}, which is often leveraged in printed media to enrich the document's visual content. Images from ILSVRC dataset~\cite{Russakovsky2015ILSVRC}, are randomly embedded within the document images for the simulation of background. The severity level is determined by the number of embedded images $\boldsymbol{N_i}$. 

\noindent \textbf{\circled{3}  Inconsistency Distortion}, which addresses distortions, \eg, (P6) Non-uniform \textbf{Illumination}, simulated through a $50$\% probability of glare or shadow, with levels determined by the brightness $\boldsymbol{V_l}$ or shadow $\boldsymbol{V_s}$, (P7) \textbf{Ink-bleeding}, achieved by using erosion process within the document image, and (P8) \textbf{Ink-holdout}, which are accomplished by using morphological dilation operations with elliptical kernel $\boldsymbol{K_e}$ and $\boldsymbol{K_d}$, respectively. 
\noindent \textbf{\circled{4}  Blur.} Blur effects arise from (P9) \textbf{Defocus}, which is approximated with Gaussian kernel function, \ie, the point spread function (PSF) with severity levels controlled by kernel size $\boldsymbol{K_g}$, and (P10) \textbf{Vibration}, which is achieved by using a convolution kernel rotated by a random angle to replicate directional motion in size $\boldsymbol{K_m}$. This perturbation can be applied via a Gaussian filtering operation.

\noindent \textbf{\circled{5} Noise.} 
Two unique noise types in document images are included. (P11) \textbf{Speckle} caused by ink clumping, can be achieved by using Gaussian noise modulated by blob density $\boldsymbol{D_b}$. It reproduces both foreground and background noise, reflecting the complex stochastic nature of noise artifacts. 
(P12) \textbf{Texture}, resembling the fibers in paper, is added to simulate the natural fiber structures of archival documents. In each fiber segment, curvature and length are independently simulated with trigonometric functions and Cauchy distribution, forming the complete fiber through segment assembly. The number of fibers $\boldsymbol{N_f}$ varies across noise levels to represent different paper qualities.

The principles of design and division of these perturbations are {three-fold}: (1) All perturbations are realistic and inspired by the real-world disturbance or document layout analysis. 
(2) The leveraged perturbations are comprehensive and occur in document from top to bottom, from global to local, from dense to sparse, and from content-wise to pixel-wise. (3) All perturbation levels are reasonably determined by Image Quality Assessment (IQA) metrics, \eg, MS-SSIM~\cite{wang2003msssim} and CW-SSIM~\cite{Sampat2009cwssim}, Degradation and our proposed metrics which are detailed as the following. 

\subsection{Perturbation Evaluation Metrics}
Quantifying the document perturbations is a crucial task, which lacks straightforward metrics. 
Previous methods~\cite{Kamann2020RobustnessSemantic, yan2023RobustLidar, hendrycks2019imagenetc, Li_2023_imagenete} are constrained by using a pre-selected baseline model performance as a reference to measure the perturbation effect, which will conflate the perturbation effect with the model robustness due to the inherent variability. 

To design an effective perturbation metric, our considerations are two-fold:
(1) Evaluation should be relatively independent to any specific model performance. (2) Evaluation should be inclusive and sensitive to all perturbations.
Inspired by the IQA methods~\cite{ding2021IQA}, we analyse all $12$ perturbations 
through $4$ metrics, as presented in Fig.~\ref{fig:IQA_visual}.

\begin{figure*}[htbp]
    \centering
    \captionsetup{type=figure}
    \centering
    \begin{subfigure}[t]{0.46\textwidth}
        \centering
        \includegraphics[width=\textwidth]{figure/IQAv2_half.pdf}
        \caption{Comparison of four evaluation metrics on 12 perturbations} \label{IQA_visual-a}
    \end{subfigure}\hfill
    \begin{subfigure}[t]{0.52\textwidth}
        \centering
        \includegraphics[width=\textwidth]{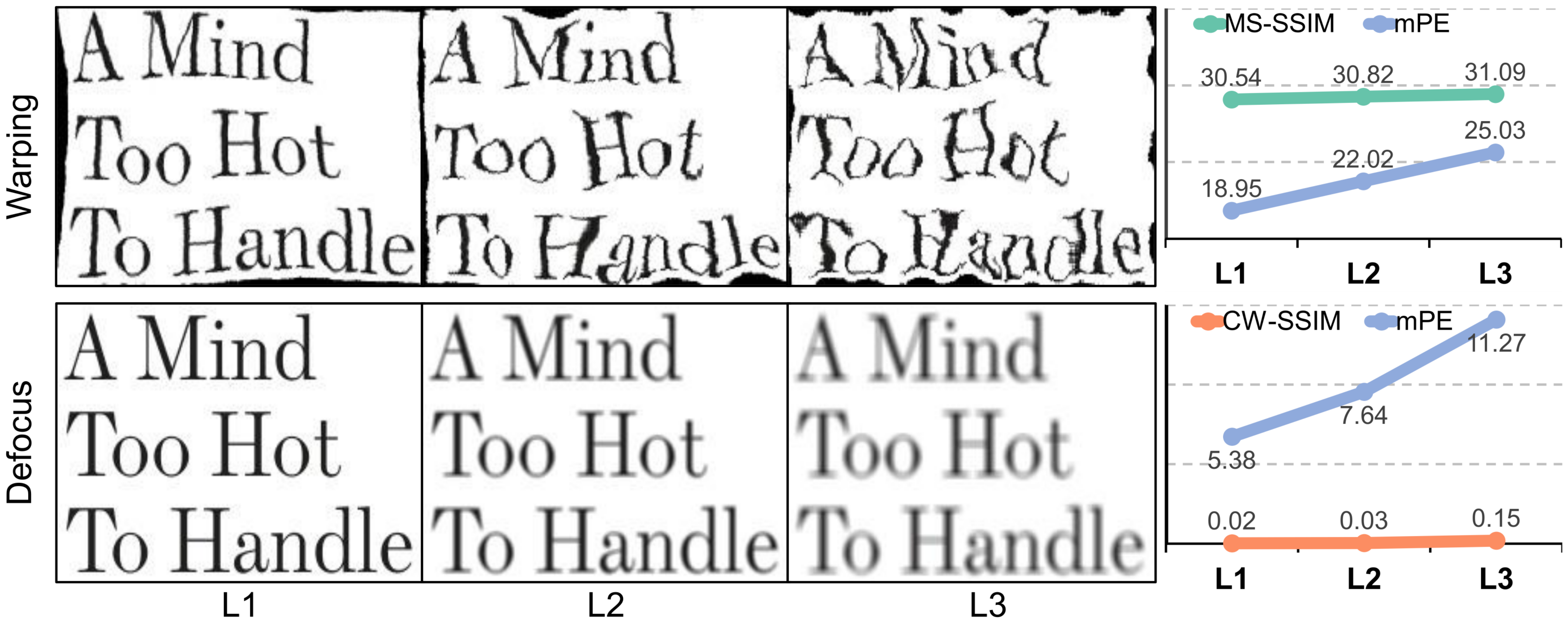}
        \caption{Comparison of the sensitivity of metrics to different perturbation levels.
        } \label{IQA_visual-b}
    \end{subfigure}
    \setcounter{figure}{2}
    \captionof{figure}{\textbf{Analysis of perturbation evaluation metrics.} (a) Comparison of perturbation metrics, including MS-SSIM, CW-SSIM, {Degradation} \textit{w.r.t} a baseline, and the proposed {mean Perturbation Effect} (mPE). mPE is more balanced and inclusive to different perturbations. (b) Six documents perturbed by \textit{warping} and \textit{defocus} and their scores indicate that mPE is more sensitive to measure different levels.} 
    \label{fig:IQA_visual}
    \vskip -2ex
\end{figure*}

\noindent \textbf{MS-SSIM} (Multi-Scale Structural Similarity Index)~\cite{wang2003msssim} evaluates image quality by considering difference in structural information 
across different resolutions, as: 
\begin{equation}\label{eq:ms-ssim}
    f^{\text{MS-SSIM}}(x,y) {=} [l_M(x,y)]^{\alpha_M} \cdot \prod_{j=1}^{M} [c_j(x,y)]^{\beta_j} [s_j(x,y)]^{\gamma_j},
\end{equation}
where $l_M(x,y)$ represents the luminance difference at the $M$-th scale, and $c_j(x,y)$, $s_j(x,y)$ are the contrast and structure comparison at the $j$-th scale. It measures the impact of most perturbations effectively but lacks sensitivity to \textit{watermark} and \textit{warping} (Fig.~\ref{IQA_visual-b}), and is overly sensitive to \textit{rotation} and \textit{keystoning}, as shown in Fig.~\ref{IQA_visual-a}.

\noindent \textbf{CW-SSIM} (Complex Wavelet Structural Similarity Index)~\cite{Sampat2009cwssim} assesses the similarity of local patterns of pixel intensities transformed into the wavelet domain, which is robust to spatial distortions, \eg, \textit{keystoning} and \textit{rotation}, formulated by the following equation:
\begin{equation}\label{eq:cw-ssim}
    f^{\text{CW-SSIM}}(x,y) = \frac{2 | \sum_{l=1}^{L} w_l x_l y_l^* | + K}{\sum_{l=1}^{L} |x_l|^2 + \sum_{l=1}^{L} |y_l|^2 + K},
\end{equation}
where $x_l$ and $y_l$ denote the wavelet coefficients, $w_l$ is the $l$-th coefficient weight, $L$ indicates the local region number, $K$ is tiny value, and $*$ denotes the complex conjugation. As shown in Fig.~\ref{IQA_visual-a}, it reflects the perturbations impact but fails for quantitive evaluation at different levels for some perturbations, \eg, \textit{defocus}, demonstrated in Fig.~\ref{IQA_visual-b}.

\noindent \textbf{Degradation} across severity levels, as indicated by ImageNet-C~\cite{hendrycks2019imagenetc}, is a common practice for various robustness benchmark~\cite{Kamann2020RobustnessSemantic, yan2023RobustLidar, hendrycks2019imagenetc, Li_2023_imagenete}.  
Degradation $D$, where $D {=} 1 - mAP$ similar with ImageNet-C, illustrates the performance impact of perturbation on a baseline model. From Fig.~\ref{IQA_visual-a}, we notice that the Degradation metric is highly dependent on the choice of the baseline and tends to be overly sensitive to perturbations, \eg, \textit{background} and \textit{texture}. 

\noindent \textbf{Mean Perturbation Effect (mPE)} is a new metric proposed to assess the compound effects of document perturbations. 
For specific perturbation $p$ at each severity level $s$, degradation of model $g$ is written as $D_{s,p}^g$, and IQA metrics is written as $f_{s,p}^i$, \ie, from Eq.~(\ref{eq:ms-ssim}) and Eq.~(\ref{eq:cw-ssim}). 
The \textbf{mPE} to measure the effect of perturbation $p$ is calculated as: 
\begin{equation}\label{eq:mpe}
    \textbf{mPE}_p = \frac{1}{NMK} \sum_{s=1}^{N} {(\sum_{i=1}^{M}{f_{s,p}^i} + \sum_{g=1}^{K}{D_{s,p}^g})}.
\end{equation}
mPE aims to isolate the model's robustness from the image alterations impact. We employ MS-SSIM and CW-SSIM as IQA metrics, and we select Faster R-CNN~\cite{ren2017FasterRcnn} as a baseline in Degradation metric. As evident from Fig.~\ref{IQA_visual-b}, our mPE metric effectively quantifies the impact of each perturbation and severity level, while other metrics are less sensitive.

\subsection{Perturbation Robustness Benchmarks}
To systematically benchmark the robustness of DLA models, we apply the mentioned $12$ perturbations to $3$ datasets. 

\noindent \textbf{PubLayNet-P}~\cite{zhong2019publaynet} is a commonly used large-scale DLA dataset. It contains over $360,000$ document images sourced from PubMed Central, and focuses on $5$ principal types of layout elements, \ie, \textit{text}, \textit{title}, \textit{list}, \textit{table}, and \textit{figure}.

\noindent \textbf{DocLayNet-P}~\cite{pfitzmann2022doclaynet} 
includes academic papers, brochures, business letters, technical papers, and more with $80,863$ document pages. It contains annotations for $11$ layout components, aiming at enabling generalizable DLA models. 

\noindent \textbf{M$^6$Doc-P}~\cite{cheng2023m6doc} 
provides more nuanced annotations ($74$ classes), capturing finer aspects of document layouts with $9,080$ document images. This dataset is tailored to advance research in sophisticated document understanding tasks.

While PubLayNet provides vast array of annotated scientific documents, DocLayNet broadens the variety of document types, and M6Doc delves into more detailed and complex annotations, catering to advanced document analysis.
\section{Robust Document Layout Analyzer}
\label{methodology}
\subsection{Framework Overview}
\begin{figure}[htbp]
  \centering
  \includegraphics[width=\linewidth, trim={4cm 2cm 5cm 0.2cm}, clip]{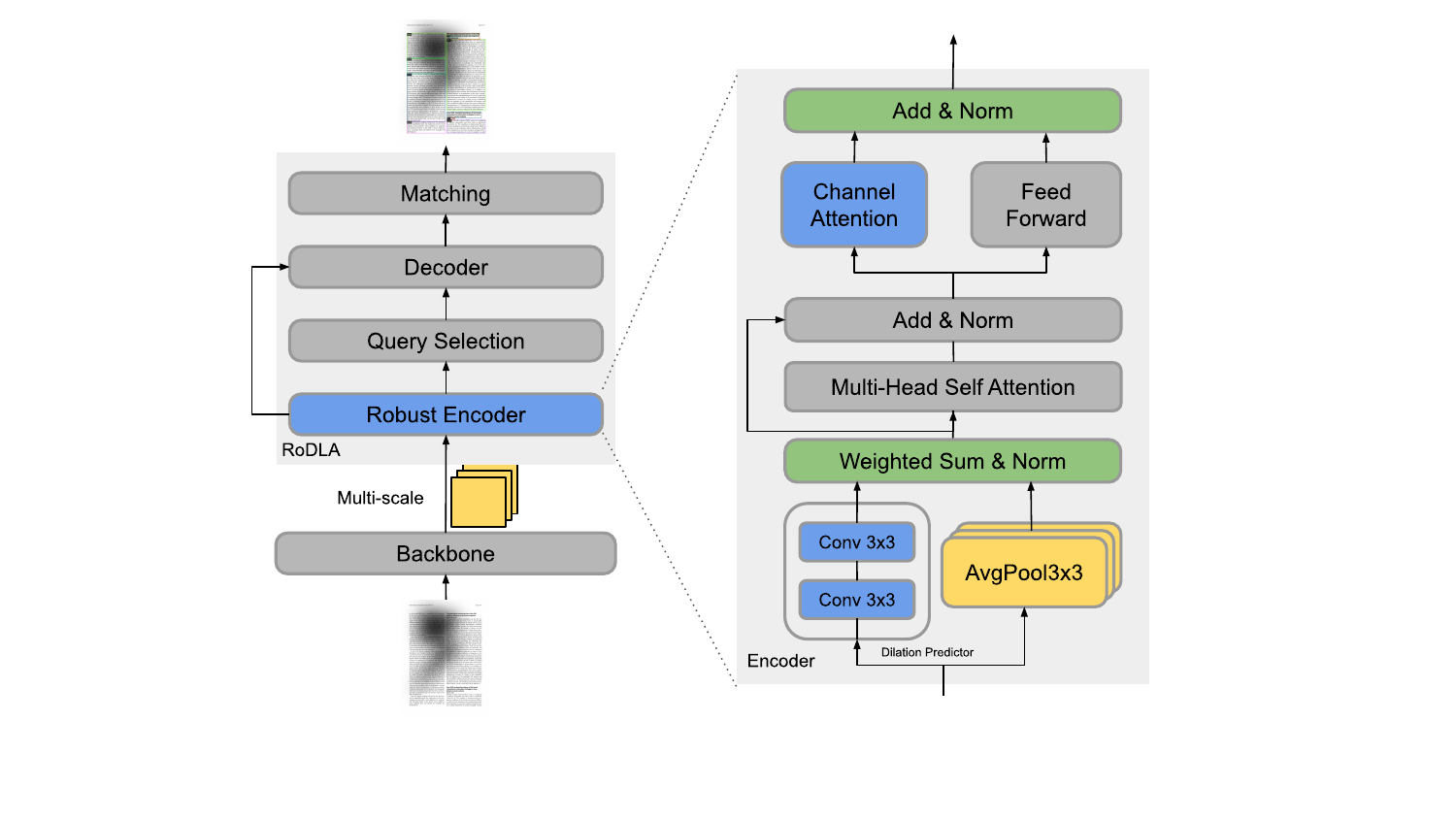}
  \vskip -0.5em\caption{\textbf{The architecture of RoDLA model.} RoDLA is comprised of Encoder, Query Selection, Decoder, and Matching components. It optimizes the attention mechanism in Encoder, heightening focus on crucial tokens and reinforcing key token connections in multi-scale features to extract stable features.}
  \vskip-2ex
  \label{fig:RoDLA}
\end{figure}

To boost the robustness performance on perturbed documents, we propose a \textit{Robust Document Layout Analyzer (RoDLA)} model. 
As shown in Fig.~\ref{fig:RoDLA}, our proposed model is inspired by the architecture of DINO~\cite{zhang2022dino} but introduces channel attention blocks and average pooling blocks in the Encoder, allowing for more robust feature extraction. 
Additionally, we incorporate InternImage\cite{wang2023internimage} as a backbone, which is pre-trained on ImageNet22K~\cite{deng2009imagenet} dataset. 
This pretraining setting enhances multi-scale feature extraction for more stable performance. The overall architecture and robustness enhancement design is depicted in Fig.~\ref{fig:RoDLA}.

\subsection{Robustness Enhancement Design}
As identified in~\cite{guo2023tap}, self-attention tends to overemphasize focus on irrelevant tokens, that leads to attention shifting in visual document analysis, where globally accumulated perturbations may interfere with the model performance. To address this, we integrate spatial-wise average pooling layers with varying dilation into the \textit{Robust Encoder}, accompanied by a \textit{Dilation Predictor} employing two $3{\times}3$ convolutional layers to predict the utilization of dilated average pooling layers. This approach leverages the attention of neighboring tokens to mitigate the overemphasis on tokens, and dynamically constrains long-distance perception based on context. 
By diminishing attention on irrelevant tokens, 
this approach effectively captures robust feature. 

As outlined in~\cite{zhou2022fan}, self-attention inherently facilitates visual grouping, filtering out irrelevant information. 
Thus, we have incorporated a \textit{Channel-wise Attention (CA)} module within the \textit{Encoder}. This module implements a self-attention operation in channel dimension $d$, as:
{\scriptsize$\textbf{Y}{=}\sigma(\frac{\texttt{Softmax}(\textbf{Q})\cdot \texttt{Softmax}(\textbf{K}^T)}{\sqrt{n}}{\cdot} \texttt{MLP}(\textbf{V}))$}, where $\textbf{Q},\textbf{K},\textbf{V} {\in} \mathbb{R}^{d \times n}$. This mechanism not only simplifies computational complexity but also aggregates local features in channel-wise. The encoder module synergies channel-wise and spatial-wise self-attention, aggregating strongly correlated feature channels while directing attention towards spatially relevant key tokens. This enhancement of global attention 
minimizes the shifting of attention tokens, leading to robust feature extraction for complex analysis task.
\vskip -4em
\section{Benchmarking and Analysis}
\label{experiments}

\subsection{Experiment Settings}
\noindent \textbf{Selected Baselines.} 
To establish a comprehensive robustness benchmark for DLA, 
we not only include 
three distinct datasets but also reproduce 
a variety of state-of-the-art 
methods and backbones, 
including CNN-based backbones (\textbf{ResNet}~\cite{kaiming2015resnet}, \textbf{LayoutParser}~\cite{shen2021layoutparser}, \textbf{InternImage}~\cite{wang2023internimage}), Transformer-based structures (\textbf{SwinDocSegmenter}~\cite{banerjee2023swindocsegmenter}, \textbf{DiT}~\cite{li2022dit}, and \textbf{LayoutLMv3}~\cite{huang2022layoutlmv3}), and methods from CNN-based structures (\textbf{Faster R-CNN}~\cite{ren2017FasterRcnn}, \textbf{Mask R-CNN}~\cite{He_2017_MaskRcnn}, \textbf{Cascade R-CNN}~\cite{cai2018cascade}), to Transformer-based frameworks (\textbf{DocSegTr}~\cite{biswas2022docsegtr}, \textbf{DINO}~\cite{zhang2022dino}, \textbf{Co-DINO}~\cite{zong2022codetr}, and our \textbf{RoDLA}). Note that all the methods are pre-trained on ImageNet dataset, while DiT and LayoutLMv3 are pre-trained on the IIT-CDIP Test Collection 1.0 document dataset~\cite{Lewis2006IITCDIP}, which gives a comparison of the pre-training data. 

\noindent \textbf{Implementation Details.} For fair comparison, we reproduce all methods in the MMDetection~\cite{mmdetection} framework. Hyperparameters are from their original settings or papers. All models are trained and validated exclusively on clean data. Only during the testing phase, the impact of perturbations is evaluated on all methods. This approach ensures that our robustness benchmark is conducted under controlled and unbiased conditions. More details are in supplementary. 

\subsection{Robustness Evaluation Metrics}
From our analysis in Fig.~\ref{fig:perturbation_effect} and Table~\ref{tab:setting_p}, it is evident that various types of perturbations impact the same document images differently, leading to a range of effects on the models' performance. Previous robustness benchmarks~\cite{hendrycks2019imagenetc, Li_2023_imagenete,michaelis2019benchmarking,yan2023RobustLidar} measure the perturbations impact solely relying on a pre-selected baseline model. This approach tends to overlook the varying degrees of robustness different models exhibit towards the perturbations impact, overly relying on the benchmark performance of a particular model. To address this limitation, we propose \textbf{Mean Robustness Degradation (mRD)} 
for evaluating model robustness. mRD is a comprehensive measure of the average performance of models under different levels and types of perturbations. Additionally, inspired by ImageNet-C, we opt to incorporate Degradation $D$ into mRD because we found that 
different perturbations pose varying levels of difficulty. Thus, the \textbf{RD} of a perturbation $p$ is calculated as: 
\begin{equation}
    \textbf{RD}_p = \frac{1}{N}{\sum_{s=1}^{N} {\frac{D_{s,p}^g}{\text{mPE}_{s,p}}}},
\end{equation}
where mPE is from Eq.~(\ref{eq:mpe}).
The overall robustness score \textbf{mRD} is the average score of $\textbf{RD}_p$ across perturbations $p{\in}[1,12]$. Exceeding $100$ indicates the model's performance degrades more than expected due to perturbations, while falling below $100$ suggests performance improvements despite these impacts. The lower the better. This metric offers a comprehensive assessment of model robustness, enabling evaluation under diverse and realistic conditions. 
To ensure a more comprehensive evaluation, we test 
DLA models on clean datasets with mAP scores, as well as the average mAP performance on $12$ perturbations (\ie, \textbf{P-Avg}). 

\subsection{Results on PubLayNet-P}
\begin{table*}[t]
\centering
\caption{\textbf{The robustness benchmark on PubLayNet-P dataset}. V, L, and T stand for Visual, Layout, and Textual modality. `Ext.' means using extra pre-training data. mAP scores are evaluated on the \textbf{clean} data, the 12 perturbation types (\textbf{P1--P12}), and the perturbation average (\textbf{P-Avg}). \textbf{mRD} is the proposed mean Robustness Degradation, the smaller the better.}\vskip -1em
\label{tab:sota_publaynet_p}
\setlength{\tabcolsep}{3pt}
\resizebox{\linewidth}{!}{
\begin{tabu}{l|l|ccc|c|c|llllllllllll|c|c}
\toprule[1.5pt]
\multirow{2}{*}{\textbf{Backbone}}&\multirow{2}{*}{\textbf{Method}} & \multicolumn{3}{c|}{\textbf{Modality}} &\multirow{2}{*}{\textbf{Ext.}} & \multirow{2}{*}{\textbf{Clean}} & \multicolumn{3}{c|}{\textbf{Spatial}} & \multicolumn{2}{c|}{\textbf{Content}} & \multicolumn{3}{c|}{\textbf{Inconsistency}}  & \multicolumn{2}{c|}{\textbf{Blur}}  & \multicolumn{2}{c|}{\textbf{Noise}}  & \multirow{2}{*}{\textbf{P-Avg$\uparrow$}} & \multirow{2}{*}{\textbf{mRD{$\downarrow$}}}\\
& & V & L & T &&&{P1} & {P2} & \multicolumn{1}{c|}{P3} & {P4} & \multicolumn{1}{c|}{P5} & {P6} & {P7} & \multicolumn{1}{c|}{P8} & {P9} & \multicolumn{1}{c|}{P10} & {P11} & {P12} &  \\
\midrule \midrule
ResNeXt~\cite{Saining2016resnext}&LayoutParser ~\cite{shen2021layoutparser}& \cmark & \xmark & \xmark & \xmark & 89.0 & 35.8 & 78.1 & 68.0 & 84.8 & 45.9 & 79.0 & 85.8 & 82.6 & 51.4 & 31.1 & 53.6 & 00.3 & 58.0 & 212.7\\
ResNet~\cite{kaiming2015resnet}&Faster R-CNN
~\cite{ren2017FasterRcnn} & \cmark & \xmark & \xmark & \xmark & 90.2 & \textbf{44.2} & 75.3 & 74.3 & 78.9 & 53.2 & 81.4 & 82.7 & 80.2 & 80.6 & 63.2 & 55.7 & 24.3 & 66.2 & 175.5\\
ResNet~\cite{kaiming2015resnet}&DocSegTr~\cite{biswas2022docsegtr}& \cmark & \xmark & \xmark & \xmark & 90.4 & 28.3 & 76.0 & 71.2 & 85.2 & 46.2 & 69.3 & 86.2 & 68.9 & 50.8 & 19.8 & 60.2 & 00.4 & 55.2 & 233.0\\
ResNet~\cite{kaiming2015resnet}&Mask R-CNN 
~\cite{He_2017_MaskRcnn} & \cmark & \xmark & \xmark & \xmark & 91.0 & 40.0 & 74.0 & 71.8 & 76.9 & 48.7 & 79.0 & 80.6 & 77.4 & 78.2 & 54.2 & 55.1 & \textbf{31.9} & 64.0 & 192.7\\
Swin~\cite{liu2021swin}&SwinDocSegmenter~\cite{banerjee2023swindocsegmenter}&\cmark & \xmark & \xmark & \xmark  & 93.7 & 39.0 & 83.3 & 61.3 & 88.9 & 46.6 & 87.9 & 88.1 & 86.6 & 37.3 & 29.2 & 36.1 & 00.5 & 57.1 & 214.4\\
\rowfont{\color{gray!30}} DiT~\cite{li2022dit}&Cascade R-CNN~\cite{cai2018cascade} & \cmark & \xmark & \xmark & \cmark &94.5 & 31.9 & 86.5 & 82.2 & 92.1 & 79.6 & 87.2 & 92.0 & 91.6 & 93.8 & 71.3 & 69.9 & 41.8 & 76.7 & 95.8\\
\rowfont{\color{gray!30}} LayoutLMv3~\cite{huang2022layoutlmv3}&Cascade R-CNN~\cite{cai2018cascade} & \cmark & \cmark & \cmark & \cmark & 95.1 & 32.7 & 85.9 & 79.8 & 92.3 & 68.5 & 86.5 & 93.1 & 86.7 & 82.9 & 47.0 & 82.1 & 45.1 & 73.6 & 116.2\\
\midrule
Swin~\cite{liu2021swin}& Co-DINO~\cite{zong2022codetr} & \cmark & \xmark & \xmark & \xmark  & 94.3 &  22.4	&43.0	&24.7	&92.6	&56.8	&86.4	&72.6	&75.8	&35.0	&20.2	&55.4	&14.9	&50.0 &254.1\\
InternImage~\cite{wang2023internimage}& Cascade R-CNN~\cite{cai2018cascade} & \cmark & \xmark & \xmark & \xmark  & 94.1 &27.7	&80.2& 79.7&	89.6&	56.6&	83.8&	91.6&	89.6&	84.9&	54.0&	55.2&	00.2&	66.1 &141.9\\
InternImage~\cite{wang2023internimage}& DINO ~\cite{zhang2022dino}& \cmark & \xmark & \xmark & \xmark  & 95.4&34.4&	\textbf{82.2} &	\textbf{82.2}&	92.3&	57.7&	\textbf{91.7} &	\textbf{92.8}&	\textbf{92.1}&	90.9 &	63.4&	47.3&	01.3&	69.0 & 120.7 \\
InternImage~\cite{wang2023internimage}& Co-DINO~\cite{zong2022codetr} & \cmark & \xmark & \xmark & \xmark  & 94.2 & 19.1 & 48.3 & 35.7 & 91.3 & 60.6 & 87.9 & 81.4 & 87.6 & 37.9 & 27.9 & 49.7 & 11.2 & 53.2 & 230.3\\
\rowcolor[gray]{.9} InternImage~\cite{wang2023internimage}&RoDLA (Ours) & \cmark & \xmark & \xmark & \xmark &\textbf{96.0} & 31.6	&79.3	&80.6	&\textbf{92.9}	&\textbf{61.6}	&91.6	&92.6	&91.6	&\textbf{91.3}	&\textbf{67.7}	&\textbf{58.8}	&00.3	&\textbf{70.0} &\textbf{116.0}\\

\bottomrule[1.5pt]

\end{tabu}
}
\vskip -1ex
\end{table*}

Our analysis of robustness results on PubLayNet-P and test results on PubLayNet~\cite{zhong2019publaynet}, as shown in Table~\ref{tab:sota_publaynet_p}, reveals a noteworthy aspect of our RoDLA model and robustness benchmark. Initially designed to enhance robustness, RoDLA surprisingly achieves state-of-the-art performance on the clean data with $96.0\%$ in mAP. Even without additional pre-training on document-specific datasets, RoDLA maintained high performance under various perturbations, achieving $70.0\%$ in P-Avg and $116.0$ in mRD. Compared to the previous model Faster R-CNN~\cite{ren2017FasterRcnn}, it realizes a $+3.8\%$ improvement in P-Avg. This suggests that a robustness-focused architecture can inherently boost performance in diverse conditions. We observe significant performance variability among models under different perturbations like blur and noise. For example, SwinDocSegmenter~\cite{banerjee2023swindocsegmenter} drastically drops from $93.7\%$ to $37.3\%$ and $29.2\%$ under blur in mAP, whereas RoDLA shows exceptional robustness, maintaining higher mAP of $91.3\%$ and $67.7\%$. These significant results prove the effectiveness of RoDLA's robust design. Moreover, our findings indicate that extra pre-training on IIT-CDIP Test Collection 1.0 document dataset of DiT~\cite{li2022dit} and LayoutLMv3~\cite{huang2022layoutlmv3} can better address unique document perturbations, \eg, \textit{texture} and \textit{speckle}. The multimodal LayoutLMv3~\cite{huang2022layoutlmv3} does not exhibit better robustness compared to the unimodal DiT~\cite{li2022dit}. 
\subsection{Results on DocLayNet-P}
\begin{table}[t]
\centering
\caption{\textbf{The robustness benchmark on DocLayNet-P dataset}. 
}\vskip -1.em
\label{tab:sota_doclaynet_p}
\setlength{\tabcolsep}{3pt}
\resizebox{\linewidth}{!}{
\begin{tabu}{l|l|ccc|c|c|c}
\toprule[1.2pt]
\multirow{2}{*}{\textbf{Backbone}}&\multirow{2}{*}{\textbf{Method}} & \multicolumn{3}{c}{\textbf{Modality}} & \multirow{2}{*}{\textbf{Clean}} & \multirow{2}{*}{\textbf{P-Avg{$\uparrow$}}} & \multirow{2}{*}{\textbf{mRD{$\downarrow$}}}\\
& & V & L & T & &  \\
\midrule \midrule
ResNet~\cite{kaiming2015resnet}&DocSegTr~\cite{biswas2022docsegtr}& \cmark & \xmark & \xmark & 69.3 & 47.5 & 234.7\\
ResNet~\cite{kaiming2015resnet}&Mask R-CNN 
~\cite{He_2017_MaskRcnn} & \cmark & \xmark & \xmark & 73.5 & 52.7 &195.7\\
ResNet~\cite{kaiming2015resnet}&Faster R-CNN 
~\cite{ren2017FasterRcnn} & \cmark & \xmark & \xmark & 73.4 & 53.9 &189.1\\
Swin~\cite{liu2021swin}&SwinDocSegmenter~\cite{banerjee2023swindocsegmenter}&\cmark & \xmark & \xmark  & 76.9 & 58.5 &282.7\\
\rowfont{\color{gray!30}}DiT~\cite{li2022dit}& Cascade R-CNN~\cite{cai2018cascade} & \cmark & \xmark & \xmark & 62.1 & 52.1 &216.5\\
\rowfont{\color{gray!30}}LayoutLMv3~\cite{huang2022layoutlmv3}& Cascade R-CNN~\cite{cai2018cascade} & \cmark & \cmark & \cmark & 75.1 & 62.1 &172.8\\
\rowcolor[gray]{.9} InternImage~\cite{wang2023internimage}&RoDLA (Ours) & \cmark & \xmark & \xmark & \textbf{80.5} & \textbf{65.6} & \textbf{135.7}\\
\bottomrule[1.2pt]
\end{tabu}
}
\vskip -1ex
\end{table}

The results in Table~\ref{tab:sota_doclaynet_p} on DocLayNet~\cite{pfitzmann2022doclaynet} and DocLayNet-P highlights a noteworthy insight. Training on DocLayNet, which includes electronic and printed document images, enables models to learn features that are nonexistent in electronic documents, \eg, PubLayNet. Our RoDLA achieves state-of-the-art performance on both clean ($80.5\%$ in mAP) and perturbed dataset ($65.6\%$ in P-Avg and $135.7$ in mRD), which is $3.6\%$ higher in mAP, $7.1\%$ higher in P-Avg and $147.0$ lower in mRD than previous state-of-the-art model~\cite{banerjee2023swindocsegmenter}. 
Besides, the multi-modal LayoutLMv3~\cite{huang2022layoutlmv3} shows notable disparities, \ie, $5.4\%$ less than our RoDLA in clean mAP, $3.5\%$ less in P-Avg, $37.1$ higher in mRD.

\subsection{Results on M$^6$Doc-P}
\begin{table}[t]
\centering
\caption{\textbf{The robustness benchmark on M$^6$Doc-P dataset}. 
}\vskip -1em
\label{tab:sota_m6doc_p}
\setlength{\tabcolsep}{3pt}
\resizebox{\linewidth}{!}{
\begin{tabu}{l|l|ccc|c|c|c}
\toprule[1.2pt]
\multirow{2}{*}{\textbf{Backbone}} &\multirow{2}{*}{\textbf{Method}} & \multicolumn{3}{c}{\textbf{Modality}} & \multirow{2}{*}{\textbf{Clean}} & \multirow{2}{*}{\textbf{P-Avg{$\uparrow$}}} & \multirow{2}{*}{\textbf{mRD{$\downarrow$}}}\\
& & V & L & T & &  &\\
\midrule \midrule
ResNet~\cite{kaiming2015resnet}&DocSegTr~\cite{biswas2022docsegtr}& \cmark & \xmark & \xmark & 60.3 & 43.2& 212.6\\
ResNet~\cite{kaiming2015resnet}&Mask R-CNN 
~\cite{He_2017_MaskRcnn} & \cmark & \xmark & \xmark & 61.9 &48.9 &216.7\\
ResNet~\cite{kaiming2015resnet}&Faster R-CNN 
~\cite{ren2017FasterRcnn} & \cmark & \xmark & \xmark & 62.0 & 49.6 &192.2\\
Swin~\cite{liu2021swin}&SwinDocSegmenter~\cite{banerjee2023swindocsegmenter}&\cmark & \xmark & \xmark  & 47.1 & 39.7 & 239.2\\
\rowfont{\color{gray!30}}DiT~\cite{li2022dit} &Cascade R-CNN~\cite{cai2018cascade} &\cmark & \xmark & \xmark & 70.2 & 60.6 & 164.8\\
\rowfont{\color{gray!30}}LayoutLMv3~\cite{huang2022layoutlmv3} &Cascade R-CNN~\cite{cai2018cascade} &\cmark & \cmark & \cmark & 64.3 & 57.0&176.1\\
\rowcolor[gray]{.9} InternImage~\cite{wang2023internimage}&RoDLA (Ours) & \cmark & \xmark & \xmark & \textbf{70.0} & \textbf{61.7}&\textbf{147.6}\\
\bottomrule[1.2pt]
\end{tabu}
}
\vskip -2ex
\end{table}

The results on M$^6$Doc~\cite{cheng2023m6doc} are detailed in Table~\ref{tab:sota_m6doc_p}. RoDLA and DiT~\cite{li2022dit} exhibit notable performances. RoDLA achieves a balanced profile with $70.0\%$ in mAP on clean data, the highest P-Avg of $61.7\%$, and the lowest mRD of $147.6$. DiT~\cite{li2022dit}, while scoring high in clean accuracy ($70.2\%$) and P-Avg ($60.6\%$), achieves 164.8 in mRD. Other models like Faster R-CNN~\cite{ren2017FasterRcnn} and LayoutLMv3~\cite{huang2022layoutlmv3} show moderate performance but lower performance in robustness. Overall, our RoDLA is $8\%$ better in mAP, $12.1\%$ better in P-Avg and $91.6$ lower in mRD compare to other state-of-the-art models under the same pre-training. 

\subsection{Ablation Study}
\begin{table}[t]
\centering
\caption{
\textbf{Ablation study} about robustness design for layout analysis models on PubLayNet and PubLayNet-P. 
}\vskip -1em
\label{tab:ablation_study}
\setlength{\tabcolsep}{10pt}
\resizebox{\linewidth}{!}{
\begin{tabular}{l|c|c|c|c}
\toprule
\textbf{Backbone}& \textbf{Method}& \textbf{Clean} & \textbf{P-Avg$\uparrow$} &\textbf{mRD$\downarrow$}\\
\midrule \midrule
\multirow{2}{*} {{Swin~\cite{liu2021swin}}} & Cascade R-CNN &93.7 & 57.1&214.4\\
&RoDLA (Ours)  &95.6 & 69.9&124.2\\
\midrule
\multirow{4}{*} {{InternImage~\cite{wang2023internimage}}} & Cascade R-CNN &94.1 &66.1& 141.9\\
& Co-DINO &94.2 & 53.2&230.3\\
& DINO &95.4 &69.1&120.7\\
& RoDLA (Ours) &\textbf{96.0} & \textbf{70.0}&\textbf{116.0}\\
\bottomrule
\end{tabular}
}
\vskip -1ex
\end{table}
An ablation study on the RoDLA model was conducted to assess the impact of its components on performance in robustness benchmarks and on clean datasets. 

\noindent \textbf{Effect of Backbone.}
The comparison in Table~\ref{tab:ablation_study} is based on Swin~\cite{liu2021swin} and InternImage~\cite{wang2023internimage} as the backbone. 
With InternImage~\cite{wang2023internimage}, the model provides an advance of $0.4\%$ in mAP on clean dataset and a $9\%$ gain in P-Avg, a $72.5$ lower mRD compared to Swin~\cite{liu2021swin}, which reveals the exceptional performance in feature extraction by InternImage~\cite{wang2023internimage}.  However, when employing our RoDLA method, the performance gap narrows: Swin~\cite{liu2021swin} trails by only $0.4\%$ in mAP on clean dataset, $0.1\%$ in P-Avg, and $8.2$ higher in mRD relative to InternImage~\cite{wang2023internimage} as the backbone. This suggests that InternImage~\cite{wang2023internimage} as a backbone outperforms Swin~\cite{liu2021swin} in both performance and robustness under the same pre-training and method conditions. Our RoDLA method effectively harnesses robust features and reduces the performance gap.

\noindent \textbf{Effect of Method.}
When using InternImage as the backbone and comparing different methods, \eg, Co-DINO~\cite{zong2022codetr}, despite a slight improvement in clean dataset performance compared to Cascade R-CNN, shows a significant decrease in P-Avg by $12.9\%$ and an increase in mRD by $88.3$, indicating a drop in robustness. Our RoDLA method 
shows an improvement over the Cascade R-CNN method. It achieves a $1.9\%$ gain in mAP, a $3.9\%$ rise in P-Avg, and a substantial $25.9$ reduction in mRD, indicating enhanced robustness.

\noindent \textbf{Analysis of Robustness Design.}
\begin{table}[t]
\centering
\caption{\textbf{Analysis of robustness design} for RoDLA on PubLayNet and PubLayNet-P. All methods use InternImage. CA: Channel Attention~\cite{zhou2022fan}. APL: Average Pooling Layer~\cite{guo2023tap}.}
\vskip -1em
\label{tab:robustness_design}
\setlength{\tabcolsep}{4pt}
\resizebox{\linewidth}{!}{
\begin{tabular}{cccc|c|c|c|c}
\toprule
\textbf{CA Encoder}& \textbf{CA Decoder}&\textbf{APL Encoder}&\textbf{APL Decoder}&\textbf{\#Params} & \textbf{Clean} & \textbf{P-Avg$\uparrow$} & \textbf{mRD$\downarrow$}\\
\midrule \midrule
&&& &335.3M& 95.4 & 69.0&120.4\\ 
\hdashline\noalign{\vskip 0.5ex}
\cmark&&& &335.7M& 95.7 & 67.8&127.3\\
&\cmark&& &335.7M& 95.7 & 65.7&139.0\\
\cmark&\cmark&& &336.1M& 95.7 & 66.5&133.8\\ 
\hdashline\noalign{\vskip 0.5ex}
&&\cmark& & 320.0M & 96.0  & 69.1 & 120.1\\
&&&\cmark &335.4M & 96.0 & 64.2 & 127.4\\
&&\cmark&\cmark &320.0M & 96.0 & 67.8 & 126.2\\
\hdashline\noalign{\vskip 0.5ex}
\cmark&&\cmark& &323.2M& 96.0 & \textbf{70.0}&\textbf{115.7}\\
&\cmark&&\cmark & 335.9M & \textbf{96.1} & 69.1&121.5\\
\cmark&\cmark&\cmark&\cmark & 323.8M & 95.9 & 67.0&132.3\\
\bottomrule
\end{tabular}
}
\vskip -1.em
\end{table}

To explore the robustness performance related to model structural designs, we conduct a detailed ablation study of RoDLA 
in Table~\ref{tab:robustness_design}. 
By integrating Channel Attention (CA) into the DINO~\cite{zhang2022dino} structure, we found that regardless of placement, adding CA can maintain $95.7\%$ mAP with clean data. CA in different positions led to P-Avg decreases of $1.2\%$, $3.3\%$, and $2.5\%$, and mRD increases of $6.9$, $18.6$, and $13.4$. These results suggest strategic placement of CA is crucial for balancing detail sensitivity and robustness of model.
Besides, introducing Average Pooling Layers (APL) to mitigate irrelative attention on damaged tokens, we found the optimal way to be within the Encoder. This adjustment improved performance across all metrics, get benefits of $0.9\%$ in P-Avg and $5.8$ in mRD compare to intersecting in Decoder and $3.0\%$ in P-Avg and $16.6$ in mRD compare to placement in both. Besides, our robustness design has fewer parameters. 

\noindent \textbf{Analysis of Multiple Perturbations.}
To analyze multiple perturbations, we conduct an experiment based on the superposition of five perturbations. The results are illustrated in Table~\ref{tab:multi_effects}. As the number of perturbations increases, \textit{i.e.}, from simple to complex perturbations, the mPE score increases and the model performance (P-Avg) decreases. This trend indicates that the model robustness is more susceptible to degradation with more perturbations. Nonetheless, our RoDLA method can obtain robust performance. 

\begin{table}[t]
\centering
\caption{\textbf{Analysis of Multiple Perturbations} on M$^6$Doc-P.}
\label{tab:multi_effects}
\vskip -1.em
\setlength{\tabcolsep}{3pt}
\renewcommand\arraystretch{1.1}
\resizebox{\linewidth}{!}{
\begin{tabular}{l|c|c}
\toprule
\textbf{Perturbations}&\textbf{mPE$\downarrow$}& \textbf{RoDLA P-Avg$\uparrow$}\\
\midrule \midrule
Warping + Watermark & 24.1 & 64.9\\
Warping + Watermark + Illumination & 29.0 & 62.0\\
Warping + Watermark + Illumination + Defocus & 32.3 & 50.8\\
Warping + Watermark + Illumination + Defocus + Speckle & 46.2 & 41.9\\
\bottomrule[1.2pt]
\end{tabular}
}
\vskip -1em
\end{table}
\section{Conclusion}
\label{conclusion}
In this work, we introduce the first robustness benchmark for Document Layout Analysis (DLA) models. Inspired by real-world document processing, we create a taxonomy with $12$ common perturbations with 3 level of severity in document images. The benchmark includes almost 450,000 documents from 3 datasets. 
To evaluate the impact of document perturbations, we propose two metrics, \ie, \textit{mean Perturbation Effect (mPE)} and \textit{mean Robustness Degradation (mRD)}. 
We hope the benchmark and these metrics can enable the future work for robustness analysis of DLA models. 
Besides, we propose a novel model, \textit{Robust Document Layout Analyzer (RoDLA)}. 
Extensive experiments on three datasets prove the effectiveness of our methods. RoDLA can obtain state-of-the-art performance on the perturbed and the clean data. 
We hope our work will establish a solid benchmark for evaluating the robustness of DLA models and foster the advancement of document understanding.

\noindent \textbf{Limitations.} The proposed robustness benchmark is constrained 
on the vision-based single-modal DLA models. The benchmark can be extended to cover multi-modal DLA models
. Besides, performing human-in-the-loop testing to evaluate the robustness of DLA models would be a crucial step before deploying in real-world applications. 
\clearpage
{\small
\bibliographystyle{ieee_fullname}
\bibliography{main}

\begin{thebibliography}{75}
\providecommand{\natexlab}[1]{#1}
\providecommand{\url}[1]{\texttt{#1}}
\expandafter\ifx\csname urlstyle\endcsname\relax
  \providecommand{\doi}[1]{doi: #1}\else
  \providecommand{\doi}{doi: \begingroup \urlstyle{rm}\Url}\fi

\bibitem[Appalaraju et~al.(2021)Appalaraju, Jasani, Kota, Xie, and Manmatha]{Appalaraju_2021_DocFormer}
Srikar Appalaraju, Bhavan Jasani, Bhargava~Urala Kota, Yusheng Xie, and R. Manmatha.
\newblock {DocFormer: End-to-End Transformer for Document Understanding}.
\newblock In \emph{ICCV}, 2021.

\bibitem[Arroyo et~al.(2021)Arroyo, Postels, and Tombari]{arroyo2021variational}
Diego~Martin Arroyo, Janis Postels, and Federico Tombari.
\newblock Variational transformer networks for layout generation.
\newblock In \emph{CVPR}, 2021.

\bibitem[Auer et~al.(2023)Auer, Nassar, Lysak, Dolfi, Livathinos, and Staar]{Auer2023ICDAR2C}
Christoph Auer, Ahmed~Samy Nassar, Maksym Lysak, Michele Dolfi, Nikolaos Livathinos, and Peter W.~J. Staar.
\newblock Icdar 2023 competition on robust layout segmentation in corporate documents.
\newblock \emph{ArXiv}, 2023.

\bibitem[Banerjee et~al.(2023)Banerjee, Biswas, Llad{\'o}s, and Pal]{banerjee2023swindocsegmenter}
Ayan Banerjee, Sanket Biswas, Josep Llad{\'o}s, and Umapada Pal.
\newblock {SwinDocSegmenter: An End-to-End Unified Domain Adaptive Transformer for Document Instance Segmentation}.
\newblock \emph{ICDAR}, 2023.

\bibitem[Biswas et~al.(2022)Biswas, Banerjee, Llad{\'o}s, and Pal]{biswas2022docsegtr}
Sanket Biswas, Ayan Banerjee, Josep Llad{\'o}s, and Umapada Pal.
\newblock {DocSegTr: An Instance-Level End-to-End Document Image Segmentation Transformer}.
\newblock \emph{arXiv preprint arXiv:2201.11438}, 2022.

\bibitem[Cai and Vasconcelos(2018)]{cai2018cascade}
Zhaowei Cai and Nuno Vasconcelos.
\newblock {Cascade r-cnn: Delving into high quality object detection}.
\newblock In \emph{CVPR}, 2018.

\bibitem[Chen et~al.(2019)Chen, Wang, Pang, Cao, Xiong, Li, Sun, Feng, Liu, Xu, Zhang, Cheng, Zhu, Cheng, Zhao, Li, Lu, Zhu, Wu, Dai, Wang, Shi, Ouyang, Loy, and Lin]{mmdetection}
Kai Chen, Jiaqi Wang, Jiangmiao Pang, Yuhang Cao, Yu Xiong, Xiaoxiao Li, Shuyang Sun, Wansen Feng, Ziwei Liu, Jiarui Xu, Zheng Zhang, Dazhi Cheng, Chenchen Zhu, Tianheng Cheng, Qijie Zhao, Buyu Li, Xin Lu, Rui Zhu, Yue Wu, Jifeng Dai, Jingdong Wang, Jianping Shi, Wanli Ouyang, Chen~Change Loy, and Dahua Lin.
\newblock {MMDetection}: Open mmlab detection toolbox and benchmark.
\newblock \emph{arXiv preprint arXiv:1906.07155}, 2019.

\bibitem[Cheng et~al.(2023)Cheng, Zhang, Wu, Zhang, Zhu, Xie, Li, Ding, and Jin]{cheng2023m6doc}
Hiuyi Cheng, Peirong Zhang, Sihang Wu, Jiaxin Zhang, Qiyuan Zhu, Zecheng Xie, Jing Li, Kai Ding, and Lianwen Jin.
\newblock {M6Doc: A Large-Scale Multi-Format, Multi-Type, Multi-Layout, Multi-Language, Multi-Annotation Category Dataset for Modern Document Layout Analysis}.
\newblock In \emph{CVPR}, 2023.

\bibitem[Coquenet et~al.(2023)Coquenet, Chatelain, and Paquet]{coquenet2023dan}
Denis Coquenet, Cl{\'e}ment Chatelain, and Thierry Paquet.
\newblock Dan: a segmentation-free document attention network for handwritten document recognition.
\newblock \emph{TPAMI}, 2023.

\bibitem[Deng et~al.(2009)Deng, Dong, Socher, Li, Li, and Fei-Fei]{deng2009imagenet}
Jia Deng, Wei Dong, Richard Socher, Li-Jia Li, Kai Li, and Li Fei-Fei.
\newblock Imagenet: A large-scale hierarchical image database.
\newblock In \emph{2009 IEEE Conference on Computer Vision and Pattern Recognition}, pages 248--255, 2009.

\bibitem[Diem et~al.(2011)Diem, Kleber, and Sablatnig]{diem2011text}
Markus Diem, Florian Kleber, and Robert Sablatnig.
\newblock Text classification and document layout analysis of paper fragments.
\newblock In \emph{ICDAR}, 2011.

\bibitem[Ding et~al.(2021)Ding, Ma, Wang, and Simoncelli]{ding2021IQA}
Keyan Ding, Kede Ma, Shiqi Wang, and Eero~P Simoncelli.
\newblock {Comparison of full-reference image quality models for optimization of image processing systems}.
\newblock \emph{IJCV}, 2021.

\bibitem[Dong et~al.(2020)Dong, Fu, Yang, Pang, Su, Xiao, and Zhu]{dong2020benchmarking}
Yinpeng Dong, Qi-An Fu, Xiao Yang, Tianyu Pang, Hang Su, Zihao Xiao, and Jun Zhu.
\newblock Benchmarking adversarial robustness on image classification.
\newblock In \emph{CVPR}, 2020.

\bibitem[Du et~al.(2022)Du, Wang, Gozum, and Li]{du2022unknown}
Xuefeng Du, Xin Wang, Gabriel Gozum, and Yixuan Li.
\newblock Unknown-aware object detection: Learning what you don't know from videos in the wild.
\newblock In \emph{CVPR}, 2022.

\bibitem[Feng et~al.(2023)Feng, Liu, Deng, Zhou, and Li]{feng2023doctr}
Hao Feng, Shaokai Liu, Jiajun Deng, Wengang Zhou, and Houqiang Li.
\newblock {Deep Unrestricted Document Image Rectification}.
\newblock \emph{arXiv preprint arXiv:2304.08796}, 2023.

\bibitem[Fronteau et~al.(2023)Fronteau, Paran, and Shabou]{fronteau2023robustDocCls}
Timoth{\'e}e Fronteau, Arnaud Paran, and Aymen Shabou.
\newblock {Evaluating Adversarial Robustness on Document Image Classification}.
\newblock \emph{arXiv preprint arXiv:2304.12486}, 2023.

\bibitem[Garz et~al.(2010)Garz, Diem, and Sablatnig]{garz2010detecting}
Angelika Garz, Markus Diem, and Robert Sablatnig.
\newblock Detecting text areas and decorative elements in ancient manuscripts.
\newblock In \emph{ICFHR}, 2010.

\bibitem[Gemelli et~al.(2022)Gemelli, Biswas, Civitelli, Llad{\'o}s, and Marinai]{gemelli2022doc2graph}
Andrea Gemelli, Sanket Biswas, Enrico Civitelli, Josep Llad{\'o}s, and Simone Marinai.
\newblock Doc2graph: a task agnostic document understanding framework based on graph neural networks.
\newblock In \emph{ECCV}, 2022.

\bibitem[Guo et~al.(2023)Guo, Stutz, and Schiele]{guo2023tap}
Yong Guo, David Stutz, and Bernt Schiele.
\newblock {Robustifying token attention for vision transformers}.
\newblock In \emph{ICCV}, 2023.

\bibitem[Hahner et~al.(2022)Hahner, Sakaridis, Bijelic, Heide, Yu, Dai, and Van~Gool]{hahner2022lidar}
Martin Hahner, Christos Sakaridis, Mario Bijelic, Felix Heide, Fisher Yu, Dengxin Dai, and Luc Van~Gool.
\newblock Lidar snowfall simulation for robust 3d object detection.
\newblock In \emph{CVPR}, 2022.

\bibitem[Harley et~al.(2015)Harley, Ufkes, and Derpanis]{harley2015RVL_CDIP}
Adam~W Harley, Alex Ufkes, and Konstantinos~G Derpanis.
\newblock {Evaluation of Deep Convolutional Nets for Document Image Classification and Retrieval}.
\newblock In \emph{ICDAR}. IEEE, 2015.

\bibitem[He et~al.(2015)He, Zhang, Ren, and Sun]{kaiming2015resnet}
Kaiming He, Xiangyu Zhang, Shaoqing Ren, and Jian Sun.
\newblock Deep residual learning for image recognition.
\newblock \emph{CoRR}, abs/1512.03385, 2015.

\bibitem[He et~al.(2017)He, Gkioxari, Dollar, and Girshick]{He_2017_MaskRcnn}
Kaiming He, Georgia Gkioxari, Piotr Dollar, and Ross Girshick.
\newblock {Mask R-CNN}.
\newblock In \emph{ICCV}, 2017.

\bibitem[Hegghammer(2022)]{Hegghammer2022ocrbenchmark}
Thomas Hegghammer.
\newblock Ocr with tesseract, amazon textract, and google document ai: a benchmarking experiment.
\newblock \emph{Journal of Computational Social Science}, 5\penalty0 (1):\penalty0 861--882, 2022.

\bibitem[Hendrycks and Dietterich(2019)]{hendrycks2019imagenetc}
Dan Hendrycks and Thomas Dietterich.
\newblock Benchmarking neural network robustness to common corruptions and perturbations.
\newblock \emph{Proceedings of the International Conference on Learning Representations}, 2019.

\bibitem[Huang et~al.(2022)Huang, Lv, Cui, Lu, and Wei]{huang2022layoutlmv3}
Yupan Huang, Tengchao Lv, Lei Cui, Yutong Lu, and Furu Wei.
\newblock {Layoutlmv3: Pre-training for document ai with unified text and image masking}.
\newblock In \emph{ACMMM}, 2022.

\bibitem[Jiang et~al.(2022)Jiang, Long, Xue, Yang, Yao, and Xia]{jiang2022dewarping}
X. Jiang, R. Long, N. Xue, Z. Yang, C. Yao, and G. Xia.
\newblock Revisiting document image dewarping by grid regularization.
\newblock In \emph{2022 IEEE/CVF Conference on Computer Vision and Pattern Recognition (CVPR)}, pages 4533--4542, Los Alamitos, CA, USA, 2022. IEEE Computer Society.

\bibitem[Kamann and Rother(2020)]{Kamann2020RobustnessSemantic}
Christoph Kamann and Carsten Rother.
\newblock Benchmarking the robustness of semantic segmentation models.
\newblock In \emph{Proceedings of the IEEE/CVF Conference on Computer Vision and Pattern Recognition (CVPR)}, 2020.

\bibitem[Krishnan and Jawahar(2019)]{Krishnan2019handwriten}
Praveen Krishnan and C.~V. Jawahar.
\newblock Hwnet v2: an efficient word image representation for handwritten documents.
\newblock \emph{International Journal on Document Analysis and Recognition (IJDAR)}, 22\penalty0 (4):\penalty0 387--405, 2019.

\bibitem[Lewis et~al.(2006)Lewis, Agam, Argamon, Frieder, Grossman, and Heard]{Lewis2006IITCDIP}
D. Lewis, G. Agam, S. Argamon, O. Frieder, D. Grossman, and J. Heard.
\newblock Building a test collection for complex document information processing.
\newblock In \emph{Proceedings of the 29th Annual International ACM SIGIR Conference on Research and Development in Information Retrieval}, page 665–666, New York, NY, USA, 2006. Association for Computing Machinery.

\bibitem[Li et~al.(2022)Li, Xu, Lv, Cui, Zhang, and Wei]{li2022dit}
Junlong Li, Yiheng Xu, Tengchao Lv, Lei Cui, Cha Zhang, and Furu Wei.
\newblock {DiT: Self-supervised Pre-training for Document Image Transformer}.
\newblock \emph{arXiv preprint arXiv:2203.02378}, 2022.

\bibitem[Li et~al.(2020)Li, Xu, Cui, Huang, Wei, Li, and Zhou]{li2020docbank}
Minghao Li, Yiheng Xu, Lei Cui, Shaohan Huang, Furu Wei, Zhoujun Li, and Ming Zhou.
\newblock Docbank: A benchmark dataset for document layout analysis.
\newblock \emph{arXiv preprint arXiv:2006.01038}, 2020.

\bibitem[Li et~al.(2023)Li, Chen, Zhu, Wang, Zhang, and Xue]{Li_2023_imagenete}
Xiaodan Li, Yuefeng Chen, Yao Zhu, Shuhui Wang, Rong Zhang, and Hui Xue.
\newblock Imagenet-e: Benchmarking neural network robustness via attribute editing.
\newblock In \emph{Proceedings of the IEEE/CVF Conference on Computer Vision and Pattern Recognition (CVPR)}, pages 20371--20381, 2023.

\bibitem[Li et~al.(2021)Li, Qian, Yu, Qin, Zhang, Liu, Yao, Han, Liu, and Ding]{li2021structext}
Yulin Li, Yuxi Qian, Yuechen Yu, Xiameng Qin, Chengquan Zhang, Yan Liu, Kun Yao, Junyu Han, Jingtuo Liu, and Errui Ding.
\newblock {StrucTexT: Structured Text Understanding with Multi-Modal Transformers}.
\newblock In \emph{ACMMM}, 2021.

\bibitem[Liu et~al.(2021)Liu, Lin, Cao, Hu, Wei, Zhang, Lin, and Guo]{liu2021swin}
Z. Liu, Y. Lin, Y. Cao, H. Hu, Y. Wei, Z. Zhang, S. Lin, and B. Guo.
\newblock Swin transformer: Hierarchical vision transformer using shifted windows.
\newblock In \emph{2021 IEEE/CVF International Conference on Computer Vision (ICCV)}, pages 9992--10002, Los Alamitos, CA, USA, 2021. IEEE Computer Society.

\bibitem[Long et~al.(2022)Long, Qin, Panteleev, Bissacco, Fujii, and Raptis]{long2022towards}
Shangbang Long, Siyang Qin, Dmitry Panteleev, Alessandro Bissacco, Yasuhisa Fujii, and Michalis Raptis.
\newblock Towards end-to-end unified scene text detection and layout analysis.
\newblock In \emph{CVPR}, 2022.

\bibitem[Luo et~al.(2022)Luo, Tang, Zheng, Yao, Jin, Li, Xue, and Si]{Luo2022BiVLDocBV}
Chuwei Luo, Guozhi Tang, Qi Zheng, Cong Yao, Lianwen Jin, Chenliang Li, Yang Xue, and Luo Si.
\newblock Bi-vldoc: Bidirectional vision-language modeling for visually-rich document understanding.
\newblock \emph{ArXiv}, 2022.

\bibitem[Luo et~al.(2023)Luo, Cheng, Zheng, and Yao]{luo2023geolayoutlm}
Chuwei Luo, Changxu Cheng, Qi Zheng, and Cong Yao.
\newblock Geolayoutlm: Geometric pre-training for visual information extraction.
\newblock In \emph{CVPR}, 2023.

\bibitem[Michaelis et~al.(2019)Michaelis, Mitzkus, Geirhos, Rusak, Bringmann, Ecker, Bethge, and Brendel]{michaelis2019benchmarking}
Claudio Michaelis, Benjamin Mitzkus, Robert Geirhos, Evgenia Rusak, Oliver Bringmann, Alexander~S Ecker, Matthias Bethge, and Wieland Brendel.
\newblock Benchmarking robustness in object detection: Autonomous driving when winter is coming.
\newblock \emph{arXiv preprint arXiv:1907.07484}, 2019.

\bibitem[Milyaev and Laptev(2017)]{milyaev2017towards}
S Milyaev and I Laptev.
\newblock Towards reliable object detection in noisy images.
\newblock \emph{TPAMI}, 2017.

\bibitem[Modas et~al.(2022)Modas, Rade, Ortiz-Jim{\'e}nez, Moosavi-Dezfooli, and Frossard]{modas2022prime}
Apostolos Modas, Rahul Rade, Guillermo Ortiz-Jim{\'e}nez, Seyed-Mohsen Moosavi-Dezfooli, and Pascal Frossard.
\newblock Prime: A few primitives can boost robustness to common corruptions.
\newblock In \emph{ECCV}, 2022.

\bibitem[Moured et~al.(2023)Moured, Zhang, Roitberg, Schwarz, and Stiefelhagen]{moured2023line}
Omar Moured, Jiaming Zhang, Alina Roitberg, Thorsten Schwarz, and Rainer Stiefelhagen.
\newblock Line graphics digitization: A step towards full automation.
\newblock In \emph{ICDAR}, pages 438--453. Springer, 2023.

\bibitem[Papakipos and Bitton(2022)]{papakipos2022augly}
Zoe Papakipos and Joanna Bitton.
\newblock Augly: Data augmentations for robustness.
\newblock \emph{arXiv preprint arXiv:2201.06494}, 2022.

\bibitem[Pei et~al.(2021)Pei, Huang, Zou, Zhang, and Wang]{8889765}
Yanting Pei, Yaping Huang, Qi Zou, Xingyuan Zhang, and Song Wang.
\newblock Effects of image degradation and degradation removal to cnn-based image classification.
\newblock \emph{TPAMI}, 2021.

\bibitem[Peng et~al.(2022)Peng, Pan, Wang, Luo, Zhang, Huang, Hu, Yin, Chen, Zhang, et~al.]{peng2022ernie}
Qiming Peng, Yinxu Pan, Wenjin Wang, Bin Luo, Zhenyu Zhang, Zhengjie Huang, Teng Hu, Weichong Yin, Yongfeng Chen, Yin Zhang, et~al.
\newblock Ernie-layout: Layout knowledge enhanced pre-training for visually-rich document understanding.
\newblock \emph{arXiv preprint arXiv:2210.06155}, 2022.

\bibitem[Pfitzmann et~al.(2022)Pfitzmann, Auer, Dolfi, Nassar, and Staar]{pfitzmann2022doclaynet}
Birgit Pfitzmann, Christoph Auer, Michele Dolfi, Ahmed~S Nassar, and Peter Staar.
\newblock {Doclaynet: A large human-annotated dataset for document-layout segmentation}.
\newblock In \emph{SIGKDD}, 2022.

\bibitem[Piao et~al.(2022)Piao, Wu, Zhang, Jiang, and Lu]{piao2022noise}
Yongri Piao, Wei Wu, Miao Zhang, Yongyao Jiang, and Huchuan Lu.
\newblock Noise-sensitive adversarial learning for weakly supervised salient object detection.
\newblock \emph{TMM}, 2022.

\bibitem[Ren et~al.(2017)Ren, He, Girshick, and Sun]{ren2017FasterRcnn}
Shaoqing Ren, Kaiming He, Ross Girshick, and Jian Sun.
\newblock {Faster R-CNN: Towards Real-Time Object Detection with Region Proposal Networks}.
\newblock \emph{TPAMI}, 2017.

\bibitem[Russakovsky et~al.(2015)Russakovsky, Deng, Su, Krause, Satheesh, Ma, Huang, Karpathy, Khosla, Bernstein, Berg, and Fei-Fei]{Russakovsky2015ILSVRC}
Olga Russakovsky, Jia Deng, Hao Su, Jonathan Krause, Sanjeev Satheesh, Sean Ma, Zhiheng Huang, Andrej Karpathy, Aditya Khosla, Michael Bernstein, Alexander~C. Berg, and Li Fei-Fei.
\newblock Imagenet large scale visual recognition challenge.
\newblock \emph{International Journal of Computer Vision}, 115\penalty0 (3):\penalty0 211--252, 2015.

\bibitem[Saifullah et~al.(2022)Saifullah, Siddiqui, Agne, Dengel, and Ahmed]{saifullah2022distortionstudy}
Saifullah, Shoaib~Ahmed Siddiqui, Stefan Agne, Andreas Dengel, and Sheraz Ahmed.
\newblock Are deep models robust against real distortions? a case study on document image classification.
\newblock In \emph{2022 26th International Conference on Pattern Recognition (ICPR)}, pages 1628--1635, 2022.

\bibitem[Sampat et~al.(2009)Sampat, Wang, Gupta, Bovik, and Markey]{Sampat2009cwssim}
Mehul~P. Sampat, Zhou Wang, Shalini Gupta, Alan~Conrad Bovik, and Mia~K. Markey.
\newblock Complex wavelet structural similarity: A new image similarity index.
\newblock \emph{IEEE Transactions on Image Processing}, 18\penalty0 (11):\penalty0 2385--2401, 2009.

\bibitem[Shen et~al.(2020{\natexlab{a}})Shen, Ji, Chen, Hong, Zheng, Liu, Xu, and Tian]{shen2020noise}
Yunhang Shen, Rongrong Ji, Zhiwei Chen, Xiaopeng Hong, Feng Zheng, Jianzhuang Liu, Mingliang Xu, and Qi Tian.
\newblock Noise-aware fully webly supervised object detection.
\newblock In \emph{CVPR}, 2020{\natexlab{a}}.

\bibitem[Shen et~al.(2020{\natexlab{b}})Shen, Zhang, and Dell]{shen2020large}
Zejiang Shen, Kaixuan Zhang, and Melissa Dell.
\newblock A large dataset of historical japanese documents with complex layouts.
\newblock In \emph{CVPR}, 2020{\natexlab{b}}.

\bibitem[Shen et~al.(2021)Shen, Zhang, Dell, Lee, Carlson, and Li]{shen2021layoutparser}
Zejiang Shen, Ruochen Zhang, Melissa Dell, Benjamin Charles~Germain Lee, Jacob Carlson, and Weining Li.
\newblock {Layoutparser: A unified toolkit for deep learning based document image analysis}.
\newblock In \emph{ICDAR}, 2021.

\bibitem[Shihab et~al.(2023)Shihab, Hasan, Emon, Hossen, Ansary, Ahmed, Rakib, Dhruvo, Dip, Pavel, et~al.]{shihab2023badlad}
Md~Istiak~Hossain Shihab, Md~Rakibul Hasan, Mahfuzur~Rahman Emon, Syed~Mobassir Hossen, Md~Nazmuddoha Ansary, Intesur Ahmed, Fazle~Rabbi Rakib, Shahriar~Elahi Dhruvo, Souhardya~Saha Dip, Akib~Hasan Pavel, et~al.
\newblock Badlad: A large multi-domain bengali document layout analysis dataset.
\newblock In \emph{ICDAR}, 2023.

\bibitem[Sulaiman et~al.(2019)Sulaiman, Omar, and Nasrudin]{jimaging2019degrated}
Alaa Sulaiman, Khairuddin Omar, and Mohammad~F. Nasrudin.
\newblock Degraded historical document binarization: A review on issues, challenges, techniques, and future directions.
\newblock \emph{Journal of Imaging}, 5\penalty0 (4), 2019.

\bibitem[Tang et~al.(2023)Tang, Yang, Wang, Fang, Liu, Zhu, Zeng, Zhang, and Bansal]{tang2023unifying}
Zineng Tang, Ziyi Yang, Guoxin Wang, Yuwei Fang, Yang Liu, Chenguang Zhu, Michael Zeng, Cha Zhang, and Mohit Bansal.
\newblock Unifying vision, text, and layout for universal document processing.
\newblock In \emph{CVPR}, 2023.

\bibitem[Tran et~al.(2017)Tran, Oh, Na, Lee, Yang, and Kim]{Tran2017ARS}
Tuan~Anh Tran, Kang~Han Oh, In~Seop Na, Gueesang Lee, Hyung-Jeong Yang, and Soohyung Kim.
\newblock A robust system for document layout analysis using multilevel homogeneity structure.
\newblock \emph{Expert Syst. Appl.}, 2017.

\bibitem[Wang et~al.(2022{\natexlab{a}})Wang, Jin, and Ding]{wang2022lilt}
Jiapeng Wang, Lianwen Jin, and Kai Ding.
\newblock {LiLT: A Simple yet Effective Language-Independent Layout Transformer for Structured Document Understanding}.
\newblock In \emph{ACL}, 2022{\natexlab{a}}.

\bibitem[Wang et~al.(2022{\natexlab{b}})Wang, Qin, Zhou, Lu, and Zhang]{wang2022ryolo}
Lucai Wang, Hongda Qin, Xuanyu Zhou, Xiao Lu, and Fengting Zhang.
\newblock {R-YOLO: A Robust Object Detector in Adverse Weather}.
\newblock \emph{TIM}, 2022{\natexlab{b}}.

\bibitem[Wang et~al.(2023)Wang, Dai, Chen, Huang, Li, Zhu, Hu, Lu, Lu, Li, et~al.]{wang2023internimage}
Wenhai Wang, Jifeng Dai, Zhe Chen, Zhenhang Huang, Zhiqi Li, Xizhou Zhu, Xiaowei Hu, Tong Lu, Lewei Lu, Hongsheng Li, et~al.
\newblock {Internimage: Exploring large-scale vision foundation models with deformable convolutions}.
\newblock In \emph{CVPR}, 2023.

\bibitem[Wang et~al.(2003)Wang, Simoncelli, and Bovik]{wang2003msssim}
Z. Wang, Eero Simoncelli, and Alan Bovik.
\newblock Multiscale structural similarity for image quality assessment.
\newblock pages 1398 -- 1402 Vol.2, 2003.

\bibitem[Xie et~al.(2016)Xie, Girshick, Doll{\'{a}}r, Tu, and He]{Saining2016resnext}
Saining Xie, Ross~B. Girshick, Piotr Doll{\'{a}}r, Zhuowen Tu, and Kaiming He.
\newblock Aggregated residual transformations for deep neural networks.
\newblock \emph{CoRR}, abs/1611.05431, 2016.

\bibitem[Xu et~al.(2020)Xu, Li, Cui, Huang, Wei, and Zhou]{xu2020layoutlm}
Yiheng Xu, Minghao Li, Lei Cui, Shaohan Huang, Furu Wei, and Ming Zhou.
\newblock {Layoutlm: Pre-training of text and layout for document image understanding}.
\newblock In \emph{KDD}, 2020.

\bibitem[Xu et~al.(2021)Xu, Xu, Lv, Cui, Wei, Wang, Lu, Florencio, Zhang, Che, Zhang, and Zhou]{xu-etal-2021-layoutlmv2}
Yang Xu, Yiheng Xu, Tengchao Lv, Lei Cui, Furu Wei, Guoxin Wang, Yijuan Lu, Dinei Florencio, Cha Zhang, Wanxiang Che, Min Zhang, and Lidong Zhou.
\newblock {{L}ayout{LM}v2: Multi-modal Pre-training for Visually-rich Document Understanding}.
\newblock In \emph{ACL}, 2021.

\bibitem[Yan et~al.(2023)Yan, Zheng, Li, Cui, and Dai]{yan2023RobustLidar}
Xu Yan, Chaoda Zheng, Zhen Li, Shuguang Cui, and Dengxin Dai.
\newblock Benchmarking the robustness of lidar semantic segmentation models.
\newblock \emph{arXiv preprint arXiv:2301.00970}, 2023.

\bibitem[Yang and Hsu(2022)]{yang2022transformer}
Huichen Yang and William Hsu.
\newblock Transformer-based approach for document layout understanding.
\newblock In \emph{ICIP}, 2022.

\bibitem[Yang et~al.(2017)Yang, Yumer, Asente, Kraley, Kifer, and Lee~Giles]{yang2017learning}
Xiao Yang, Ersin Yumer, Paul Asente, Mike Kraley, Daniel Kifer, and C Lee~Giles.
\newblock Learning to extract semantic structure from documents using multimodal fully convolutional neural networks.
\newblock In \emph{CVPR}, 2017.

\bibitem[Zhang et~al.(2022)Zhang, Li, Liu, Zhang, Su, Zhu, Ni, and Shum]{zhang2022dino}
Hao Zhang, Feng Li, Shilong Liu, Lei Zhang, Hang Su, Jun Zhu, Lionel~M Ni, and Heung-Yeung Shum.
\newblock {DINO: DETR with improved denoising anchor boxes for end-to-end object detection}.
\newblock \emph{arXiv preprint arXiv:2203.03605}, 2022.

\bibitem[Zhang et~al.(2023)Zhang, Cao, Liu, Niu, Meng, and Zhou]{Zhang2023WeLayoutWL}
Mingliang Zhang, Zhen Cao, Juntao Liu, Liqiang Niu, Fandong Meng, and Jie Zhou.
\newblock Welayout: Wechat layout analysis system for the icdar 2023 competition on robust layout segmentation in corporate documents.
\newblock \emph{ArXiv}, 2023.

\bibitem[Zhang et~al.(2021)Zhang, Li, Qiao, Cheng, Pu, Niu, and Wu]{zhang2021vsr}
Peng Zhang, Can Li, Liang Qiao, Zhanzhan Cheng, Shiliang Pu, Yi Niu, and Fei Wu.
\newblock {VSR: a unified framework for document layout analysis combining vision, semantics and relations}.
\newblock In \emph{ICDAR}, 2021.

\bibitem[Zhong et~al.(2019)Zhong, Tang, and Yepes]{zhong2019publaynet}
Xu Zhong, Jianbin Tang, and Antonio~Jimeno Yepes.
\newblock {PubLayNet: largest dataset ever for document layout analysis}.
\newblock In \emph{ICDAR}, 2019.

\bibitem[Zhou et~al.(2022)Zhou, Yu, Xie, Xiao, Anandkumar, Feng, and Alvarez]{zhou2022fan}
Daquan Zhou, Zhiding Yu, Enze Xie, Chaowei Xiao, Animashree Anandkumar, Jiashi Feng, and Jose~M Alvarez.
\newblock {Understanding The Robustness in Vision Transformers}.
\newblock In \emph{ICML}, 2022.

\bibitem[Zhu et~al.(2022)Zhu, Lei, Feng, Wang, Zhang, and Chua]{zhu2022towards}
Fengbin Zhu, Wenqiang Lei, Fuli Feng, Chao Wang, Haozhou Zhang, and Tat-Seng Chua.
\newblock Towards complex document understanding by discrete reasoning.
\newblock In \emph{ACMMM}, 2022.

\bibitem[Zong et~al.(2023)Zong, Song, and Liu]{zong2022codetr}
Zhuofan Zong, Guanglu Song, and Yu Liu.
\newblock {DETRs with Collaborative Hybrid Assignments Training}.
\newblock In \emph{ICCV}, 2023.

\end{thebibliography}
}

\clearpage
\twocolumn[
\centering
\Large
\textbf{RoDLA: Benchmarking the Robustness of Document Layout Analysis Models} \\
\vspace{2.0em} 
\Large \textbf{(Supplementary Material)} \\
\vspace{5.0em}
]
\appendix

\section{Implementation Details}

\noindent \textbf{Hardware Setup.} In this work, we have trained all models (including reproduced models) on machines equipped with four A100, each having $40$ GB of memory. Each node would also with $300$ GB CPU memory.

\noindent \textbf{Training Settings.} After input batch normalization, we have applied flip with a $0.5$ filp ratio and crop with a crop size ($384, 600$) as data augmentation method. For a fair comparison in robustness benchmark for DLA models, we have trained all models (including reproduced models) using the same training strategy as in Table~\ref{tab:training_settings}.
\begin{table}[h]
\centering
\caption{Training settings.}
\vskip -2ex
\setlength{\tabcolsep}{28pt}
\resizebox{\columnwidth}{!}{
\label{tab:training_settings}
\begin{tabular}{l|c}
\toprule[1.5pt]
\textbf{Configurations} & \textbf{Parameter}  \\
\midrule \midrule
Optimizer             & AdamW                 \\
Learning Rate         & $2e^{-4}$ \\ 
Weight Decay          & $1e^{-4}$ \\ 
Scheduler             & step-base                \\
Training Epochs       & 24                    \\
Warm-up Step          & \{16, 22\}                   \\
Warm-up Ratio         & $1e^{-3}$ \\ 
Batch-size per GPU    & 2                    \\ 
\bottomrule[1.5pt]
\end{tabular}
}
\end{table}

To create the benchmark, we have re-trained $38$ models for this robustness benchmark for DLA models: On PubLayNet~\cite{zhong2019publaynet} dataset, we have re-trained $24$ models (including ablation study). On DocLayNet~\cite{pfitzmann2022doclaynet} and M$^6$Doc~\cite{cheng2023m6doc} datasets, we have re-trained $7$ models each, as we have only re-trained the models with representative performance, \ie, high \textbf{mRD} or \textbf{mAP} for specific perturbation, on the robustness benchmark for PubLayNet~\cite{zhong2019publaynet} dataset.

\section{Detail of Perturbation Taxonomy}
In this section, we provide more details about our $12$ document image perturbations in $3$ severity levels. 

\noindent \textbf{(P1) Rotation.} We apply a random rotation to document images, along with corresponding annotations. The rotation operation on an image of a document is an affine transformation, mathematically described by a rotation matrix. If $\theta$ is the angle of rotation, the transformation for rotating a point $(x,y)$ around the origin is given by:
\begin{equation}
\begin{pmatrix}
x' \\
y'
\end{pmatrix}
=
\begin{pmatrix}
\cos(\theta) & -\sin(\theta) \\
\sin(\theta) & \cos(\theta)
\end{pmatrix}
\begin{pmatrix}
x \\
y
\end{pmatrix}.
\end{equation}
Here, $(x', y')$ are the coordinates of the point after rotation. For L1, the $\theta$ is selected randomly from the range $[-5, 5]$. For L2, the $\theta$ is chosen randomly from $[-10, -5]$ or $[5, 10]$, each with 50\% probability. For L3, the $\theta$ is taken randomly from $[-15, -10]$ or $[10, 15]$ for simulating real-world scenarios where object orientations vary.

\noindent \textbf{(P2) Warping.} We apply a pixel-wise displacement defined by a displacement field $D$. This field is typically generated using the Gaussian smoothing of random noise to simulate elastic deformation on document paper. The warping operation is as follows:
\begin{equation}
\begin{aligned}
       D(x, y) = \alpha \cdot G_{\sigma}(R(x, y)), 
\end{aligned}
\end{equation}
\begin{equation}
\begin{aligned}
     \left\{\begin{matrix}
    x' = x + D_x(x, y) \\ y' = y + D_y(x, y), 
    \end{matrix}\right.
\end{aligned}
\end{equation}
where $R(x,y)$ is a random field for displacement in both the $x$ and $y$ directions. $G_\sigma$ is a Gaussian function with standard deviation $\sigma$; the intensity or amplitude of the displacement is controlled by a factor $\alpha$. $D_x$ and $D_y$ are the $x$ and $y$ components of the displacement field $D$.

\noindent \textbf{(P3) Keystoning.} We apply a 3D transformation to a 2D plane through a $3 \times 3$ matrix $H$, preserving lines but not necessarily the actual angles or lengths. This operation maps the homogeneous coordinates of a point in the source image to its new coordinates in the destination image:
\begin{equation}
\begin{pmatrix}
x' \\
y' \\
w'
\end{pmatrix}
= H \cdot
\begin{pmatrix}
x \\
y \\
1
\end{pmatrix}.
\end{equation}
Then the actual position in the transformed image is given by normalizing with $w'$ by $(\frac{x'}{w'}, \frac{y'}{w'})$. The elements of $H$ are typically derived from corner-point correspondences between the source and destination images. The coordinates of destination images are selected randomly from a Gaussian distribution centered around the original coordinates. The standard deviation of the Gaussian distribution is determined by the level.
 
\noindent \textbf{(P4) Watermark.} The process of adding a watermark involves several steps, primarily dealing with image composition and potential rotation. The watermark image is rotated by a random angle $\theta$ from range $[0^\circ, 360^\circ]$ before the blending process. Then the watermark $W$ is blended onto the original image $O$ using a technique called alpha blending. The resulting pixel value $I$ is calculated as:
\begin{equation}
    I = \alpha_w \cdot W + (1 - \alpha_w) \cdot O ,
\end{equation}
where $\alpha_w$ is the transparency level of the watermark, which allows the original image to show through to varying severity levels.

\noindent \textbf{(P5) Background.} For complex background simulation, we overlay multiple images onto the original image. Before background alpha composition, multiple background images are resized and placed on a copy image $B$ of the original image $A$. The placement is defined by the position $(x_{pos}, y_{pos})$, which is randomly generated. The alpha composition can be described as:
\begin{equation}
    I = \alpha_A \cdot A + (1 - \alpha_A) \cdot \alpha_B \cdot B ,
\end{equation}
where $\alpha_A$ and $\alpha_B$ are the alpha values of the original image and the background image, respectively.

\noindent \textbf{(P6) Illumination.} We introduce non-uniform illumination into document images, simulating effects such as shadows or glare. Mask $M$ is created with random polygons filled with black on a white canvas, which is then blurred using a Gaussian filter. The illumination adjustment can be described mathematically as a pixel-wise multiplication of the image $I$ with mask $M$ :
\begin{equation}
    I'(x, y) = V \cdot I(x, y) \cdot M(x, y) ,
\end{equation}
where $V$ is the illumination scaling factor, determined by the severity levels and type of illumination adjustment, \ie, shadow with $V_s$ and glare with $V_l$. 

\noindent \textbf{(P7) Ink-Bleeding.} We apply an erosion operation for ink-bleeding simulation with an elliptical structuring element. The kernel size $K_e$ determines the extent of erosion, depending on severity levels. The basic mathematical formula for erosion $\ominus$ of an image $A$ by a structuring element $B$ is:
\begin{equation}
    (A \ominus B)(x, y) = \min_{(b_x, b_y) \in B}\{A(x + b_x, y + b_y)\} .
\end{equation}
To improve image quality during erosion, we upscale the image tenfold in both dimensions before applying the erosion. This is followed by downscaling to the original size, ensuring enhanced detail and quality in the final image.

\noindent \textbf{(P8) Ink-Holdout.} To simulate Ink-Holdout, which is the opposite of Ink-Bleeding, we use the dilation operation, the inverse of erosion. The parameters for the dilation process, including the kernel size and the number of iterations, remain the same as those used for the erosion operation to maintain consistency in simulating these opposing ink behaviors. The mathematical formula for dilation $\oplus$ of an document image $A$ by a elliptical structuring element $B$ is:
\begin{equation}
    (A \oplus B)(x, y) = \max_{(b_x, b_y) \in B}\{A(x - b_x, y - b_y)\} .
\end{equation}

\noindent \textbf{(P9) Defocus.} The simulation of defocus blur is inherently complex due to the variability of point spread functions (PSFs) within diverse photographic conditions. Nevertheless, given that document images are frequently captured at close quarters, it is feasible to approximate the PSF with a Gaussian kernel function for simulating defocus blur which demonstrated as follows:
\begin{equation} 
I_{\text{defocus}}(x, y) = (I * G)(x, y),
\end{equation}
with
\begin{equation}
    G(x, y) = \frac{1}{2\pi\sigma^2} e^{-\frac{x^2 + y^2}{2\sigma^2}} \:.
\end{equation}
where parameters of Gaussian kernel $G$ are correspond to the level of severity which calibrated to manipulate the scope and the depth of field of the blur.

\noindent \textbf{(P10) Vibration.} Document vibration is simulated by motion blur. The kernel for motion blur is a matrix with non-zero values along a line. This line simulates the path of motion. The kernel for a horizontal motion blur is: 
\begin{equation}
K = \frac{1}{n}
\begin{bmatrix}
1 & 1 & \cdots & 1 \\
0 & 0 & \cdots & 0 \\
\vdots & \vdots & \ddots & \vdots \\
0 & 0 & \cdots & 0
\end{bmatrix},
\end{equation}
where $n$ is the number of non-zero elements in the kernel. This kernel $K$ is then rotated through a random angle $\theta$ within a predetermined range, which emulates the directional motion effect. Similar to defocus blur, the motion blur effect is applied using a convolution operation between the image and the rotated kernel:
\begin{equation}
    B(x, y) = (I * K_{\text{rotated}})(x, y),
\end{equation}
where $I$ is the original image, $K$ is the motion blur kernel, and $B$ is the blurred image.

\noindent \textbf{(P11) Speckle.} Document speckle is generated by superimposing random light (background) and dark (foreground) blobs onto a document image. We generate random blobs based on density, size, and roughness through randomly placed points and Gaussian smoothing. These foreground and background blobs are combined with the original image $I$ as:
\begin{equation}
I_{\text{modified}} =
\min\left(\max\left(I_{\text{original}}, N_{\text{fg}}\right), 1 - N_{\text{bg}}\right),
\end{equation}
where $N_{\text{fg}}$, $N_{\text{bg}}$ represent the foreground and background blob noise intensity. In the mathematical simulation of speckle and blotch noise on document images, Gaussian distributed blob noise are generated within the image domain, modulated by a blob density factor $D_b$ , which is parametrically governed by designated severity levels.

\noindent \textbf{(P12) Texture.} We have endeavored to replicate the texture interference patterns characteristic of document imagery. This approach aims to emulate texture interference by simulating the complex plant fiber structures historically present in archival documents. We have modeled the random walk of fiber paths as follows:
\begin{equation}
    \text{FiberPath} = \left[ \sum_{k=1}^{n} \cos(\theta_k) \cdot \delta, \sum_{k=1}^{n} \sin(\theta_k) \cdot \delta \right],
\end{equation}
where $\theta_k$ are angles drawn from a Cauchy distribution, $\delta$ is the step length, and $k$ is the step number. The final fibrous image is obtained by blending fibrous textures:
\begin{equation}
    I' = (\text{M} \cdot I_\text{ink}) + ((1 - \text{M}) \cdot (1 - I_\text{paper})) \times 255 ,
\end{equation}
where mask $M$ determines the application of ink and paper textures to the original image.
Within this simulation, the spatial distribution of the fibers predominantly conforms to a Gaussian distribution, thereby reflecting the randomness inherent in the physical composition of paper. To impart authenticity to the fiber noise and facilitate a more accurate representation of document wear and quality, we have modulated the fiber density across varying noise levels.

\section{Evaluation Metrics}
In this work, there are two types of evaluation metrics, including: (1) the metrics for quantifying the impacts of perturbations will be presented in Sec.~\ref{sec:metric_mPE}, such as MS-SSIM, CW-SSIM, and our proposed mPE. (2) the metrics for assessing the robustness of models will be detailed in Sec.~\ref{sec:metric_mRD}, such as mAP and our proposed mRD. 

\subsection{Details of Perturbation Evaluation Metrics}\label{sec:metric_mPE}

\begin{figure*}[t]
 \footnotesize	
 \centering
 \subfloat{\frame{\includegraphics[width=0.6\columnwidth]{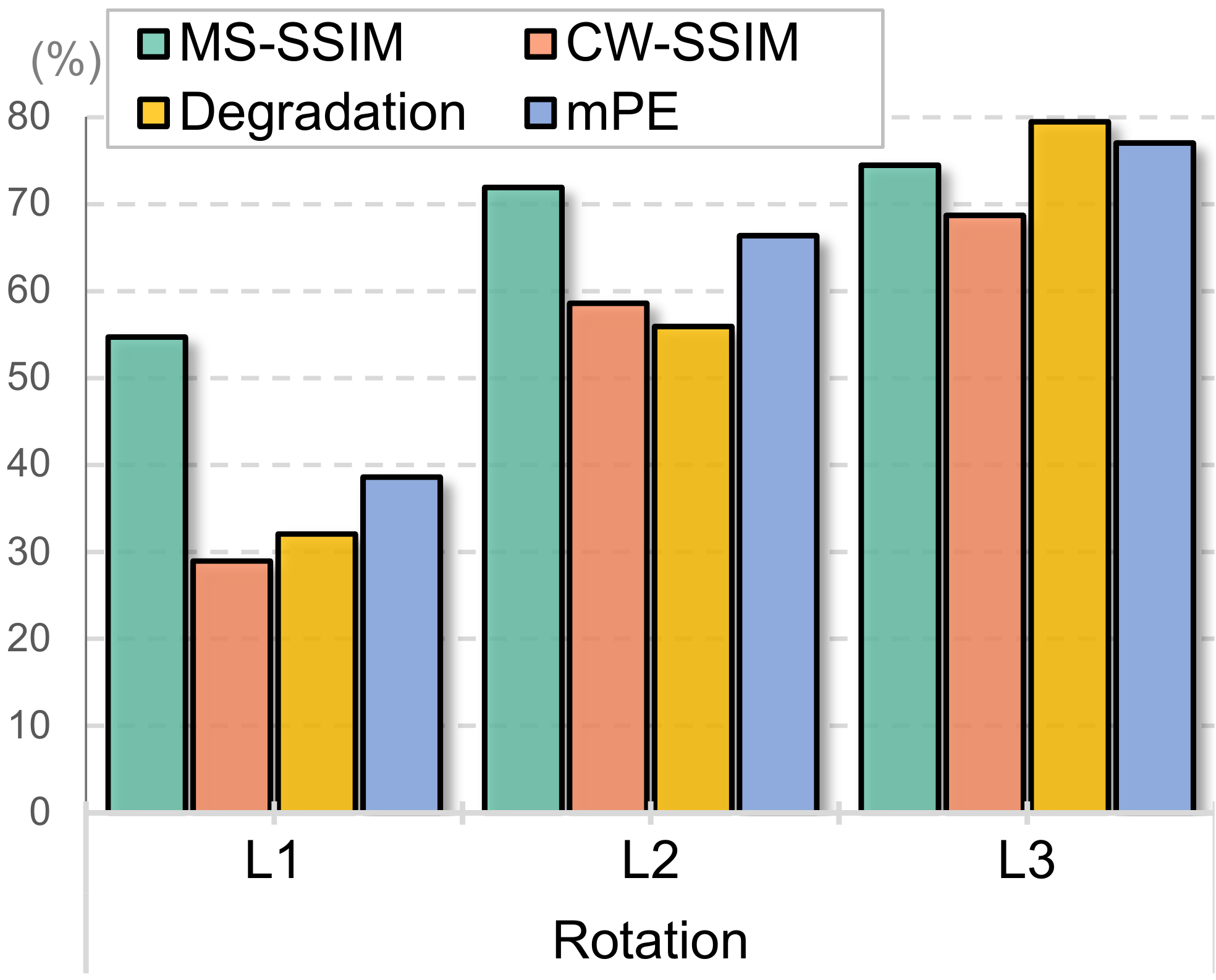}}}\hspace{18pt}
 \subfloat{\frame{\includegraphics[width=0.6\columnwidth]{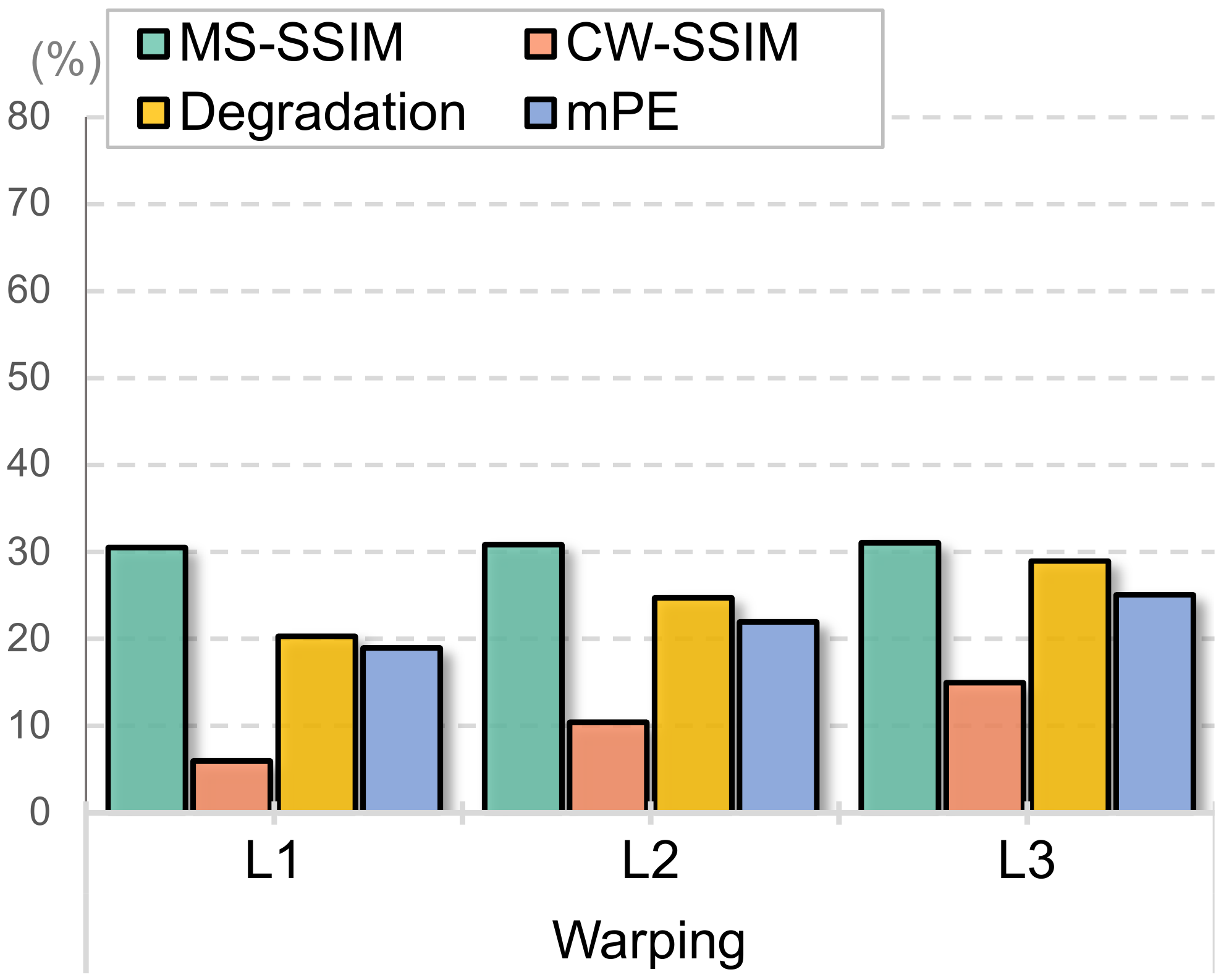}}}\hspace{18pt}
 \subfloat{\frame{\includegraphics[width=0.6\columnwidth]{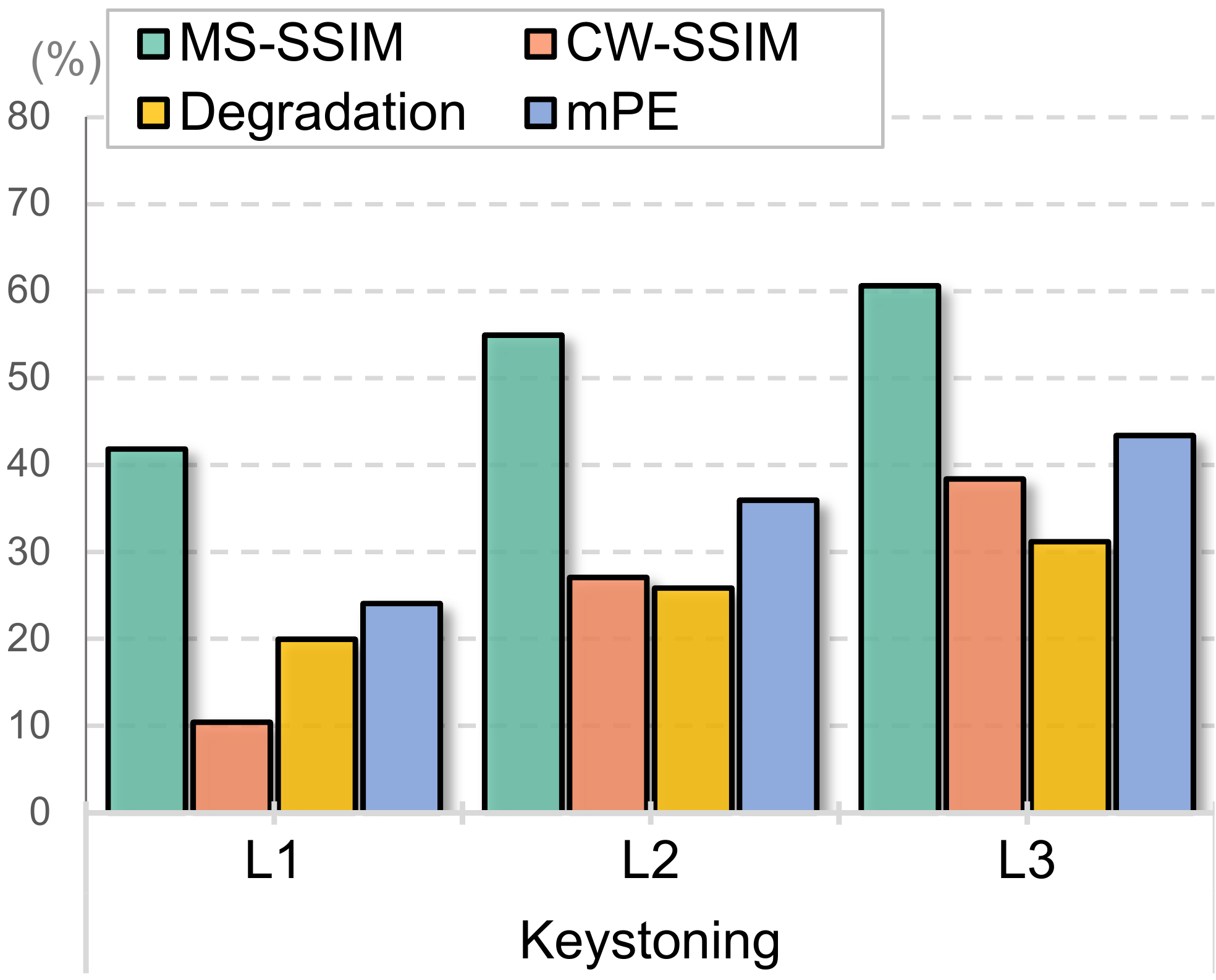}}}\hspace{18pt}
 \\ \vspace{14pt}
 \subfloat{\frame{\includegraphics[width=0.6\columnwidth]{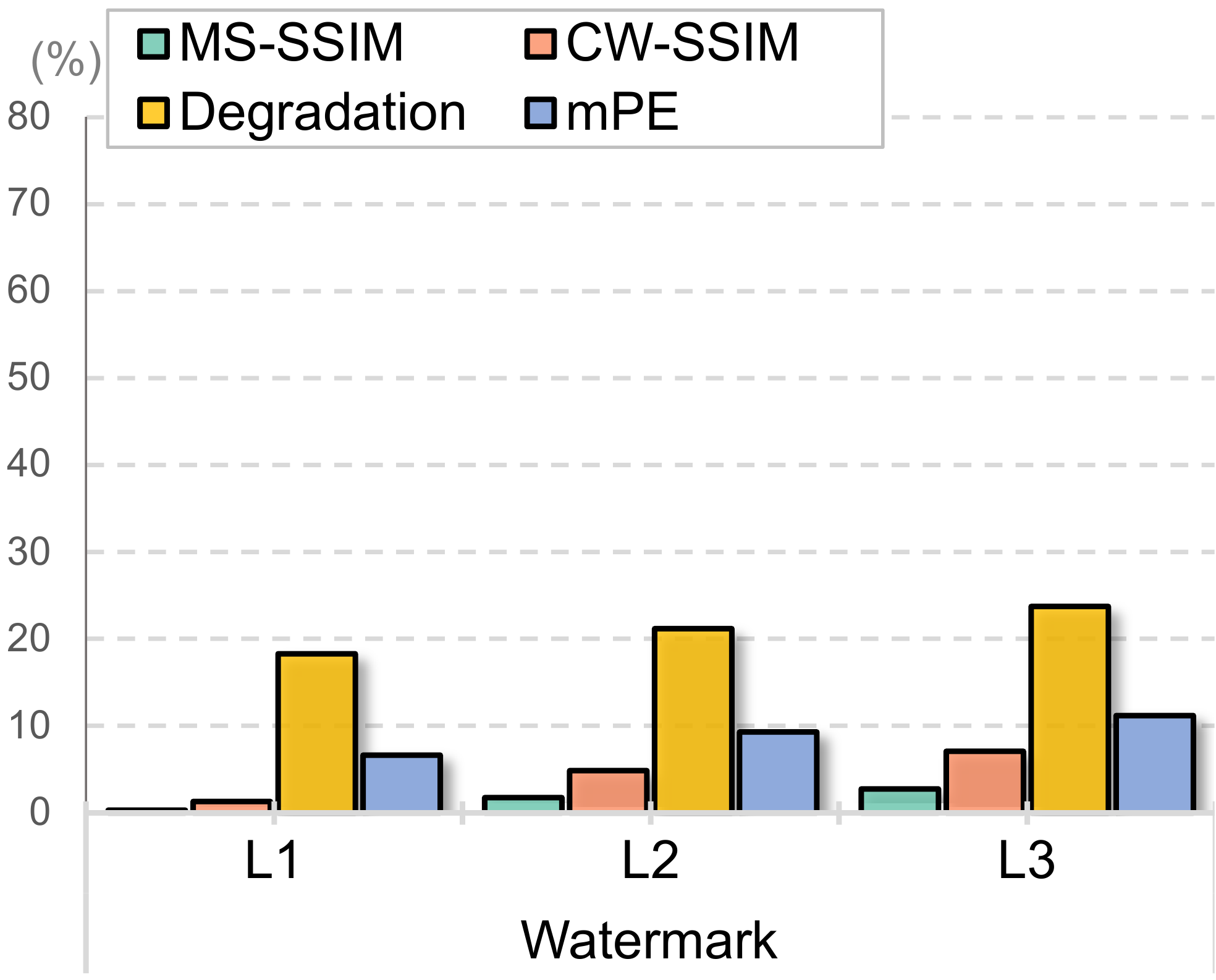}}}\hspace{18pt}
 \subfloat{\frame{\includegraphics[width=0.6\columnwidth]{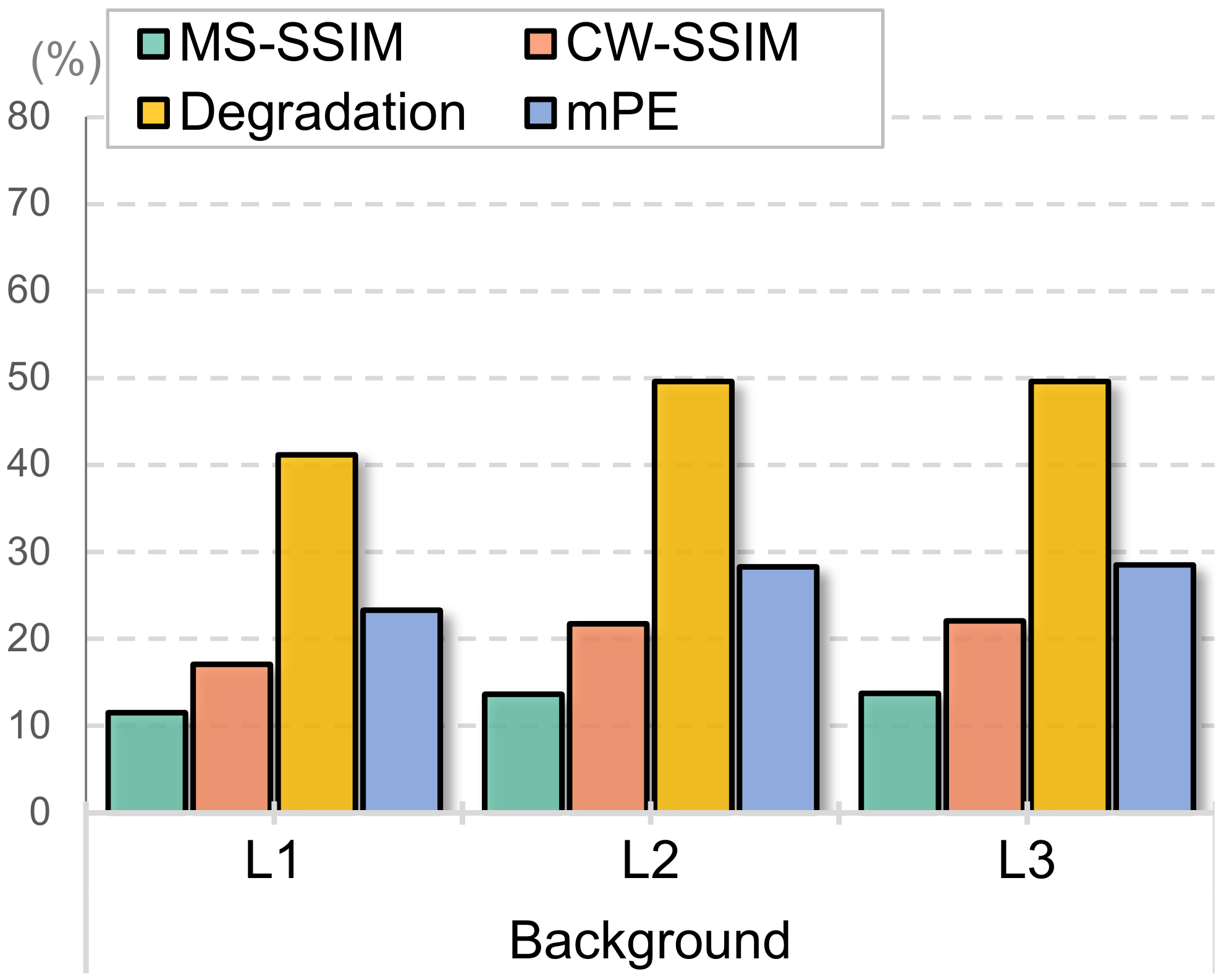}}}\hspace{18pt}
 \subfloat{\frame{\includegraphics[width=0.6\columnwidth]{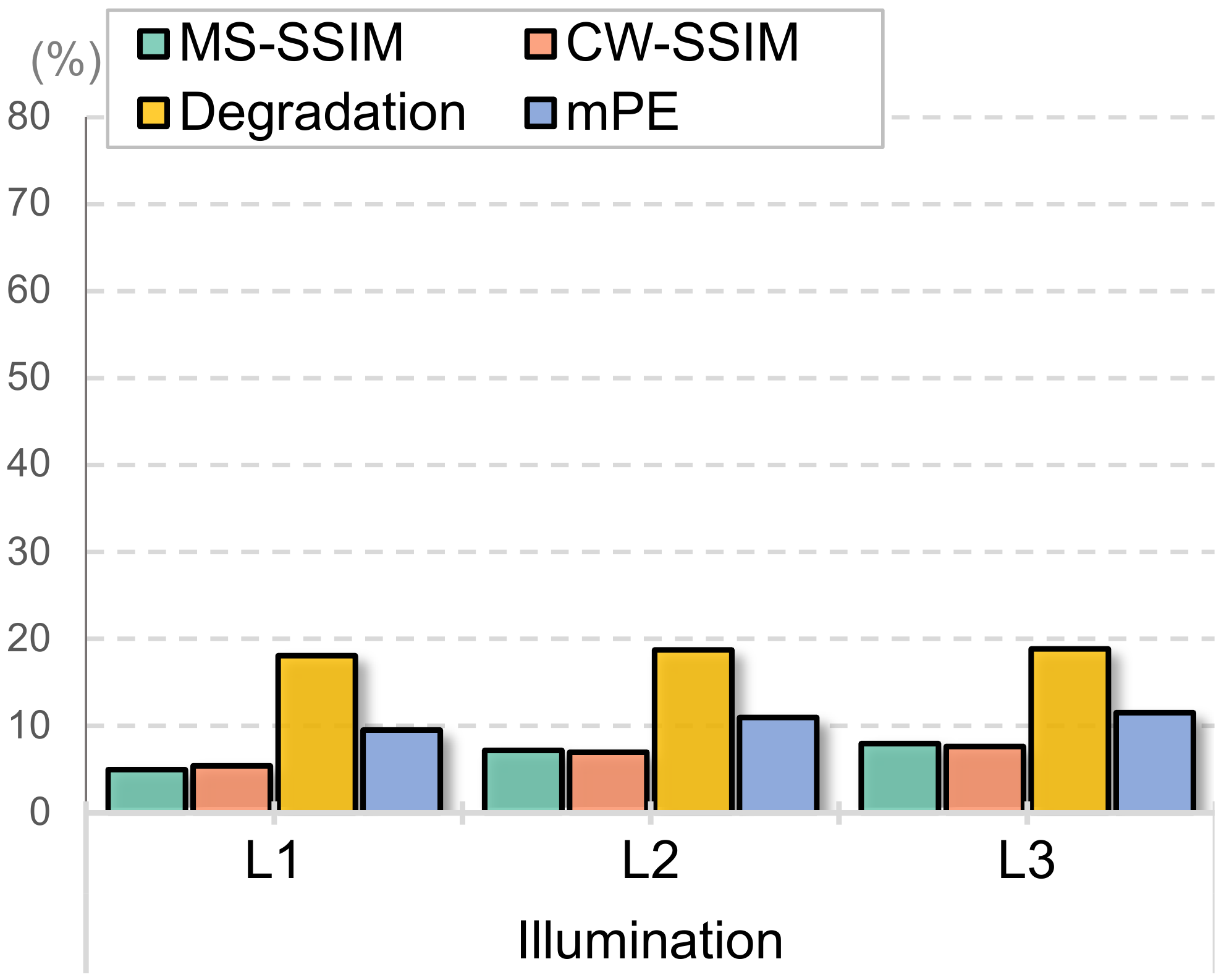}}}\hspace{18pt}
  \\ \vspace{14pt}
 \subfloat{\frame{\includegraphics[width=0.6\columnwidth]{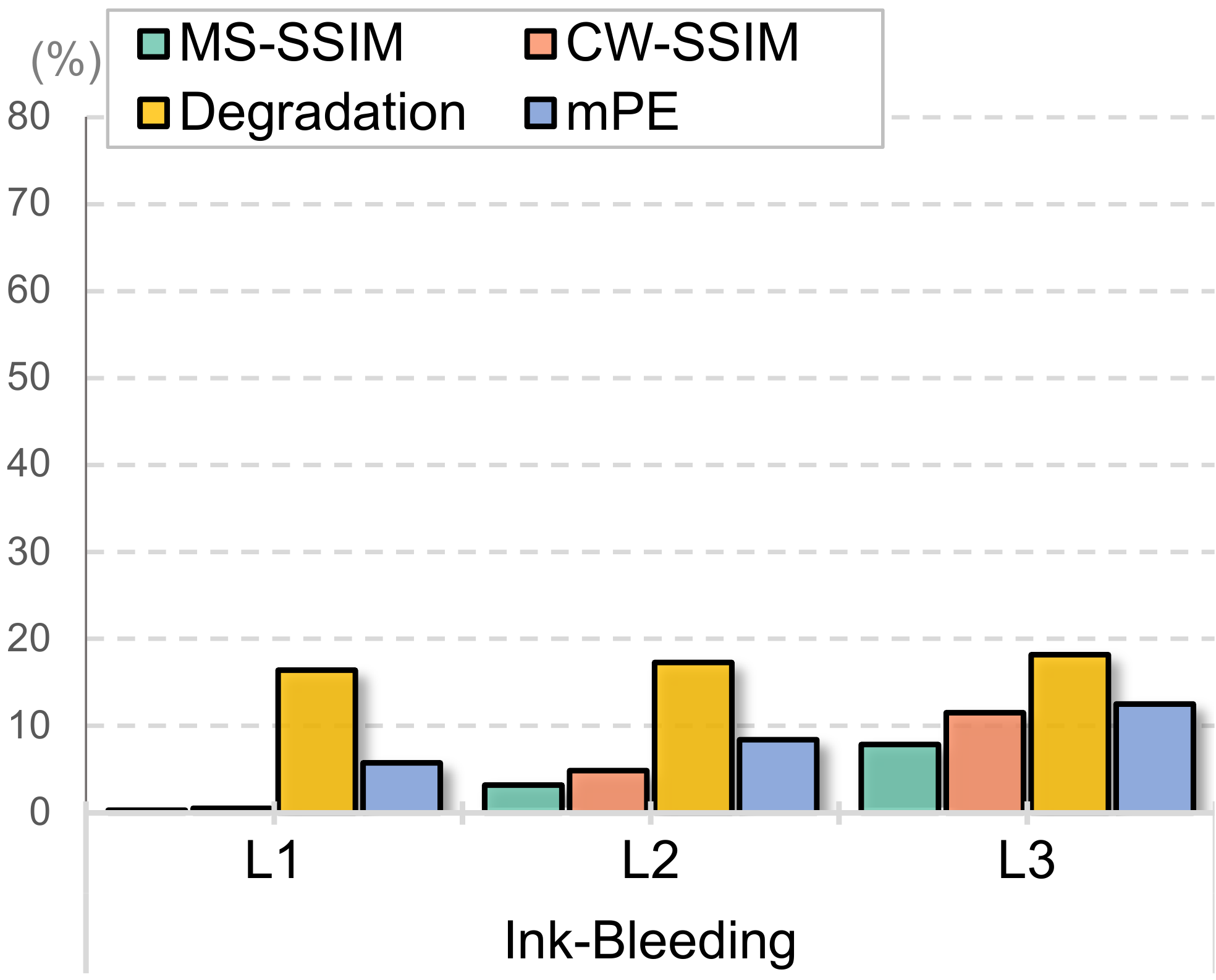}}}\hspace{18pt}
 \subfloat{\frame{\includegraphics[width=0.6\columnwidth]{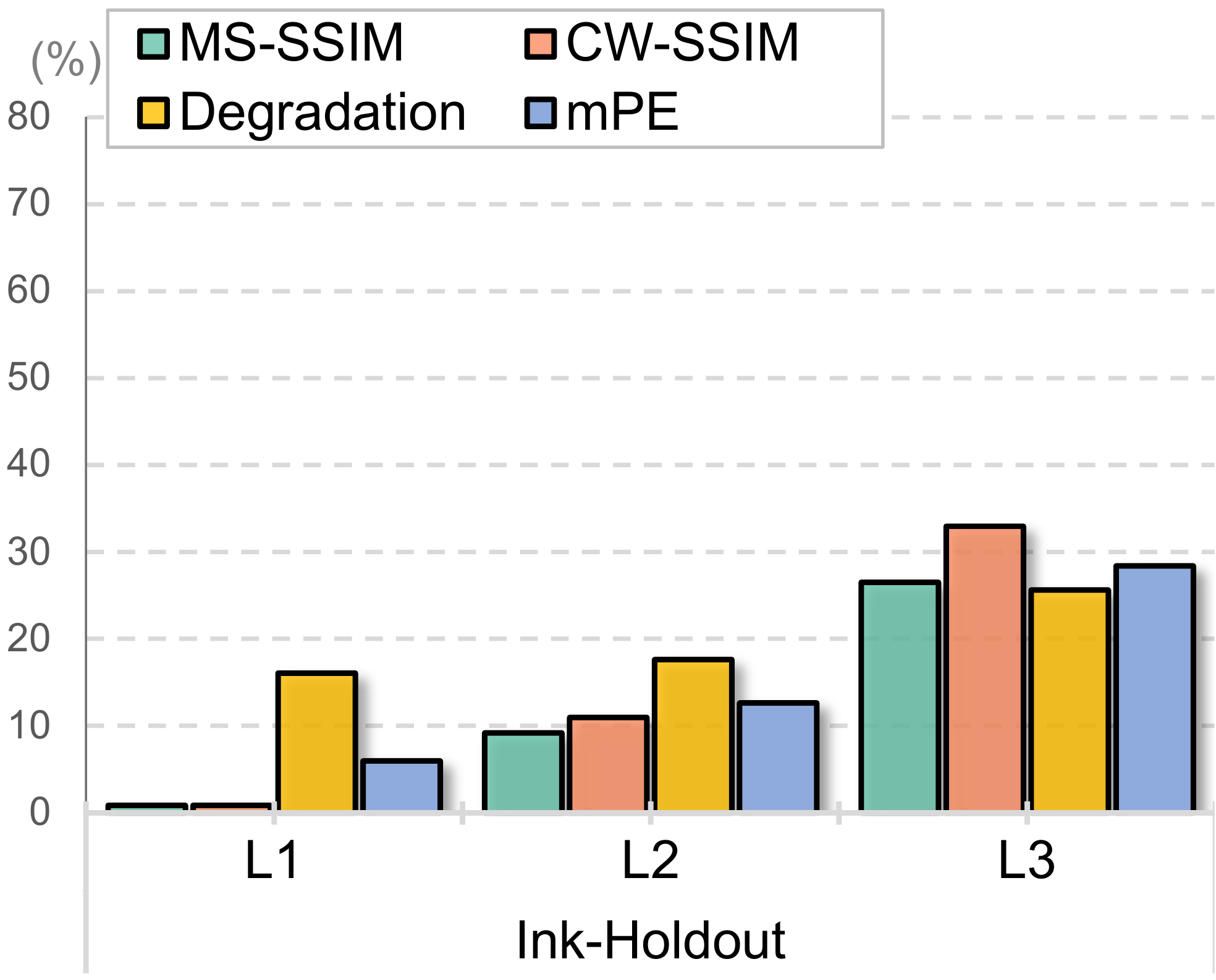}}}\hspace{18pt}
 \subfloat{\frame{\includegraphics[width=0.6\columnwidth]{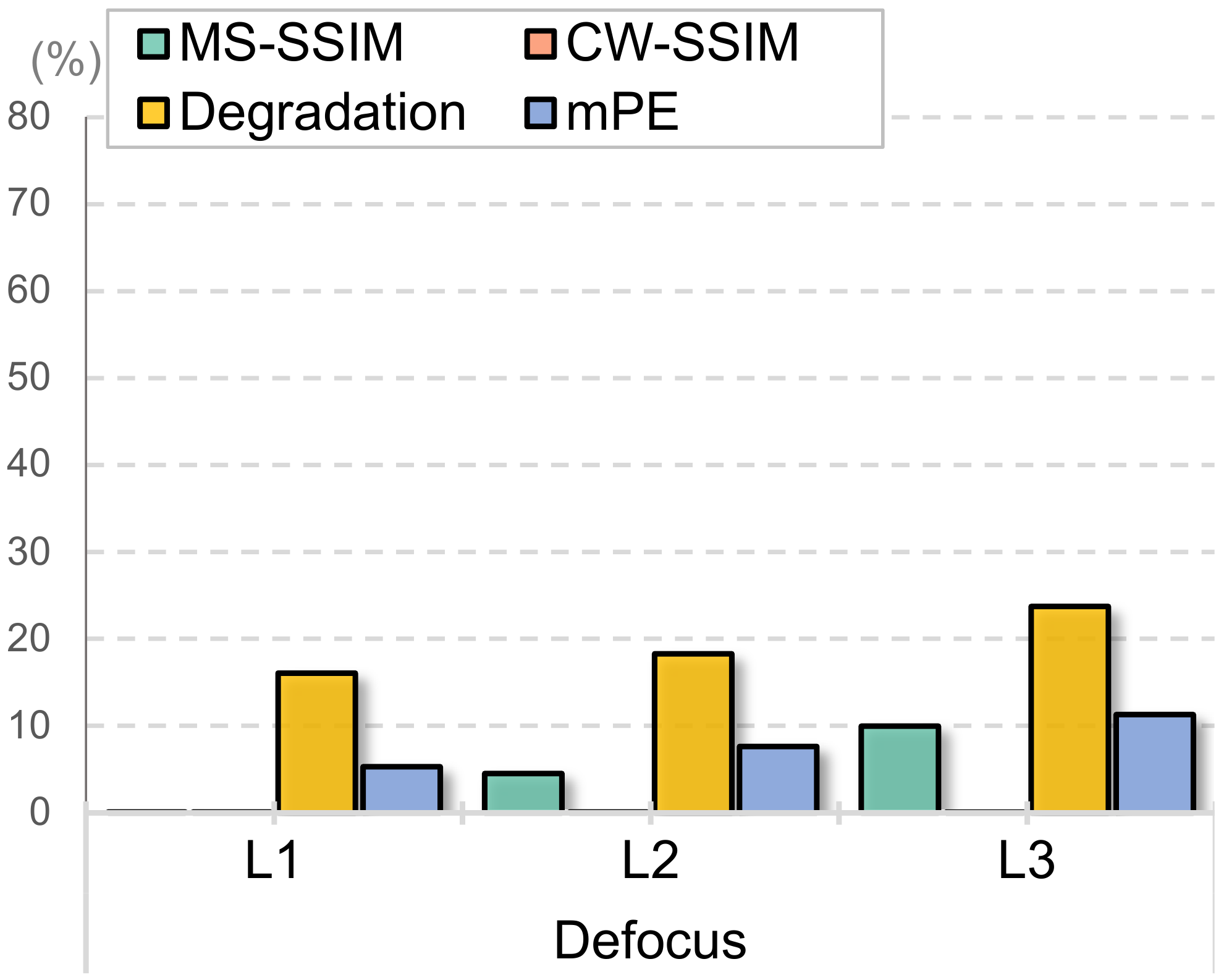}}}\hspace{18pt}
 \\ \vspace{14pt}
 \subfloat{\frame{\includegraphics[width=0.6\columnwidth]{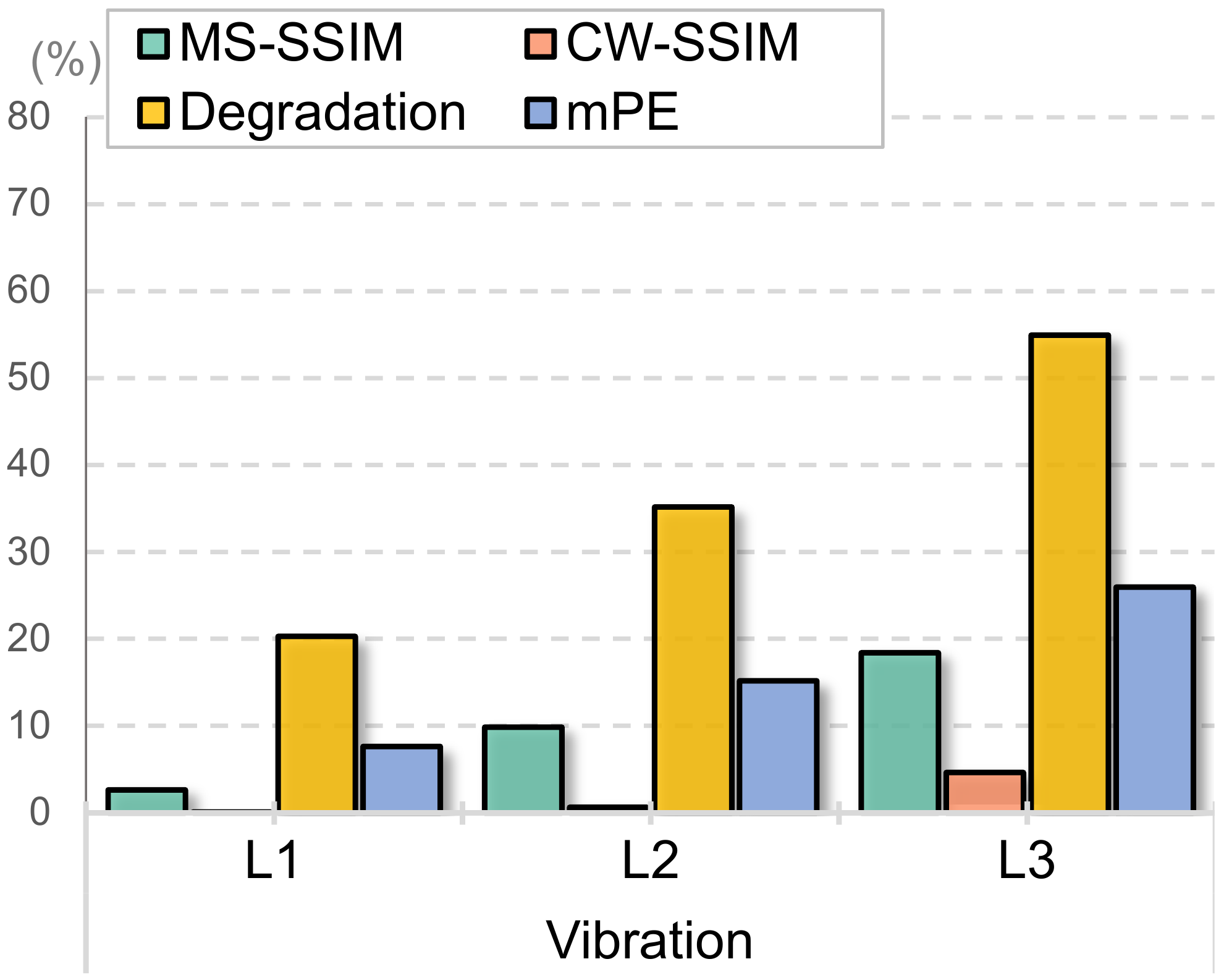}}}\hspace{18pt}
 \subfloat{\frame{\includegraphics[width=0.6\columnwidth]{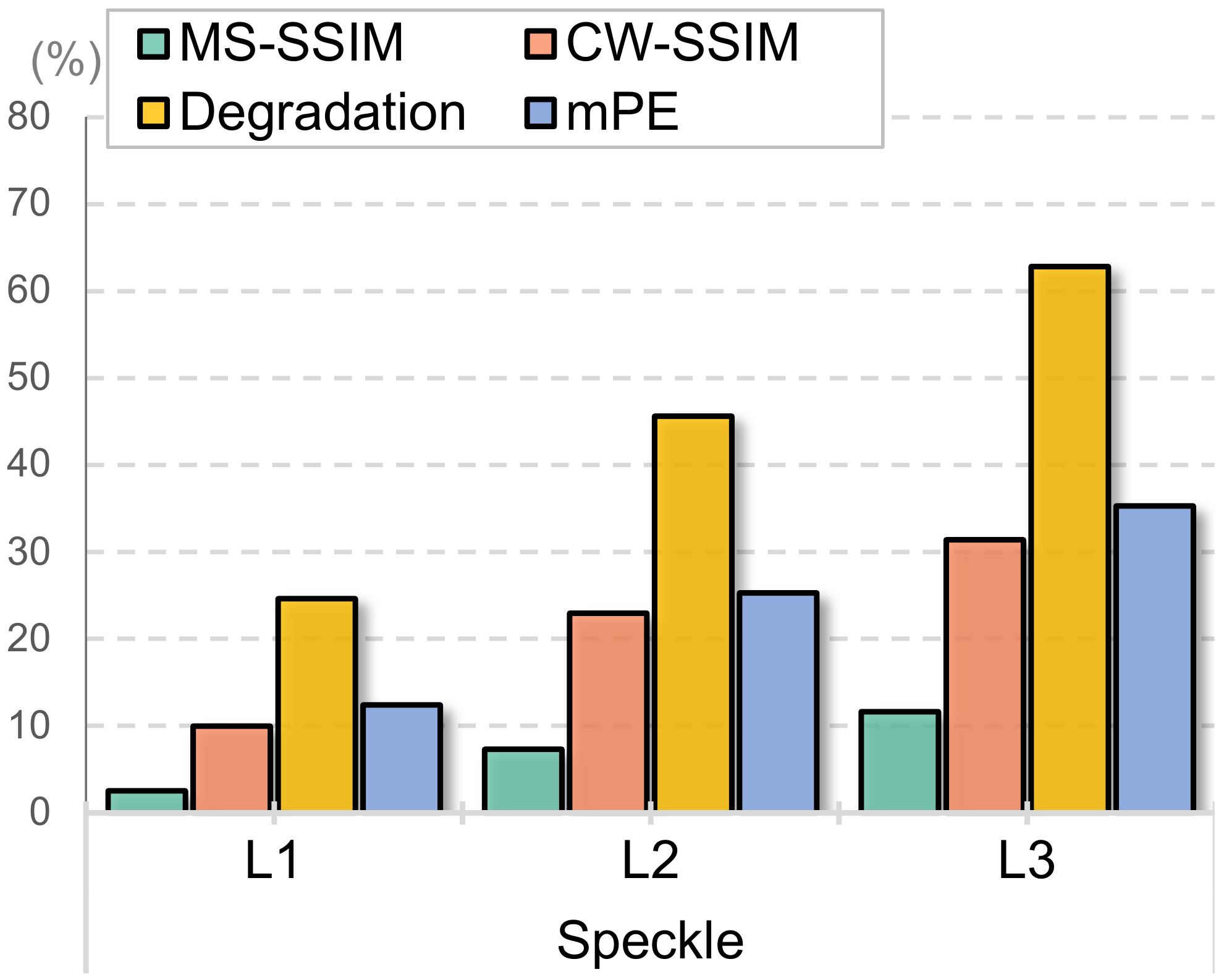}}}\hspace{18pt}
 \subfloat{\frame{\includegraphics[width=0.6\columnwidth]{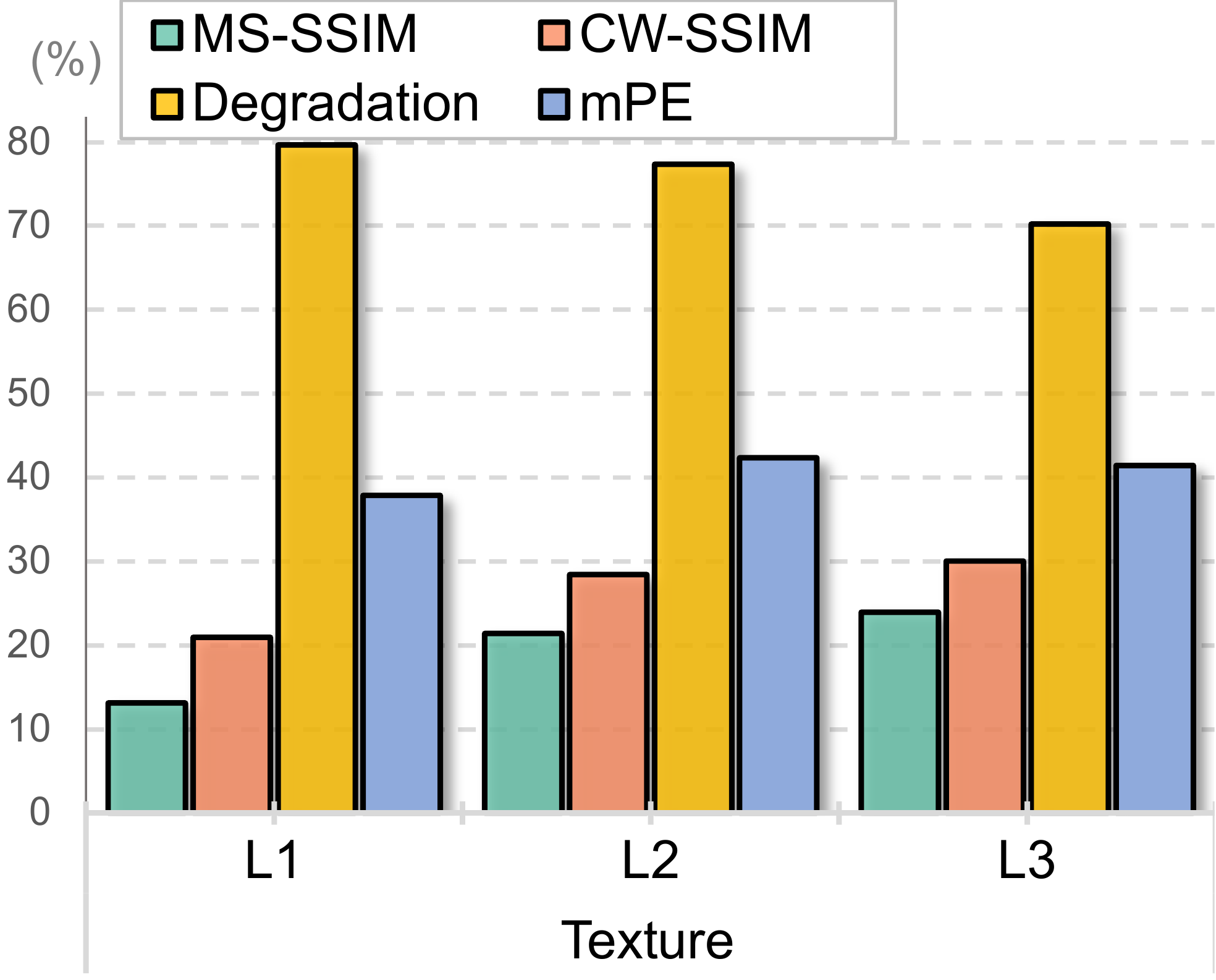}}}
 \caption{\textbf{Comparison between perturbation evaluation metrics} on 12 perturbation categories and 3 severity levels, including {Image Quality Assessment} methods (MS-SSIM and CW-SSIM), {Degradation} \textit{w.r.t} a baseline, and the proposed {mean Perturbation Effect} (mPE). Other metrics cannot assess specific perturbations, for example MS-SSIM is insensitive to \textit{warping} perturbation, and Degradation inversely measures \textit{texture} perturbation across levels. In contrast, mPE is more balanced and inclusive to all perturbations and severity levels.}
\label{fig:iqa_levels}
\vskip -3ex
\end{figure*}

To elucidate the effects of different perturbations and compare perturbation evaluation metrics, we present detailed analyses in Fig.~\ref{fig:iqa_levels}, showcasing the impact of various perturbation categories and levels on document images.
\renewcommand\thesubfigure{P\arabic{subfigure}}
\renewcommand\thesubfigure{\alph{subfigure}}

\noindent \textbf{MS-SSIM \& CW-SSIM.} In our robustness benchmark, we utilize MS-SSIM (Multi-Scale Structural Similarity Index) and CW-SSIM (Complex Wavelet Structural Similarity Index) metrics, both widely recognized for assessing the similarity between two images and pertinent for evaluating the extent of information loss caused by such perturbations. These indices exhibit varying sensitivity to image perturbations, as in Fig.~\ref{fig:iqa_levels}. However, in this study, we deviate from the conventional usage of MS-SSIM and CW-SSIM as mere similarity measures. Given that these metrics yield a value of 100 for identical images, we propose using their complements relative to 100 to represent the loss of information, \ie, $100{-}f^{\text{MS-SSIM}}$ and $100{-}f^{\text{CW-SSIM}}$. This approach enables a nuanced assessment of the impact of perturbations on document images, thereby enhancing the evaluation of model robustness in handling document perturbations.

\noindent \textbf{mPE.}
The Mean Perturbation Effect (mPE) metric integrates the effects of image quality degradation and model performance reduction under various perturbations in DLA. Our mPE metric reveals a consistent trend, with an escalation in values corresponding to increased severity, particularly evident in Keystoning and Texture perturbations, as shown in Fig.~\ref{fig:iqa_levels}. It highlights the compounded effects of perturbations, underscoring the importance of robustness in document analysis models. While all metrics show heightened impact with more severe perturbations, mPE uniquely captures the overall impact, serving as a dependable measure of model robustness against document perturbations and offering a comprehensive view of model robustness.

\subsection{Details of Robustness Evaluation Metrics}\label{sec:metric_mRD}

\noindent \textbf{mAP.} The mean Average Precision (mAP) is a crucial metric in object detection, assessing a model's performance across various classes. It is calculated as the mean of the Average Precision (AP) for each category, where AP is the area under the Precision-Recall curve.
\begin{equation}\label{eq:map}
    \text{mAP} = \frac{1}{N} \sum_{i=1}^{N} AP_i \:.
\end{equation}
Here, $AP_i$ is the Average Precision for the $i^{th}$ class. mAP is especially important in multi-class detection tasks with varying Intersection over Union (IoU) thresholds.

\noindent \textbf{P-Avg.} We introduce P-Avg (Perturbation Average), a novel metric based on the mAP framework, designed to evaluate a model's robustness in document layout recognition across various levels and types of perturbations. P-Avg extends mAP to quantify a model's ability to maintain recognition accuracy under diverse perturbation scenarios. Based on Eq.~(\ref{eq:map}), the P-Avg can be mathematically expressed as:
\begin{equation}
    \text{P-Avg} = \frac{1}{M N} \sum_{s=1}^{M} \sum_{p=1}^{N} \text{mAP}_{s,p} \:.
\end{equation}
In this formula, $s$ represents perturbation level, $p$ represents perturbation categories, and $\text{mAP}_{s,p}$ is the mAP calculated for the $s^{th}$ level of perturbation in the $p^{th}$ category. This metric provides insights into the model's adaptability and consistency in recognizing document layouts despite the presence of diverse and challenging distortions.

\noindent \textbf{mRD.} The mathematical underpinning of mRD pivots on the interplay between degradation $D$ and the Mean Perturbation Effect (mPE). The metric is designed to normalize the degradation observed for a given perturbation by the perturbation's inherent difficulty as captured by mPE. This normalization is crucial as it accounts for the perturbation's baseline impact on the images, thus offering a relativized robustness measure. The degradation $D$ represents how much a model's performance deviates from its unperturbed state when subjected to a specific perturbation and severity level.

\section{More Results}

\subsection{Detailed Results on PubLayNet-P}
\begin{table*}[t]
\centering
\caption{The robustness benchmark with the \textbf{best-case} result on PubLayNet-P dataset. V, L, and T stand for visual, layout, and textual modality. `Ext.' means using extra pre-training data. mAP scores are evaluated on the \textbf{clean} data. \textbf{Best-case} scores are measured to show the upper-bound robustness performance that the DLA model could achieve at each perturbation in our robustness benchmark.}
\vskip -2ex
\label{tab:sota_publaynet_p_best}
\setlength{\tabcolsep}{3pt}
\renewcommand{\arraystretch}{0.89}
\resizebox{\linewidth}{!}{
\begin{tabu}{l|l|ccc|c|c|llllllllllll|c|c}
\toprule[1.5pt]
\multirow{2}{*}{\textbf{Backbone}}&\multirow{2}{*}{\textbf{Method}} & \multicolumn{3}{c|}{\textbf{Modality}} &\multirow{2}{*}{\textbf{Ext.}} & \multirow{2}{*}{\textbf{Clean}} & \multicolumn{3}{c|}{\textbf{Spatial}} & \multicolumn{2}{c|}{\textbf{Content}} & \multicolumn{3}{c|}{\textbf{Inconsistency}}  & \multicolumn{2}{c|}{\textbf{Blur}}  & \multicolumn{2}{c|}{\textbf{Noise}}  & \multirow{2}{*}{\textbf{P-Avg$\uparrow$}} & \multirow{2}{*}{\textbf{mRD{$\downarrow$}}}\\
& & V & L & T &&&{P1} & {P2} & \multicolumn{1}{c|}{P3} & {P4} & \multicolumn{1}{c|}{P5} & {P6} & {P7} & \multicolumn{1}{c|}{P8} & {P9} & \multicolumn{1}{c|}{P10} & {P11} & {P12} &  \\
\midrule \midrule
ResNeXt~\cite{Saining2016resnext}&LayoutParser ~\cite{shen2021layoutparser}& \cmark & \xmark & \xmark & \xmark & 89.0 & 74.8 & 83.3 & 76.8 & 87.5 & 50.9 & 80.5 & 89.1 & 89.2 & 86.7 & 74.8 & 78.3 & 00.5 & 72.7 & 169.0\\
ResNet~\cite{kaiming2015resnet}&Faster R-CNN
~\cite{ren2017FasterRcnn} & \cmark & \xmark & \xmark & \xmark & 90.2 & 67.9 & 79.7 & 80.0 & 81.7 & 58.8 & 81.9 & 83.6 & 83.9 & 83.9 & 79.7 & 75.4 & 29.8 & 73.9  &151.4\\
ResNet~\cite{kaiming2015resnet}&DocSegTr~\cite{biswas2022docsegtr}& \cmark & \xmark & \xmark & \xmark & 90.4 & 72.0 & 84.1 & 79.1 & 87.1 & 49.5 & 72.7 & 88.3 & 88.2 & 89.2 & 55.0 & 81.3 & 00.9 & 70.6 & 173.7\\
ResNet~\cite{kaiming2015resnet}&Mask R-CNN 
~\cite{He_2017_MaskRcnn} & \cmark & \xmark & \xmark & \xmark & 91.0 & 65.2 & 77.9 & 78.0 & 79.4 & 54.7 & 79.4 & 81.3 & 81.6 & 81.4 & 77.0 & 72.1 & 36.4 & 72.0& 165.0\\
Swin~\cite{liu2021swin}&SwinDocSegmenter~\cite{banerjee2023swindocsegmenter}&\cmark & \xmark & \xmark & \xmark  & 93.7 & 75.6 & 88.4 & 66.3 & 90.2 & 49.5 & 88.3 & 92.6 & 92.9 & 93.3 & 81.6 & 68.5 & 01.1 & 74.0& 139.8\\
\rowfont{\color{gray!30}} DiT~\cite{li2022dit}&Cascade R-CNN~\cite{cai2018cascade} & \cmark & \xmark & \xmark & \cmark &94.5 & 70.7 & 90.0 & 90.1 & 93.2 & 80.1 & 88.1 & 94.2 & 94.4 & 94.8 & 94.6 & 85.8 & 43.4 & 85.0& 80.0\\
\rowfont{\color{gray!30}} LayoutLMv3~\cite{huang2022layoutlmv3}&Cascade R-CNN~\cite{cai2018cascade} & \cmark & \cmark & \cmark & \cmark & 95.1 & 70.7 & 89.2 & 88.1 & 93.6 & 75.1 & 90.1 & 94.6 & 94.2 & 95.1 & 92.2 & 89.9 & 57.9 & 85.9& 81.4\\
\midrule
Swin~\cite{liu2021swin}& Co-DINO~\cite{zong2022codetr} & \cmark & \xmark & \xmark & \xmark  & 94.3 &  57.0 & 71.5 & 29.0 & 92.9 & 61.3 & 87.2 & 92.5 & 92.1 & 93.6 & 54.5 & 71.8 & 22.0 & 68.8&153.5\\
InternImage~\cite{wang2023internimage}& Cascade R-CNN~\cite{cai2018cascade} & \cmark & \xmark & \xmark & \xmark  & 94.1 &71.5 & 88.8 & 88.6 & 91.8 & 59.1 & 84.7 & 93.4 & 92.8 & 93.6 & 91.6 & 81.2 & 00.5 & 78.1 &110.3\\
InternImage~\cite{wang2023internimage}& DINO ~\cite{zhang2022dino}& \cmark & \xmark & \xmark & \xmark  & 95.4&74.6 & 89.5 & 89.2 & 93.9 & 59.8 & 92.0 & 94.4 & 94.2 & 95.1 & 93.5 & 80.4 & 01.6 & 79.9& 94.4 \\
InternImage~\cite{wang2023internimage}& Co-DINO~\cite{zong2022codetr} & \cmark & \xmark & \xmark & \xmark  & 94.2 & 49.8 & 73.0 & 43.3 & 92.5 & 63.6 & 88.5 & 91.9 & 92.2 & 93.5 & 72.4 & 75.7 & 13.8 & 70.9& 153.5\\
\rowcolor[gray]{.9} InternImage~\cite{wang2023internimage}&RoDLA (Ours) & \cmark & \xmark & \xmark & \xmark & 96.0 & 71.9 & 89.0 & 88.4 & 94.2 & 64.4 & 92.0 & 94.6 & 94.6 & 95.7 & 93.3 & 81.6 & 00.5 & 80.0 & 93.8\\

\bottomrule[1.5pt]

\end{tabu}
}
\end{table*}
\begin{table*}[t]
\centering
\caption{The robustness benchmark with the \textbf{worst-case} result on PubLayNet-P dataset. \textbf{Worst-case} scores are measured to show the lower-bound robustness performance that the DLA model could achieve at each perturbation in our robustness benchmark. 
}
\vskip -2ex
\label{tab:sota_publaynet_p_worst}
\renewcommand{\arraystretch}{0.89}
\setlength{\tabcolsep}{3pt}
\resizebox{\linewidth}{!}{
\begin{tabu}{l|l|ccc|c|c|llllllllllll|c|c}
\toprule[1.5pt]
\multirow{2}{*}{\textbf{Backbone}}&\multirow{2}{*}{\textbf{Method}} & \multicolumn{3}{c|}{\textbf{Modality}} &\multirow{2}{*}{\textbf{Ext.}} & \multirow{2}{*}{\textbf{Clean}} & \multicolumn{3}{c|}{\textbf{Spatial}} & \multicolumn{2}{c|}{\textbf{Content}} & \multicolumn{3}{c|}{\textbf{Inconsistency}}  & \multicolumn{2}{c|}{\textbf{Blur}}  & \multicolumn{2}{c|}{\textbf{Noise}}  & \multirow{2}{*}{\textbf{P-Avg$\uparrow$}} & \multirow{2}{*}{\textbf{mRD{$\downarrow$}}}\\
& & V & L & T &&&{P1} & {P2} & \multicolumn{1}{c|}{P3} & {P4} & \multicolumn{1}{c|}{P5} & {P6} & {P7} & \multicolumn{1}{c|}{P8} & {P9} & \multicolumn{1}{c|}{P10} & {P11} & {P12} &  \\
\midrule \midrule
ResNeXt~\cite{Saining2016resnext}&LayoutParser ~\cite{shen2021layoutparser}& \cmark & \xmark & \xmark & \xmark & 89.0 & 03.7 & 73.5 & 59.2 & 81.9 & 42.9 & 78.1 & 81.2 & 71.8 & 15.5 & 03.6 & 28.9 & 00.1 & 45.0& 256.4\\
ResNet~\cite{kaiming2015resnet}&Faster R-CNN
~\cite{ren2017FasterRcnn} & \cmark & \xmark & \xmark & \xmark & 90.2 & 20.6 & 71.0 & 68.8 & 76.3 & 50.4 & 81.1 & 81.8 & 74.4 & 76.3 & 45.1 & 37.2 & 20.3 & 58.6& 206.3\\
ResNet~\cite{kaiming2015resnet}&DocSegTr~\cite{biswas2022docsegtr}& \cmark & \xmark & \xmark & \xmark & 90.4 & 00.7 & 68.1 & 61.5 & 82.9 & 42.8 & 67.2 & 83.5 & 34.5 & 12.6 & 1.5 & 39.1 & 00.1 & 41.2 & 278.3\\
ResNet~\cite{kaiming2015resnet}&Mask R-CNN 
~\cite{He_2017_MaskRcnn} & \cmark & \xmark & \xmark & \xmark & 91.0 & 15.4 & 70.4 & 65.8 & 74.5 & 45.6 & 78.8 & 79.9 & 70.2 & 73.9 & 30.1 & 39.5 & 28.2 & 56.0& 228.2\\
Swin~\cite{liu2021swin}&SwinDocSegmenter~\cite{banerjee2023swindocsegmenter}&\cmark & \xmark & \xmark & \xmark  & 93.7 & 03.6 & 78.1 & 55.3 & 87.7 & 42.8 & 87.7 & 82.3 & 76.2 & 02.3 & 02.0 & 12.1 & 00.2 & 44.2& 281.2\\
\rowfont{\color{gray!30}} DiT~\cite{li2022dit}&Cascade R-CNN~\cite{cai2018cascade} & \cmark & \xmark & \xmark & \cmark &94.5 & 01.5 & 83.7 & 73.9 & 91.0 & 78.7 & 86.4 & 89.7 & 87.2 & 91.9 & 33.6 & 55.8 & 41.0 & 67.9 & 116.0\\
\rowfont{\color{gray!30}} LayoutLMv3~\cite{huang2022layoutlmv3}&Cascade R-CNN~\cite{cai2018cascade} & \cmark & \cmark & \cmark & \cmark & 95.1 & 03.1 & 83.2 & 71.2 & 91.4 & 65.2 & 84.2 & 91.5 & 74.0 & 63.0 & 06.3 & 74.8 & 36.3 & 62.0& 146.0\\
\midrule
Swin~\cite{liu2021swin}& Co-DINO~\cite{zong2022codetr} & \cmark & \xmark & \xmark & \xmark  & 94.3 &  00.9 & 17.5 & 20.2 & 92.3 & 53.9 & 86.0 & 42.3 & 45.7 & 01.4 & 02.0 & 45.2 & 10.5 & 34.8 &342.8\\
InternImage~\cite{wang2023internimage}& Cascade R-CNN~\cite{cai2018cascade} & \cmark & \xmark & \xmark & \xmark  & 94.1 &00.3 & 70.1 & 69.4 & 87.5 & 53.6 & 83.3 & 89.8 & 84.4 & 69.0 & 12.8 & 31.5 & 00.0 & 54.3&174.6\\
InternImage~\cite{wang2023internimage}& DINO ~\cite{zhang2022dino}& \cmark & \xmark & \xmark & \xmark  & 95.4&02.2 & 74.3 & 74.9 & 90.5 & 54.2 & 91.6 & 91.5 & 88.4 & 83.8 & 25.8 & 22.0 & 00.9 & 58.3& 147.0 \\
InternImage~\cite{wang2023internimage}& Co-DINO~\cite{zong2022codetr} & \cmark & \xmark & \xmark & \xmark  & 94.2 & 00.4 & 26.8 & 28.1 & 90.1 & 58.6 & 87.5 & 66.2 & 80.7 & 02.3 & 03.2 & 30.8 & 09.9 & 40.4 & 301.5\\
\rowcolor[gray]{.9} InternImage~\cite{wang2023internimage}&RoDLA (Ours) & \cmark & \xmark & \xmark & \xmark & 96.0 & 03.2 & 57.8 & 72.3 & 88.0 & 48.2 & 89.6 & 89.6 & 87.3 & 72.5 & 19.8 & 16.2 & 00.2 & 53.7 & 138.7\\

\bottomrule[1.5pt]

\end{tabu}
}
\vskip -1ex
\end{table*}
\begin{table*}[!ht]
\centering
\caption{The detailed per-level \textbf{P-Avg$\uparrow$} results on PubLayNet-P dataset. \textbf{L1}, \textbf{L2}, and \textbf{L3} stand for the severity levels from light to heavy.}
\vskip -2ex
\label{tab:pavg_detail_publaynet}
\setlength{\tabcolsep}{3pt}
\resizebox{\linewidth}{!}{
\begin{tabu}{l|l|ccc|c|c|ccc|ccc|ccc|ccc|ccc|ccc|ccc|ccc|ccc|ccc|ccc|ccc}
\toprule[1.5pt]
\multirow{2}{*}{\textbf{Backbone}}&\multirow{2}{*}{\textbf{Method}} & \multicolumn{3}{c|}{\textbf{Modality}} &\multirow{2}{*}{\textbf{Ext.}} & \multirow{2}{*}{\textbf{Clean}} & \multicolumn{3}{c|}{Rotation} & \multicolumn{3}{c|}{Warping} & \multicolumn{3}{c|}{Keystoning} & \multicolumn{3}{c|}{Watermark} & \multicolumn{3}{c|}{Background} & \multicolumn{3}{c|}{Illumination} & \multicolumn{3}{c|}{Ink-Bleeding} & \multicolumn{3}{c|}{Ink-Holdout} & \multicolumn{3}{c|}{Defocus} & \multicolumn{3}{c|}{Vibration} & \multicolumn{3}{c|}{Speckle} & \multicolumn{3}{c}{Texture} \\
& & V & L & T && & L1 & L2 & L3 & L1 & L2 & L3 & L1 & L2 & L3 & L1 & L2 & L3 & L1 & L2 & L3 & L1 & L2 & L3 & L1 & L2 & L3 & L1 & L2 & L3 & L1 & L2 & L3 & L1 & L2 & L3 & L1 & L2 & L3 & L1 & L2 & L3 \\
\midrule \midrule
ResNeXt~\cite{Saining2016resnext}&LayoutParser ~\cite{shen2021layoutparser}& \cmark & \xmark & \xmark & \xmark & 89.0 & 74.8 & 29.0 & 03.7 & 83.3 & 77.6 & 73.5 & 76.8 & 68.0 & 59.2 & 87.5 & 85.0 & 81.9 & 50.9 & 42.9 & 44.0 & 80.5 & 78.4 & 78.1 & 89.1 & 87.2 & 81.2 & 89.2 & 86.9 & 71.8 & 86.7 & 51.9 & 15.5 & 74.8 & 14.9 & 3.6 & 78.3 & 53.5 & 28.9 & 00.5 & 00.2 & 00.1\\
ResNet~\cite{kaiming2015resnet}&Faster R-CNN
~\cite{ren2017FasterRcnn} & \cmark & \xmark & \xmark & \xmark & 90.2& 67.9&44.1&20.6&79.7&75.2&71.0&80.0&74.1&68.8&81.7&78.8&76.3&58.8&50.4&50.4&81.9&81.3&81.1&83.6&82.7&81.8&83.9&82.4&74.4&83.9&81.7&76.3&79.7&64.8&45.1&75.4&54.4&37.2&20.3&22.7&29.8
\\
ResNet~\cite{kaiming2015resnet}&DocSegTr~\cite{biswas2022docsegtr}& \cmark & \xmark & \xmark & \xmark & 90.4 & 72.0&12.2&00.7&84.1&75.7&68.1&79.1&72.9&61.5&87.1&85.5&82.9&49.5&42.8&46.2&72.7&67.9&67.2&88.3&86.7&83.5&88.2&83.9&34.5&89.2&50.6&12.6&55.0&02.9&01.5&81.3&60.3&39.1&00.9&00.1&00.1
\\
ResNet~\cite{kaiming2015resnet}&Mask R-CNN 
~\cite{He_2017_MaskRcnn} & \cmark & \xmark & \xmark & \xmark & 91.0 & 65.2&39.3&15.4&77.9&73.8&70.4&78.0&71.5&65.8&79.4&76.9&74.5&54.7&45.9&45.6&79.4&78.9&78.8&81.3&80.7&79.9&81.6&80.5&70.2&81.4&79.3&73.9&77.0&55.6&30.1&72.1&53.6&39.5&28.2&31.1&36.4
\\
Swin~\cite{liu2021swin}&SwinDocSegmenter~\cite{banerjee2023swindocsegmenter}&\cmark & \xmark & \xmark & \xmark  & 93.7 & 75.6&37.9&03.6&88.4&83.5&78.1&66.3&62.3&55.3&90.2&88.8&87.7&49.5&42.8&47.4&88.3&87.7&87.7&92.6&89.4&82.3&92.9&90.7&76.2&93.3&16.4&02.3&81.6&04.1&02.0&68.5&27.8&12.1&01.1&00.2&00.2
\\
\rowfont{\color{gray!30}} DiT~\cite{li2022dit}&Cascade R-CNN~\cite{cai2018cascade} & \cmark & \xmark & \xmark & \cmark &94.5 &70.7&23.5&01.5&90.0&85.9&83.7&90.1&82.5&73.9&93.2&92.0&91.0&80.1&78.7&79.9&88.1&87.0&86.4&94.2&92.2&89.7&94.4&93.1&87.2&94.8&94.7&91.9&94.6&85.7&33.6&85.8&68.2&55.8&43.4&41.0&41.0
\\
\rowfont{\color{gray!30}} LayoutLMv3~\cite{huang2022layoutlmv3}&Cascade R-CNN~\cite{cai2018cascade} & \cmark & \cmark & \cmark & \cmark & 95.1 &70.7&24.3&03.1&89.2&85.2&83.2&88.1&80.2&71.2&93.6&92.0&91.4&75.1&65.2&65.2&90.1&85.3&84.2&94.6&93.1&91.5&94.2&91.9&74.0&95.1&90.7&63.0&92.2&42.5&06.3&89.9&81.5&74.8&57.9&41.0&36.3
\\
\midrule
Swin~\cite{liu2021swin}& Co-DINO~\cite{zong2022codetr} & \cmark & \xmark & \xmark & \xmark  & 94.3 & 57.0&09.3&00.9&71.5&40.0&17.5&29.0&24.8&20.2&92.9&92.7&92.3&53.9&55.3&61.3&87.2&86.1&86.0&92.5&83.0&42.3&92.1&89.5&45.7&93.6&09.9&01.4&54.5&04.1&02.0&71.8&49.1&45.2&22.0&12.2&10.5
\\
InternImage~\cite{wang2023internimage}& Cascade R-CNN~\cite{cai2018cascade} & \cmark & \xmark & \xmark & \xmark  & 94.1 &71.5&11.2&00.3&88.8&81.7&70.1&88.6&81.2&69.4&91.8&89.5&87.5&59.1&53.6&57.1&84.7&83.3&83.3&93.4&91.5&89.8&92.8&91.5&84.4&93.6&92.2&69.0&91.6&57.6&12.8&81.2&52.8&31.5&00.5&00.1&00.0
\\
InternImage~\cite{wang2023internimage}& DINO ~\cite{zhang2022dino}& \cmark & \xmark & \xmark & \xmark  & 95.4& 74.6&26.5&02.2&89.5&82.7&74.3&89.2&82.6&74.9&93.9&92.4&90.5&59.8&54.2&59.2&92.0&91.6&91.6&94.4&92.4&91.5&94.2&93.6&88.4&95.1&93.7&83.8&93.5&70.9&25.8&80.4&39.5&22.0&01.3&00.9&01.6
\\
InternImage~\cite{wang2023internimage}& Co-DINO~\cite{zong2022codetr} & \cmark & \xmark & \xmark & \xmark  & 94.2 & 49.8&07.2&00.4&73.0&45.0&26.8&43.3&35.8&28.1&92.5&91.3&90.1&59.7&58.6&63.6&88.5&87.8&87.5&91.9&86.0&66.2&92.2&89.9&80.7&93.5&17.8&02.3&72.4&08.0&03.2&75.7&42.6&30.8&13.8&09.9&10.0
\\
\rowcolor[gray]{.9} InternImage~\cite{wang2023internimage}&RoDLA (Ours) & \cmark & \xmark & \xmark & \xmark & 96.0 &71.9&19.9&02.9&89.0&80.4&68.5&88.4&81.2&72.1&94.2&93.0&91.6&64.4&58.4&61.9&92.0&91.4&91.5&94.6&92.4&90.8&94.6&92.9&87.3&95.7&94.4&83.8&93.3&71.2&38.5&81.6&51.2&43.6&00.5&00.2&00.1
\\

\bottomrule[1.5pt]

\end{tabu}
}
\vskip -1ex
\end{table*}
\begin{table*}[!ht]
\centering
\caption{
The detailed per-level \textbf{RD$\downarrow$} results on PubLayNet-P dataset. \textbf{L1}, \textbf{L2}, and \textbf{L3} stand for the severity levels from light to heavy.
}
\vskip -2ex
\label{tab:rd_detail_publaynet}
\setlength{\tabcolsep}{3pt}
\resizebox{\linewidth}{!}{
\begin{tabu}{l|l|ccc|c|ccc|ccc|ccc|ccc|ccc|ccc|ccc|ccc|ccc|ccc|ccc|ccc}
\toprule[1.5pt]
\multirow{2}{*}{\textbf{Backbone}}&\multirow{2}{*}{\textbf{Method}} & \multicolumn{3}{c|}{\textbf{Modality}} &\multirow{2}{*}{\textbf{Ext.}} & \multicolumn{3}{c|}{Rotation} & \multicolumn{3}{c|}{Warping} & \multicolumn{3}{c|}{Keystoning} & \multicolumn{3}{c|}{Watermark} & \multicolumn{3}{c|}{Background} & \multicolumn{3}{c|}{Illumination} & \multicolumn{3}{c|}{Ink-Bleeding} & \multicolumn{3}{c|}{Ink-Holdout} & \multicolumn{3}{c|}{Defocus} & \multicolumn{3}{c|}{Vibration} & \multicolumn{3}{c|}{Speckle} & \multicolumn{3}{c}{Texture} \\
& & V & L & T & & L1 & L2 & L3 & L1 & L2 & L3 & L1 & L2 & L3 & L1 & L2 & L3 & L1 & L2 & L3 & L1 & L2 & L3 & L1 & L2 & L3 & L1 & L2 & L3 & L1 & L2 & L3 & L1 & L2 & L3 & L1 & L2 & L3 & L1 & L2 & L3 \\
\midrule \midrule
ResNeXt~\cite{Saining2016resnext}&LayoutParser ~\cite{shen2021layoutparser}& \cmark & \xmark & \xmark & \xmark & 065.3 & 114.3 & 129.8 & 088.1 & 101.7 & 105.9 & 096.3 & 089.0 & 094.0 & 187.3 & 161.6 & 161.6 & 210.8 & 201.5 & 196.7 & 204.6 & 196.6 & 190.6 & 189.1 & 151.1 & 150.3 & 181.6 & 103.8 & 099.4 & 247.1 & 629.6 & 749.6 & 329.3 & 558.3 & 371.0 & 175.3 & 183.9 & 201.5 & 262.3 & 235.5 & 241.3\\
ResNet~\cite{kaiming2015resnet}&Faster R-CNN
~\cite{ren2017FasterRcnn} & \cmark & \xmark & \xmark & \xmark & 083.2 & 090.0 & 107.0 & 107.1 & 112.6 & 115.9 & 083.0 & 072.0 & 071.9 & 274.2 & 228.4 & 211.5 & 176.8 & 175.0 & 174.2 & 189.9 & 170.2 & 164.5 & 284.6 & 204.3 & 145.5 & 270.7 & 139.5 & 090.2 & 299.1 & 239.5 & 210.2 & 265.2 & 230.9 & 211.3 & 198.7 & 180.4 & 178.0 & 210.1 & 182.4 & 169.5\\
ResNet~\cite{kaiming2015resnet}&DocSegTr~\cite{biswas2022docsegtr}& \cmark & \xmark & \xmark & \xmark &  072.6 & 141.4 & 133.9 & 083.9 & 110.4 & 127.5 & 086.8 & 075.3 & 088.7 & 193.3 & 156.2 & 152.6 & 216.8 & 201.8 & 188.9 & 286.5 & 292.2 & 285.5 & 203.0 & 157.0 & 131.9 & 198.4 & 127.6 & 230.9 & 200.6 & 646.6 & 775.3 & 588.0 & 637.0 & 379.1 & 151.1 & 157.0 & 172.6 & 261.3 & 235.8 & 241.3\\
ResNet~\cite{kaiming2015resnet}&Mask R-CNN 
~\cite{He_2017_MaskRcnn} & \cmark & \xmark & \xmark & \xmark & 090.2 & 097.7 & 114.1 & 116.6 & 119.0 & 118.3 & 091.3 & 079.2 & 078.8 & 308.7 & 248.8 & 227.6 & 194.4 & 190.9 & 191.0 & 216.2 & 192.1 & 184.5 & 324.5 & 227.9 & 160.7 & 309.4 & 154.6 & 105.0 & 345.5 & 270.9 & 231.5 & 300.5 & 291.3 & 269.0 & 225.4 & 183.5 & 171.5 & 189.3 & 162.6 & 153.6
\\
Swin~\cite{liu2021swin}&SwinDocSegmenter~\cite{banerjee2023swindocsegmenter}&\cmark & \xmark & \xmark & \xmark   & 063.3 & 100.0 & 130.0 & 061.2 & 074.9 & 087.5 & 139.9 & 104.8 & 103.0 & 146.9 & 120.6 & 109.8 & 216.8 & 201.8 & 184.7 & 122.8 & 112.0 & 107.0 & 128.4 & 125.1 & 141.5 & 119.4 & 073.7 & 083.9 & 124.5 & 1094.2 & 866.6 & 240.4 & 629.1 & 377.2 & 254.4 & 285.6 & 249.1 & 260.7 & 235.5 & 241.0\\
\rowfont{\color{gray!30}} DiT~\cite{li2022dit}&Cascade R-CNN~\cite{cai2018cascade} & \cmark & \xmark & \xmark & \cmark &076.0 & 123.2 & 132.8 & 052.8 & 064.0 & 065.1 & 041.1 & 048.6 & 060.1 & 101.9 & 086.2 & 080.3 & 085.4 & 075.2 & 070.6 & 124.9 & 118.3 & 118.4 & 100.6 & 092.1 & 082.3 & 094.2 & 054.7 & 045.1 & 096.6 & 069.4 & 071.9 & 070.6 & 093.8 & 255.5 & 114.7 & 125.8 & 125.3 & 149.2 & 139.2 & 142.5\\
\rowfont{\color{gray!30}} LayoutLMv3~\cite{huang2022layoutlmv3}&Cascade R-CNN~\cite{cai2018cascade} & \cmark & \cmark & \cmark & \cmark  &076.0 & 121.9 & 130.6 & 057.0 & 067.2 & 067.1 & 049.4 & 055.0 & 066.4 & 095.9 & 086.2 & 076.8 & 106.9 & 122.8 & 122.2 & 103.9 & 133.8 & 137.5 & 093.7 & 081.5 & 067.9 & 097.5 & 064.2 & 091.6 & 091.0 & 121.7 & 328.2 & 101.9 & 377.2 & 360.6 & 081.6 & 073.2 & 071.4 & 111.0 & 139.2 & 153.8\\
\midrule
Swin~\cite{liu2021swin}& Co-DINO~\cite{zong2022codetr} & \cmark & \xmark & \xmark & \xmark  & 065.8 & 118.4 & 131.8 & 055.4 & 078.6 & 102.7 & 044.8 & 048.4 & 057.8 & 091.4 & 081.9 & 084.8 & 172.6 & 161.6 & 143.3 & 083.9 & 076.5 & 073.1 & 097.2 & 089.7 & 067.9 & 097.5 & 050.7 & 040.9 & 091.0 & 082.5 & 143.7 & 084.9 & 190.9 & 285.6 & 158.3 & 239.3 & 221.1 & 260.2 & 233.9 & 237.6\\
InternImage~\cite{wang2023internimage}& Cascade R-CNN~\cite{cai2018cascade} & \cmark & \xmark & \xmark & \xmark &073.9 & 143.0 & 134.4 & 059.1 & 083.1 & 119.5 & 047.3 & 052.3 & 070.5 & 122.9 & 113.1 & 111.6 & 175.6 & 163.7 & 150.6 & 160.5 & 152.0 & 145.3 & 114.5 & 100.4 & 081.5 & 121.1 & 067.4 & 055.0 & 118.9 & 102.1 & 275.0 & 109.8 & 278.2 & 335.6 & 151.9 & 186.7 & 194.1 & 262.3 & 235.8 & 241.5\\
InternImage~\cite{wang2023internimage}& DINO ~\cite{zhang2022dino}& \cmark & \xmark & \xmark & \xmark  & 065.8 & 118.4 & 131.8 & 055.4 & 078.6 & 102.7 & 044.8 & 048.4 & 057.8 & 091.4 & 081.9 & 084.8 & 172.6 & 161.6 & 143.3 & 083.9 & 076.5 & 073.1 & 097.2 & 089.7 & 067.9 & 097.5 & 050.7 & 040.9 & 091.0 & 082.5 & 143.7 & 084.9 & 190.9 & 285.6 & 158.3 & 239.3 & 221.1 & 260.2 & 233.9 & 237.6\\
InternImage~\cite{wang2023internimage}& Co-DINO~\cite{zong2022codetr} & \cmark & \xmark & \xmark & \xmark  & 130.1 & 149.4 & 134.3 & 142.5 & 249.8 & 292.5 & 235.4 & 178.5 & 165.7 & 112.4 & 093.7 & 088.4 & 173.0 & 146.1 & 127.8 & 120.7 & 111.0 & 108.8 & 140.5 & 165.3 & 270.2 & 131.2 & 080.1 & 068.0 & 120.7 & 1075.9 & 866.6 & 360.6 & 603.5 & 372.5 & 196.3 & 227.1 & 196.1 & 227.3 & 212.7 & 217.4\\
\rowcolor[gray]{.9} InternImage~\cite{wang2023internimage}&RoDLA (Ours) & \cmark & \xmark & \xmark & \xmark & 072.8 & 129.0 & 130.9 & 058.0 & 089.0 & 125.9 & 048.2 & 052.3 & 064.3 & 086.9 & 075.4 & 075.0 & 152.8 & 146.8 & 133.8 & 083.9 & 078.3 & 074.0 & 093.7 & 089.7 & 073.5 & 090.8 & 056.3 & 044.8 & 079.9 & 073.3 & 143.7 & 087.5 & 188.9 & 236.7 & 148.6 & 193.0 & 159.8 & 262.3 & 235.5 & 241.3\\

\bottomrule[1.5pt]

\end{tabu}
}
\vskip -1ex
\end{table*}
To comprehensively evaluate methods on PubLayNet-P, we report the best and the worst results in Table~\ref{tab:sota_publaynet_p_best} and Table~\ref{tab:sota_publaynet_p_worst} which according to the severity levels. The best-case metric in our robustness benchmark is calculated using the best outcomes under each perturbation according to result in severity level, while the worst-case uses the poorest outcomes which selected from result of severity levels. These metrics reveal the range of a model's robustness to the perturbations we have established, showing the potential fluctuation in robustness performance across different models. Besides, exhaustive results for the \textbf{P-Avg} metric are presented in Table~\ref{tab:pavg_detail_publaynet}, while the ones for the \textbf{RD} metric are detailed in Table~\ref{tab:rd_detail_publaynet}.

\subsection{Detailed Results on DocLayNet-P}
To comprehensively compare the performance of different methods across various datasets, we also present the best and worst outcomes on the DocLayNet-P in Table~\ref{tab:sota_doclaynet_p_best} and Table~\ref{tab:sota_doclaynet_p_worst}, respectively. Furthermore, comparisons between different models performance in \textbf{P-Avg} and \textbf{RD} at different severity levels are provided in Table~\ref{tab:pavg_detail_doclaynet} and Table~\ref{tab:rd_detail_doclaynet}.
\begin{table*}[t]
\centering
\caption{The robustness benchmark with the \textbf{best-case} result on DocLayNet-P dataset.}
\vskip -2ex
\label{tab:sota_doclaynet_p_best}
\setlength{\tabcolsep}{3pt}
\resizebox{\linewidth}{!}{
\begin{tabu}{l|l|ccc|c|c|llllllllllll|c|c}
\toprule[1.5pt]
\multirow{2}{*}{\textbf{Backbone}}&\multirow{2}{*}{\textbf{Method}} & \multicolumn{3}{c|}{\textbf{Modality}} &\multirow{2}{*}{\textbf{Ext.}} & \multirow{2}{*}{\textbf{Clean}} & \multicolumn{3}{c|}{\textbf{Spatial}} & \multicolumn{2}{c|}{\textbf{Content}} & \multicolumn{3}{c|}{\textbf{Inconsistency}}  & \multicolumn{2}{c|}{\textbf{Blur}}  & \multicolumn{2}{c|}{\textbf{Noise}}  & \multirow{2}{*}{\textbf{P-Avgt$\uparrow$}} & \multirow{2}{*}{\textbf{mRD{$\downarrow$}}}\\
& & V & L & T &&&{P1} & {P2} & \multicolumn{1}{c|}{P3} & {P4} & \multicolumn{1}{c|}{P5} & {P6} & {P7} & \multicolumn{1}{c|}{P8} & {P9} & \multicolumn{1}{c|}{P10} & {P11} & {P12} &  \\
\midrule \midrule
ResNet~\cite{kaiming2015resnet}&Faster R-CNN
~\cite{ren2017FasterRcnn} & \cmark & \xmark & \xmark & \xmark & 73.4&37.6 & 65.2 & 62.1 & 70.2 & 65.8 & 70.6 & 72.1 & 72.5 & 72.3 & 69.4 & 48.3 & 35.5 & 61.8 &178.1\\
ResNet~\cite{kaiming2015resnet}&DocSegTr~\cite{biswas2022docsegtr}& \cmark & \xmark & \xmark & \xmark & 69.3 & 40.3 & 60.5 & 62.0 & 69.9 & 65.3 & 67.6 & 46.3 & 56.9 & 61.7 & 26.5 & 49.7 & 35.1 & 53.5&190.7
\\
ResNet~\cite{kaiming2015resnet}&Mask R-CNN 
~\cite{He_2017_MaskRcnn} & \cmark & \xmark & \xmark & \xmark & 73.5 & 44.2 & 65.2 & 62.6 & 70.4 & 65.3 & 71.1 & 72.6 & 72.7 & 72.9 & 69.3 & 44.7 & 35.2 & 62.2& 176.1\\
Swin~\cite{liu2021swin}&SwinDocSegmenter~\cite{banerjee2023swindocsegmenter}&\cmark & \xmark & \xmark & \xmark  & 76.9 & 48.6 & 67.4 & 68.6 & 69.9 & 67.8 & 75.1 & 75.2 & 76.7 & 75.2 & 70.6 & 47.6 & 39.1 & 65.2 & 161.7\\
\rowfont{\color{gray!30}} DiT~\cite{li2022dit}&Cascade R-CNN~\cite{cai2018cascade} & \cmark & \xmark & \xmark & \cmark &62.1 &35.2 & 58.5 & 58.3 & 61.1 & 52.5 & 60.2 & 61.4 & 62.3 & 63.0 & 63.0 & 52.3 & 56.3 & 57.0 & 170.9
\\
\rowfont{\color{gray!30}} LayoutLMv3~\cite{huang2022layoutlmv3}&Cascade R-CNN~\cite{cai2018cascade} & \cmark & \cmark & \cmark & \cmark & 75.1 &56.7 & 75.1 & 65.8 & 75.2 & 67.5 & 75.2 & 69.1 & 75.5 & 62.6 & 39.6 & 72.3 & 73.7 & 67.4 &151.8
\\
\midrule
\rowcolor[gray]{.9} InternImage~\cite{wang2023internimage}&RoDLA (Ours) & \cmark & \xmark & \xmark & \xmark & 80.5 & 49.6 & 72.6 & 73.2 & 79.0 & 74.2 & 80.3 & 80.7 & 81.6 & 82.0 & 81.8 & 58.9 & 59.7 & 72.8 &117.9
\\

\bottomrule[1.5pt]

\end{tabu}
}
\end{table*}
\begin{table*}[!ht]
\centering
\caption{The robustness benchmark with the \textbf{worst-case} result on DocLayNet-P dataset.}
\vskip -2ex
\label{tab:sota_doclaynet_p_worst}
\setlength{\tabcolsep}{3pt}
\resizebox{\linewidth}{!}{
\begin{tabu}{l|l|ccc|c|c|llllllllllll|c|c}
\toprule[1.5pt]
\multirow{2}{*}{\textbf{Backbone}}&\multirow{2}{*}{\textbf{Method}} & \multicolumn{3}{c|}{\textbf{Modality}} &\multirow{2}{*}{\textbf{Ext.}} & \multirow{2}{*}{\textbf{Clean}} & \multicolumn{3}{c|}{\textbf{Spatial}} & \multicolumn{2}{c|}{\textbf{Content}} & \multicolumn{3}{c|}{\textbf{Inconsistency}}  & \multicolumn{2}{c|}{\textbf{Blur}}  & \multicolumn{2}{c|}{\textbf{Noise}}  & \multirow{2}{*}{\textbf{P-Avgt$\uparrow$}} & \multirow{2}{*}{\textbf{mRD{$\downarrow$}}}\\
& & V & L & T &&&{P1} & {P2} & \multicolumn{1}{c|}{P3} & {P4} & \multicolumn{1}{c|}{P5} & {P6} & {P7} & \multicolumn{1}{c|}{P8} & {P9} & \multicolumn{1}{c|}{P10} & {P11} & {P12} &  \\
\midrule \midrule
ResNet~\cite{kaiming2015resnet}&Faster R-CNN
~\cite{ren2017FasterRcnn} & \cmark & \xmark & \xmark & \xmark & 73.4&01.4 & 58.8 & 46.3 & 67.4 & 61.0 & 70.1 & 66.2 & 63.2 & 27.7 & 05.3 & 09.9 & 27.5 & 42.1&221.2\\
ResNet~\cite{kaiming2015resnet}&DocSegTr~\cite{biswas2022docsegtr}& \cmark & \xmark & \xmark & \xmark & 69.3 & 02.6 & 49.1 & 42.7 & 65.8 & 32.5 & 60.3 & 39.8 & 44.7 & 44.6 & 05.4 & 23.4 & 22.0 & 36.1 & 314.9
\\
ResNet~\cite{kaiming2015resnet}&Mask R-CNN 
~\cite{He_2017_MaskRcnn} & \cmark & \xmark & \xmark & \xmark & 73.5 & 08.4 & 58.7 & 53.4 & 67.3 & 60.7 & 70.5 & 66.4 & 62.5 & 26.7 & 05.3 & 08.4 & 27.5 & 43.0&220.3
\\
Swin~\cite{liu2021swin}&SwinDocSegmenter~\cite{banerjee2023swindocsegmenter}&\cmark & \xmark & \xmark & \xmark  & 76.9 & 07.2 & 54.6 & 55.9 & 68.0 & 58.2 & 74.9 & 67.8 & 71.0 & 31.9 & 05.0 & 25.2 & 35.7 & 46.3 & 208.1
\\
\rowfont{\color{gray!30}} DiT~\cite{li2022dit}&Cascade R-CNN~\cite{cai2018cascade} & \cmark & \xmark & \xmark & \cmark &62.1 &01.3 & 51.4 & 44.1 & 55.9 & 49.1 & 58.4 & 58.5 & 57.7 & 61.9 & 45.6 & 48.3 & 51.5 & 48.7&271.0
\\
\rowfont{\color{gray!30}} LayoutLMv3~\cite{huang2022layoutlmv3}&Cascade R-CNN~\cite{cai2018cascade} & \cmark & \cmark & \cmark & \cmark & 75.1 &21.8 & 67.8 & 59.5 & 74.2 & 59.3 & 59.6 & 63.8 & 07.0 & 47.3 & 00.6 & 70.1 & 67.8 & 49.9&234.7
\\
\midrule
\rowcolor[gray]{.9} InternImage~\cite{wang2023internimage}&RoDLA (Ours) & \cmark & \xmark & \xmark & \xmark & 80.5 & 04.3 & 59.4 & 59.1 & 74.5 & 70.4 & 80.0 & 77.1 & 73.4 & 77.3 & 44.7 & 26.3 & 48.7 & 57.9 &157.3
\\

\bottomrule[1.5pt]

\end{tabu}
}
\end{table*}
\begin{table*}[!ht]
\centering
\caption{The detailed per-level \textbf{P-Avg$\uparrow$} results on DocLayNet-P dataset. \textbf{L1}, \textbf{L2}, and \textbf{L3} stand for the severity levels from light to heavy. 
}
\vskip -2ex
\label{tab:pavg_detail_doclaynet}
\setlength{\tabcolsep}{3pt}
\renewcommand{\arraystretch}{1.3}
\resizebox{\linewidth}{!}{
\begin{tabu}{l|l|ccc|c|c|ccc|ccc|ccc|ccc|ccc|ccc|ccc|ccc|ccc|ccc|ccc|ccc}
\toprule[1.5pt]
\multirow{2}{*}{\textbf{Backbone}}&\multirow{2}{*}{\textbf{Method}} & \multicolumn{3}{c|}{\textbf{Modality}} &\multirow{2}{*}{\textbf{Ext.}} & \multirow{2}{*}{\textbf{Clean}} & \multicolumn{3}{c|}{Rotation} & \multicolumn{3}{c|}{Warping} & \multicolumn{3}{c|}{Keystoning} & \multicolumn{3}{c|}{Watermark} & \multicolumn{3}{c|}{Background} & \multicolumn{3}{c|}{Illumination} & \multicolumn{3}{c|}{Ink-Bleeding} & \multicolumn{3}{c|}{Ink-Holdout} & \multicolumn{3}{c|}{Defocus} & \multicolumn{3}{c|}{Vibration} & \multicolumn{3}{c|}{Speckle} & \multicolumn{3}{c}{Texture} \\
& & V & L & T && & L1 & L2 & L3 & L1 & L2 & L3 & L1 & L2 & L3 & L1 & L2 & L3 & L1 & L2 & L3 & L1 & L2 & L3 & L1 & L2 & L3 & L1 & L2 & L3 & L1 & L2 & L3 & L1 & L2 & L3 & L1 & L2 & L3 & L1 & L2 & L3 \\
\midrule \midrule
ResNet~\cite{kaiming2015resnet}&Faster R-CNN
~\cite{ren2017FasterRcnn} & \cmark & \xmark & \xmark & \xmark & 73.4&37.6 & 11.0 & 01.4 & 65.2 & 60.4 & 58.8 & 62.1 & 52.7 & 46.3 & 70.2 & 68.6 & 67.4 & 65.8 & 61.6 & 61.0 & 70.6 & 70.5 & 70.1 & 72.1 & 69.8 & 66.2 & 72.5 & 70.0 & 63.2 & 72.3 & 66.5 & 27.7 & 69.4 & 19.6 & 05.3 & 48.3 & 21.5 & 09.9 & 35.5 & 27.5 & 27.7

\\
ResNet~\cite{kaiming2015resnet}&DocSegTr~\cite{biswas2022docsegtr}& \cmark & \xmark & \xmark & \xmark & 69.3 & 40.3 & 23.7 & 02.6 & 60.5 & 58.4 & 49.1 & 62.0 & 51.7 & 42.7 & 69.9 & 68.2 & 65.8 & 65.3 & 54.1 & 32.5 & 64.3 & 67.6 & 60.3 & 46.3 & 42.6 & 39.8 & 56.9 & 49.3 & 44.7 & 61.7 & 45.2 & 44.6 & 26.5 & 24.2 & 05.4 & 49.7 & 23.4 & 24.6 & 35.1 & 32.9 & 22.0

\\
ResNet~\cite{kaiming2015resnet}&Mask R-CNN 
~\cite{He_2017_MaskRcnn} & \cmark & \xmark & \xmark & \xmark & 73.5 & 44.2 & 18.1 & 08.4 & 65.2 & 60.4 & 58.7 & 62.6 & 53.4 & 56.9 & 70.4 & 68.7 & 67.3 & 65.3 & 61.3 & 60.7 & 71.1 & 71.0 & 70.5 & 72.6 & 70.3 & 66.4 & 72.7 & 69.5 & 62.5 & 72.9 & 66.3 & 26.7 & 69.3 & 21.0 & 05.3 & 44.7 & 18.1 & 08.4 & 35.2 & 28.2 & 27.5
\\
Swin~\cite{liu2021swin}&SwinDocSegmenter~\cite{banerjee2023swindocsegmenter}&\cmark & \xmark & \xmark & \xmark  & 76.9 & 48.6 & 19.0 & 07.2 & 67.4 & 58.6 & 54.6 & 68.6 & 61.9 & 55.9 & 68.0 & 69.1 & 69.9 & 67.8 & 59.7 & 58.2 & 74.9 & 75.1 & 74.9 & 75.2 & 72.1 & 67.8 & 76.7 & 75.6 & 71.0 & 75.2 & 65.9 & 31.9 & 70.6 & 21.5 & 05.0 & 47.6 & 32.7 & 25.2 & 39.1 & 35.7 & 37.0

\\
\rowfont{\color{gray!30}} DiT~\cite{li2022dit}&Cascade R-CNN~\cite{cai2018cascade} & \cmark & \xmark & \xmark & \cmark &62.1 &35.2 & 08.3 & 01.3 & 58.5 & 54.1 & 51.4 & 58.3 & 50.3 & 44.1 & 61.1 & 55.9 & 55.9 & 52.5 & 49.3 & 49.1 & 60.2 & 59.1 & 58.4 & 61.4 & 59.9 & 58.5 & 62.3 & 61.9 & 57.7 & 61.9 & 62.8 & 63.0 & 63.0 & 58.0 & 45.6 & 48.3 & 52.2 & 52.3 & 56.3 & 52.8 & 51.5

\\
\rowfont{\color{gray!30}} LayoutLMv3~\cite{huang2022layoutlmv3}&Cascade R-CNN~\cite{cai2018cascade} & \cmark & \cmark & \cmark & \cmark & 75.1 & 56.7 & 33.2 & 21.8 & 75.1 & 73.0 & 67.8 & 65.8 & 60.9 & 59.5 & 75.2 & 75.1 & 74.2 & 67.5 & 61.4 & 59.3 & 75.1 & 75.2 & 59.6 & 69.1 & 65.1 & 63.8 & 75.5 & 41.5 & 07.0 & 62.6 & 54.4 & 47.3 & 39.6 & 07.5 & 00.6 & 72.3 & 71.8 & 70.1 & 73.7 & 68.9 & 67.8\\
\midrule
\rowcolor[gray]{.9} InternImage~\cite{wang2023internimage}&RoDLA (Ours) & \cmark & \xmark & \xmark & \xmark & 80.5 & 49.6 & 17.8 & 04.3 & 72.6 & 64.2 & 59.4 & 73.2 & 65.8 & 59.1 & 79.0 & 76.4 & 74.5 & 74.2 & 71.1 & 70.4 & 80.3 & 80.2 & 80.0 & 80.7 & 78.4 & 77.1 & 81.6 & 79.0 & 73.4 & 81.5 & 82.0 & 77.3 & 81.8 & 67.5 & 44.7 & 58.9 & 37.8 & 26.3 & 59.7 & 51.3 & 48.7

\\

\bottomrule[1.5pt]

\end{tabu}
}
\end{table*}
\begin{table*}[!ht]
\centering
\caption{
The detailed per-level \textbf{RD$\downarrow$} results on DocLayNet-P dataset. \textbf{L1}, \textbf{L2}, and \textbf{L3} stand for the severity levels from light to heavy.
}
\vskip -2ex
\label{tab:rd_detail_doclaynet}
\setlength{\tabcolsep}{3pt}
\renewcommand{\arraystretch}{1.3}
\resizebox{\linewidth}{!}{
\begin{tabu}{l|l|ccc|c|ccc|ccc|ccc|ccc|ccc|ccc|ccc|ccc|ccc|ccc|ccc|ccc}
\toprule[1.5pt]
\multirow{2}{*}{\textbf{Backbone}}&\multirow{2}{*}{\textbf{Method}} & \multicolumn{3}{c|}{\textbf{Modality}} &\multirow{2}{*}{\textbf{Ext.}} & \multicolumn{3}{c|}{Rotation} & \multicolumn{3}{c|}{Warping} & \multicolumn{3}{c|}{Keystoning} & \multicolumn{3}{c|}{Watermark} & \multicolumn{3}{c|}{Background} & \multicolumn{3}{c|}{Illumination} & \multicolumn{3}{c|}{Ink-Bleeding} & \multicolumn{3}{c|}{Ink-Holdout} & \multicolumn{3}{c|}{Defocus} & \multicolumn{3}{c|}{Vibration} & \multicolumn{3}{c|}{Speckle} & \multicolumn{3}{c}{Texture} \\
& & V & L & T & & L1 & L2 & L3 & L1 & L2 & L3 & L1 & L2 & L3 & L1 & L2 & L3 & L1 & L2 & L3 & L1 & L2 & L3 & L1 & L2 & L3 & L1 & L2 & L3 & L1 & L2 & L3 & L1 & L2 & L3 & L1 & L2 & L3 & L1 & L2 & L3 \\
\midrule \midrule
ResNet~\cite{kaiming2015resnet}&Faster R-CNN
~\cite{ren2017FasterRcnn} & \cmark & \xmark & \xmark & \xmark & 128.2 & 121.7 & 122.3 & 146.3 & 146.9 & 141.7 & 126.0 & 109.7 & 105.5 & 283.6 & 247.6 & 230.1 & 163.2 & 156.0 & 156.3 & 221.0 & 202.3 & 197.2 & 290.7 & 236.4 & 190.9 & 282.1 & 179.1 & 114.6 & 299.5 & 263.6 & 263.1 & 276.0 & 265.3 & 241.3 & 241.5 & 216.6 & 203.0 & 196.2 & 177.9 & 171.7\\
ResNet~\cite{kaiming2015resnet}&DocSegTr~\cite{biswas2022docsegtr}& \cmark & \xmark & \xmark & \xmark & 122.8 & 104.3 & 121.0 & 166.2 & 154.4 & 174.8 & 126.6 & 111.9 & 112.6 & 286.6 & 250.8 & 241.2 & 165.6 & 186.5 & 270.6 & 268.6 & 222.0 & 261.9 & 560.1 & 450.3 & 339.6 & 442.7 & 302.6 & 172.3 & 414.2 & 431.4 & 201.8 & 662.5 & 250.1 & 240.9 & 234.6 & 211.3 & 169.8 & 197.7 & 164.5 & 185.1\\
ResNet~\cite{kaiming2015resnet}&Mask R-CNN 
~\cite{He_2017_MaskRcnn} & \cmark & \xmark & \xmark & \xmark & 114.7 & 111.9 & 113.7 & 146.3 & 146.9 & 141.8 & 124.5 & 108.0 & 084.6 & 281.9 & 246.7 & 230.7 & 165.5 & 157.3 & 157.5 & 217.2 & 198.7 & 195.0 & 286.4 & 232.9 & 189.5 & 280.0 & 182.0 & 116.9 & 293.6 & 265.4 & 266.9 & 277.0 & 260.6 & 241.2 & 258.0 & 225.9 & 206.4 & 197.3 & 176.0 & 172.2
\\
Swin~\cite{liu2021swin}&SwinDocSegmenter~\cite{banerjee2023swindocsegmenter}&\cmark & \xmark & \xmark & \xmark  &105.6 & 110.7 & 115.2 & 137.2 & 153.8 & 156.1 & 104.4 & 088.4 & 086.6 & 304.2 & 243.5 & 212.1 & 153.5 & 163.9 & 167.4 & 188.7 & 170.6 & 165.5 & 258.6 & 218.9 & 181.7 & 239.5 & 145.8 & 090.4 & 268.4 & 268.2 & 248.1 & 264.6 & 258.9 & 242.0 & 244.6 & 185.7 & 168.5 & 185.5 & 157.8 & 149.5
\\
\rowfont{\color{gray!30}} DiT~\cite{li2022dit}&Cascade R-CNN~\cite{cai2018cascade} & \cmark & \xmark & \xmark & \cmark &133.2 & 125.3 & 122.5 & 174.5 & 170.2 & 166.9 & 138.6 & 115.4 & 109.7 & 370.1 & 347.2 & 311.3 & 226.7 & 206.0 & 204.0 & 299.9 & 280.2 & 274.5 & 402.4 & 314.3 & 234.2 & 387.0 & 227.7 & 131.7 & 411.9 & 293.0 & 134.7 & 333.1 & 138.5 & 138.6 & 241.5 & 132.0 & 107.5 & 133.1 & 115.7 & 115.1
\\
\rowfont{\color{gray!30}} LayoutLMv3~\cite{huang2022layoutlmv3}&Cascade R-CNN~\cite{cai2018cascade} & \cmark & \cmark & \cmark & \cmark & 089.0 & 091.3 & 097.0 & 104.7 & 100.2 & 110.7 & 113.8 & 090.6 & 079.6 & 235.8 & 196.4 & 182.3 & 155.2 & 157.1 & 163.0 & 187.6 & 170.2 & 266.4 & 322.3 & 273.8 & 204.5 & 252.0 & 349.4 & 289.7 & 404.7 & 358.9 & 191.8 & 544.3 & 305.2 & 253.1 & 129.4 & 077.9 & 067.4 & 080.0 & 076.2 & 076.5

\\
\midrule
\rowcolor[gray]{.9} InternImage~\cite{wang2023internimage}&RoDLA (Ours) & \cmark & \xmark & \xmark & \xmark &103.6 & 112.4 & 118.8 & 115.2 & 132.9 & 139.5 & 089.2 & 079.3 & 080.3 & 199.8 & 186.0 & 179.9 & 123.0 & 117.5 & 118.7 & 148.3 & 135.7 & 132.0 & 201.4 & 169.3 & 129.3 & 188.9 & 125.4 & 082.8 & 200.1 & 141.7 & 082.7 & 164.1 & 107.2 & 140.8 & 191.8 & 171.6 & 166.1 & 122.7 & 119.4 & 121.8\\

\bottomrule[1.5pt]

\end{tabu}
}
\vskip -1ex
\end{table*}

\subsection{Detailed Results on M$^6$Doc-P}
We showcase the best and worst model performances on the M$^6$Doc-P dataset in Table~\ref{tab:sota_m6doc_p_best} and Table~\ref{tab:sota_m6doc_p_worst}. We hope the elaborate result from three datasets illustrates the DLA model's robustness using data. Detailed \textbf{P-Avg} and \textbf{RD} metrics for M$^6$Doc-P are in Table~\ref{tab:pavg_detail_m6doc} and Table~\ref{tab:rd_detail_m6doc}, respectively.
\begin{table*}[t]
\centering
\caption{The robustness benchmark with the \textbf{best-case} result on M$^6$Doc dataset.}
\vskip -2ex
\label{tab:sota_m6doc_p_best}
\setlength{\tabcolsep}{3pt}
\resizebox{\linewidth}{!}{
\begin{tabu}{l|l|ccc|c|c|llllllllllll|c|c}
\toprule[1.5pt]
\multirow{2}{*}{\textbf{Backbone}}&\multirow{2}{*}{\textbf{Method}} & \multicolumn{3}{c|}{\textbf{Modality}} &\multirow{2}{*}{\textbf{Ext.}} & \multirow{2}{*}{\textbf{Clean}} & \multicolumn{3}{c|}{\textbf{Spatial}} & \multicolumn{2}{c|}{\textbf{Content}} & \multicolumn{3}{c|}{\textbf{Inconsistency}}  & \multicolumn{2}{c|}{\textbf{Blur}}  & \multicolumn{2}{c|}{\textbf{Noise}}  & \multirow{2}{*}{\textbf{P-Avgt$\uparrow$}} & \multirow{2}{*}{\textbf{mRD{$\downarrow$}}}\\
& & V & L & T &&&{P1} & {P2} & \multicolumn{1}{c|}{P3} & {P4} & \multicolumn{1}{c|}{P5} & {P6} & {P7} & \multicolumn{1}{c|}{P8} & {P9} & \multicolumn{1}{c|}{P10} & {P11} & {P12} &  \\
\midrule \midrule
ResNet~\cite{kaiming2015resnet}&Faster R-CNN
~\cite{ren2017FasterRcnn} & \cmark & \xmark & \xmark & \xmark & 62.0&44.6 & 60.1 & 56.6 & 61.7 & 54.0 & 59.3 & 62.5 & 61.9 & 62.1 & 59.6 & 52.2 & 40.7 & 56.3&177.2
\\
ResNet~\cite{kaiming2015resnet}&DocSegTr~\cite{biswas2022docsegtr}& \cmark & \xmark & \xmark & \xmark & 60.3 & 38.4 & 52.1 & 48.5 & 52.7 & 32.6 & 72.7 & 88.3 & 43.7 & 89.2 & 58.1 & 41.3 & 48.8 & 55.5&168.6
\\
ResNet~\cite{kaiming2015resnet}&Mask R-CNN 
~\cite{He_2017_MaskRcnn} & \cmark & \xmark & \xmark & \xmark & 61.9 & 61.7 & 60.6 & 62.3 & 60.3 & 61.8 & 55.8 & 58.9 & 55.9 & 51.7 & 61.9 & 59.6 & 62.2 & 59.4&144.2
\\
Swin~\cite{liu2021swin}&SwinDocSegmenter~\cite{banerjee2023swindocsegmenter}&\cmark & \xmark & \xmark & \xmark  & 47.1 & 31.4 & 45.5 & 42.9 & 47.0 & 40.3 & 44.8 & 47.3 & 46.8 & 47.2 & 46.3 & 40.7 & 39.1 & 43.3&196.5\\
\rowfont{\color{gray!30}} DiT~\cite{li2022dit}&Cascade R-CNN~\cite{cai2018cascade} & \cmark & \xmark & \xmark & \cmark &70.2 &61.0 & 61.1 & 66.7 & 60.5 & 50.6 & 70.9 & 62.4 & 61.4 & 64.7 & 53.7 & 66.2 & 69.9 & 62.4&141.9
\\
\rowfont{\color{gray!30}} LayoutLMv3~\cite{huang2022layoutlmv3}&Cascade R-CNN~\cite{cai2018cascade} & \cmark & \cmark & \cmark & \cmark & 64.3 &60.7 & 64.6 & 63.1 & 64.4 & 56.1 & 64.8 & 52.2 & 63.4 & 59.1 & 45.2 & 64.0 & 68.4 & 60.5&156.3
\\
\midrule
\rowcolor[gray]{.9} InternImage~\cite{wang2023internimage}&RoDLA (Ours) & \cmark & \xmark & \xmark & \xmark & 70.0 & 58.4 & 68.2 & 66.3 & 68.8 & 62.4 & 68.2 & 69.7 & 69.9 & 69.2 & 68.5 & 62.9 & 55.6 & 65.7&125.3
\\

\bottomrule[1.5pt]

\end{tabu}
}
\vskip -1ex
\end{table*}
\begin{table*}[!ht]
\centering
\caption{The robustness benchmark with the \textbf{worst-case} result on M$^6$Doc dataset.}
\vskip -2ex
\label{tab:sota_m6doc_p_worst}
\setlength{\tabcolsep}{3pt}
\resizebox{\linewidth}{!}{
\begin{tabu}{l|l|ccc|c|c|llllllllllll|c|c}
\toprule[1.5pt]
\multirow{2}{*}{\textbf{Backbone}}&\multirow{2}{*}{\textbf{Method}} & \multicolumn{3}{c|}{\textbf{Modality}} &\multirow{2}{*}{\textbf{Ext.}} & \multirow{2}{*}{\textbf{Clean}} & \multicolumn{3}{c|}{\textbf{Spatial}} & \multicolumn{2}{c|}{\textbf{Content}} & \multicolumn{3}{c|}{\textbf{Inconsistency}}  & \multicolumn{2}{c|}{\textbf{Blur}}  & \multicolumn{2}{c|}{\textbf{Noise}}  & \multirow{2}{*}{\textbf{P-Avgt$\uparrow$}} & \multirow{2}{*}{\textbf{mRD{$\downarrow$}}}\\
& & V & L & T &&&{P1} & {P2} & \multicolumn{1}{c|}{P3} & {P4} & \multicolumn{1}{c|}{P5} & {P6} & {P7} & \multicolumn{1}{c|}{P8} & {P9} & \multicolumn{1}{c|}{P10} & {P11} & {P12} &  \\
\midrule \midrule
ResNet~\cite{kaiming2015resnet}&Faster R-CNN
~\cite{ren2017FasterRcnn} & \cmark & \xmark & \xmark & \xmark & 62.0&06.2 & 57.8 & 48.8 & 59.3 & 50.0 & 56.6 & 54.3 & 53.7 & 34.0 & 09.0 & 19.5 & 29.7 & 39.9&223.1\\
ResNet~\cite{kaiming2015resnet}&DocSegTr~\cite{biswas2022docsegtr}& \cmark & \xmark & \xmark & \xmark & 60.3 & 07.0 & 32.2 & 44.0 & 33.9 & 24.5 & 67.2 & 45.8 & 39.2 & 12.6 & 07.5 & 19.3 & 42.8 & 31.3&260.0\\
ResNet~\cite{kaiming2015resnet}&Mask R-CNN 
~\cite{He_2017_MaskRcnn} & \cmark & \xmark & \xmark & \xmark & 61.9 & 17.8 & 53.0 & 57.2 & 39.4 & 29.8 & 35.0 & 50.2 & 08.7 & 44.2 & 07.0 & 55.7 & 59.4 & 38.1&298.7\\
Swin~\cite{liu2021swin}&SwinDocSegmenter~\cite{banerjee2023swindocsegmenter}&\cmark & \xmark & \xmark & \xmark  & 47.1 & 07.9 & 40.1 & 35.2 & 45.8 & 37.5 & 43.2 & 46.1 & 42.9 & 40.0 & 20.7 & 30.6 & 35.7 & 35.5&295.6\\
\rowfont{\color{gray!30}} DiT~\cite{li2022dit}&Cascade R-CNN~\cite{cai2018cascade} & \cmark & \xmark & \xmark & \cmark &70.2 &51.0 & 56.9 & 60.2 & 52.6 & 49.6 & 69.0 & 60.1 & 39.8 & 58.6 & 12.3 & 64.8 & 68.9 & 53.6&203.7\\
\rowfont{\color{gray!30}} LayoutLMv3~\cite{huang2022layoutlmv3}&Cascade R-CNN~\cite{cai2018cascade} & \cmark & \cmark & \cmark & \cmark & 64.3 &46.4 & 54.4 & 60.6 & 62.0 & 48.6 & 55.7 & 50.3 & 29.6 & 51.6 & 06.2 & 62.5 & 60.1 & 49.0&217.7\\
\midrule
\rowcolor[gray]{.9} InternImage~\cite{wang2023internimage}&RoDLA (Ours) & \cmark & \xmark & \xmark & \xmark & 70.0 & 23.7 & 64.0 & 60.9 & 67.5 & 58.5 & 67.5 & 67.5 & 68.3 & 63.6 & 34.4 & 47.7 & 51.7 & 56.3&178.0\\

\bottomrule[1.5pt]

\end{tabu}
}
\vskip -1ex
\end{table*}
\begin{table*}[!ht]
\centering
\caption{
The detailed per-level \textbf{P-Avg$\uparrow$} results on M$^6$Doc-P dataset. \textbf{L1}, \textbf{L2}, and \textbf{L3} stand for the severity levels from light to heavy.
}
\vskip -2ex
\label{tab:pavg_detail_m6doc}
\renewcommand{\arraystretch}{1.3}
\setlength{\tabcolsep}{3pt}
\resizebox{\linewidth}{!}{
\begin{tabu}{l|l|ccc|c|c|ccc|ccc|ccc|ccc|ccc|ccc|ccc|ccc|ccc|ccc|ccc|ccc}
\toprule[1.5pt]
\multirow{2}{*}{\textbf{Backbone}}&\multirow{2}{*}{\textbf{Method}} & \multicolumn{3}{c|}{\textbf{Modality}} &\multirow{2}{*}{\textbf{Ext.}} & \multirow{2}{*}{\textbf{Clean}} & \multicolumn{3}{c|}{Rotation} & \multicolumn{3}{c|}{Warping} & \multicolumn{3}{c|}{Keystoning} & \multicolumn{3}{c|}{Watermark} & \multicolumn{3}{c|}{Background} & \multicolumn{3}{c|}{Illumination} & \multicolumn{3}{c|}{Ink-Bleeding} & \multicolumn{3}{c|}{Ink-Holdout} & \multicolumn{3}{c|}{Defocus} & \multicolumn{3}{c|}{Vibration} & \multicolumn{3}{c|}{Speckle} & \multicolumn{3}{c}{Texture} \\
& & V & L & T && & L1 & L2 & L3 & L1 & L2 & L3 & L1 & L2 & L3 & L1 & L2 & L3 & L1 & L2 & L3 & L1 & L2 & L3 & L1 & L2 & L3 & L1 & L2 & L3 & L1 & L2 & L3 & L1 & L2 & L3 & L1 & L2 & L3 & L1 & L2 & L3 \\
\midrule \midrule
ResNet~\cite{kaiming2015resnet}&Faster R-CNN
~\cite{ren2017FasterRcnn} & \cmark & \xmark & \xmark & \xmark & 62.0&44.6 & 24.4 & 06.2 & 60.1 & 58.4 & 57.8 & 56.6 & 50.5 & 48.8 & 61.7 & 60.1 & 59.3 & 54.0 & 50.0 & 52.1 & 59.3 & 58.1 & 56.6 & 62.5 & 60.6 & 54.3 & 61.9 & 60.0 & 53.7 & 62.1 & 56.9 & 34.0 & 59.6 & 20.6 & 09.0 & 52.2 & 32.1 & 19.5 & 40.7 & 31.5 & 29.7\\
ResNet~\cite{kaiming2015resnet}&DocSegTr~\cite{biswas2022docsegtr}& \cmark & \xmark & \xmark & \xmark & 60.3 & 38.4 & 23.4 & 07.0 & 52.1 & 37.3 & 32.2 & 48.5 & 47.8 & 44.0 & 52.7 & 34.3 & 33.9 & 30.3 & 32.6 & 24.5 & 72.7 & 67.9 & 67.2 & 88.3 & 86.7 & 45.8 & 42.7 & 43.7 & 39.2 & 89.2 & 50.6 & 12.6 & 58.1 & 14.6 & 07.5 & 41.3 & 30.4 & 19.3 & 48.8 & 46.6 & 42.8

\\
ResNet~\cite{kaiming2015resnet}&Mask R-CNN 
~\cite{He_2017_MaskRcnn} & \cmark & \xmark & \xmark & \xmark & 61.9 & 31.4 & 17.8 & 61.7 & 60.5 & 53.0 & 60.6 & 59.3 & 57.2 & 62.3 & 60.3 & 53.7 & 39.4 & 32.8 & 29.8 & 61.8 & 55.8 & 35.0 & 51.8 & 50.2 & 51.7 & 58.9 & 20.8 & 08.7 & 55.9 & 51.7 & 49.1 & 44.2 & 24.7 & 07.0 & 61.9 & 59.6 & 58.6 & 55.7 & 62.2 & 59.8 & 59.4\\
Swin~\cite{liu2021swin}&SwinDocSegmenter~\cite{banerjee2023swindocsegmenter}&\cmark & \xmark & \xmark & \xmark  & 47.1 & 31.4 & 16.3 & 07.9 & 45.5 & 42.3 & 40.1 & 42.9 & 37.3 & 35.2 & 47.0 & 46.2 & 45.8 & 40.3 & 38.7 & 37.5 & 44.7 & 44.8 & 43.2 & 47.3 & 46.9 & 46.1 & 46.8 & 45.6 & 42.9 & 47.2 & 45.2 & 40.0 & 46.3 & 33.9 & 20.7 & 40.7 & 34.4 & 30.6 & 39.1 & 35.7 & 37.0\\
\rowfont{\color{gray!30}} DiT~\cite{li2022dit}&Cascade R-CNN~\cite{cai2018cascade} & \cmark & \xmark & \xmark & \cmark &70.2 &61.0 & 54.8 & 51.0 & 61.1 & 60.6 & 56.9 & 66.7 & 62.5 & 60.2 & 60.2 & 60.5 & 52.6 & 50.6 & 50.1 & 49.6 & 70.4 & 70.9 & 69.0 & 62.4 & 61.6 & 60.1 & 61.4 & 54.5 & 39.8 & 64.7 & 62.6 & 58.6 & 53.7 & 34.9 & 12.3 & 64.8 & 66.2 & 64.9 & 69.9 & 69.1 & 68.9\\
\rowfont{\color{gray!30}} LayoutLMv3~\cite{huang2022layoutlmv3}&Cascade R-CNN~\cite{cai2018cascade} & \cmark & \cmark & \cmark & \cmark & 64.3 &60.7 & 53.2 & 46.4 & 64.6 & 62.8 & 54.4 & 63.1 & 61.3 & 60.6 & 64.4 & 63.6 & 62.0 & 56.1 & 50.8 & 48.6 & 64.8 & 62.7 & 55.7 & 52.2 & 50.3 & 51.8 & 63.4 & 48.8 & 29.6 & 59.1 & 54.7 & 51.6 & 45.2 & 22.6 & 06.2 & 64.0 & 62.8 & 62.5 & 68.4 & 64.4&60.1\\
\midrule
\rowcolor[gray]{.9} InternImage~\cite{wang2023internimage}&RoDLA (Ours) & \cmark & \xmark & \xmark & \xmark & 70.0 & 58.4 & 40.9 & 23.7 & 68.2 & 66.0 & 64.0 & 66.3 & 63.4 & 60.9 & 68.8 & 68.7 & 67.5 & 62.4 & 60.2 & 58.5 & 67.5 & 68.2 & 67.8 & 69.7 & 67.8 & 67.5 & 68.8 & 69.9 & 68.3 & 69.2 & 68.8 & 63.6 & 68.5 & 54.2 & 34.4 & 62.9 & 53.6 & 47.7 & 55.6 & 51.7&52.3

\\

\bottomrule[1.5pt]

\end{tabu}
}
\vskip -1ex
\end{table*}
\begin{table*}[!ht]
\centering
\caption{
The detailed per-level \textbf{RD$\downarrow$} results on M$^6$Doc-P dataset. \textbf{L1}, \textbf{L2}, and \textbf{L3} stand for the severity levels from light to heavy.
}
\vskip -2ex
\label{tab:rd_detail_m6doc}
\renewcommand{\arraystretch}{1.3}
\setlength{\tabcolsep}{3pt}
\resizebox{\linewidth}{!}{
\begin{tabu}{l|l|ccc|c|ccc|ccc|ccc|ccc|ccc|ccc|ccc|ccc|ccc|ccc|ccc|ccc}
\toprule[1.5pt]
\multirow{2}{*}{\textbf{Backbone}}&\multirow{2}{*}{\textbf{Method}} & \multicolumn{3}{c|}{\textbf{Modality}} &\multirow{2}{*}{\textbf{Ext.}} & \multicolumn{3}{c|}{Rotation} & \multicolumn{3}{c|}{Warping} & \multicolumn{3}{c|}{Keystoning} & \multicolumn{3}{c|}{Watermark} & \multicolumn{3}{c|}{Background} & \multicolumn{3}{c|}{Illumination} & \multicolumn{3}{c|}{Ink-Bleeding} & \multicolumn{3}{c|}{Ink-Holdout} & \multicolumn{3}{c|}{Defocus} & \multicolumn{3}{c|}{Vibration} & \multicolumn{3}{c|}{Speckle} & \multicolumn{3}{c}{Texture} \\
& & V & L & T & & L1 & L2 & L3 & L1 & L2 & L3 & L1 & L2 & L3 & L1 & L2 & L3 & L1 & L2 & L3 & L1 & L2 & L3 & L1 & L2 & L3 & L1 & L2 & L3 & L1 & L2 & L3 & L1 & L2 & L3 & L1 & L2 & L3 & L1 & L2 & L3 \\
\midrule \midrule
ResNet~\cite{kaiming2015resnet}&Faster R-CNN
~\cite{ren2017FasterRcnn} & \cmark & \xmark & \xmark & \xmark & 119.5 & 110.1 & 118.8 & 156.6 & 150.7 & 100.6 & 074.9 & 077.1 & 102.2 & 287.1 & 257.2 & 241.2 & 184.7 & 175.6 & 171.5 & 238.5 & 223.8 & 220.8 & 293.1 & 248.8 & 210.9 & 286.9 & 199.2 & 131.3 & 299.6 & 271.0 & 260.1 & 281.5 & 264.9 & 239.4 & 237.7 & 207.6 & 195.5 & 190.5 & 173.7&169.6

\\
ResNet~\cite{kaiming2015resnet}&DocSegTr~\cite{biswas2022docsegtr}& \cmark & \xmark & \xmark & \xmark & 132.9 & 111.5 & 117.8 & 187.9 & 226.9 & 161.6 & 088.9 & 081.3 & 112.0 & 355.1 & 423.1 & 391.9 & 280.0 & 236.9 & 270.6 & 160.1 & 171.6 & 166.8 & 091.3 & 084.0 & 249.7 & 431.3 & 280.2 & 172.4 & 085.3 & 310.5 & 344.5 & 292.0 & 284.9 & 243.4 & 291.5 & 212.9 & 195.9 & 164.5 & 135.5 & 138.0

\\
ResNet~\cite{kaiming2015resnet}&Mask R-CNN 
~\cite{He_2017_MaskRcnn} & \cmark & \xmark & \xmark & \xmark &  148.1 & 119.7 & 048.5 & 155.1 & 170.0 & 093.9 & 070.2 & 066.7 & 075.3 & 298.0 & 298.0 & 359.6 & 270.2 & 246.6 & 137.1 & 259.5 & 347.5 & 245.1 & 389.0 & 305.3 & 189.5 & 596.1 & 454.4 & 125.0 & 381.5 & 320.0 & 220.1 & 525.4 & 310.2 & 100.2 & 200.5 & 126.6 & 107.6 & 121.5 & 102.0&098.1

\\
Swin~\cite{liu2021swin}&SwinDocSegmenter~\cite{banerjee2023swindocsegmenter}&\cmark & \xmark & \xmark & \xmark  & 148.1 & 121.9 & 116.7 & 214.0 & 208.8 & 142.7 & 098.5 & 097.7 & 129.5 & 398.0 & 346.7 & 321.4 & 239.8 & 215.2 & 224.0 & 324.3 & 295.2 & 288.9 & 411.4 & 335.1 & 248.4 & 399.9 & 271.0 & 161.8 & 417.5 & 344.2 & 236.3 & 374.5 & 220.6 & 208.8 & 294.4 & 200.5 & 168.4 & 195.8 & 163.1&152.0

\\
\rowfont{\color{gray!30}} DiT~\cite{li2022dit}&Cascade R-CNN~\cite{cai2018cascade} & \cmark & \xmark & \xmark & \cmark &084.2 & 065.8 & 062.1 & 152.5 & 142.6 & 102.8 & 057.4 & 058.5 & 079.6 & 298.8 & 254.3 & 281.0 & 198.5 & 175.3 & 180.7 & 173.8 & 155.7 & 157.7 & 293.6 & 242.7 & 184.1 & 290.0 & 226.7 & 170.7 & 278.6 & 235.3 & 163.1 & 323.1 & 217.0 & 230.7 & 175.1 & 103.3 & 085.1 & 096.7 & 078.4&075.1

\\
\rowfont{\color{gray!30}} LayoutLMv3~\cite{huang2022layoutlmv3}&Cascade R-CNN~\cite{cai2018cascade} & \cmark & \cmark & \cmark & \cmark & 084.9 & 068.2 & 067.9 & 138.9 & 134.7 & 108.8 & 063.8 & 060.3 & 078.7 & 267.5 & 234.7 & 225.1 & 176.2 & 172.9 & 184.1 & 206.7 & 199.3 & 225.4 & 372.9 & 313.6 & 222.4 & 275.2 & 255.0 & 199.6 & 323.0 & 284.7 & 190.8 & 381.8 & 258.1 & 246.7 & 178.9 & 113.6 & 090.9 & 101.5 & 090.4 & 096.3

\\
\midrule
\rowcolor[gray]{.9} InternImage~\cite{wang2023internimage}&RoDLA (Ours) & \cmark & \xmark & \xmark & \xmark & 089.8 & 086.1 & 096.6 & 124.8 & 123.0 & 085.8 & 058.2 & 057.0 & 078.1 & 234.1 & 201.7 & 192.7 & 151.1 & 139.8 & 148.8 & 190.6 & 170.0 & 163.8 & 236.6 & 203.3 & 149.8 & 234.7 & 149.9 & 089.9 & 243.4 & 196.1 & 143.5 & 219.7 & 152.8 & 172.6 & 184.3 & 141.8 & 126.9 & 142.7 & 122.5 & 115.1

\\

\bottomrule[1.5pt]

\end{tabu}
}
\vskip -1ex
\end{table*}

\section{Comparison with image rectification}
To fully demonstrate the significance of researching the robustness of DLA models, we will compare the direct prediction results using our RoDLA model with the results obtained by first rectifying the document images using a Document Image Rectification model, followed by applying the DLA model. For the Document Image Rectification task, we employ the state-of-the-art DocTr~\cite{feng2023doctr} model. However, it is important to note that, currently, the task of document image rectification only considers the recovery of four types of perturbations in our robustness benchmark: \textbf{Rotation}, \textbf{Warping}, \textbf{Keystoning}, and \textbf{Illumination}. Therefore, the comparison will be located in these four perturbations.

\subsection{Pipeline Comparison}
\begin{figure*}[t]
 \footnotesize	
 \centering
 \subfloat[Our RoDLA inference pipeline under perturbations]{\includegraphics[width=0.75\textwidth]{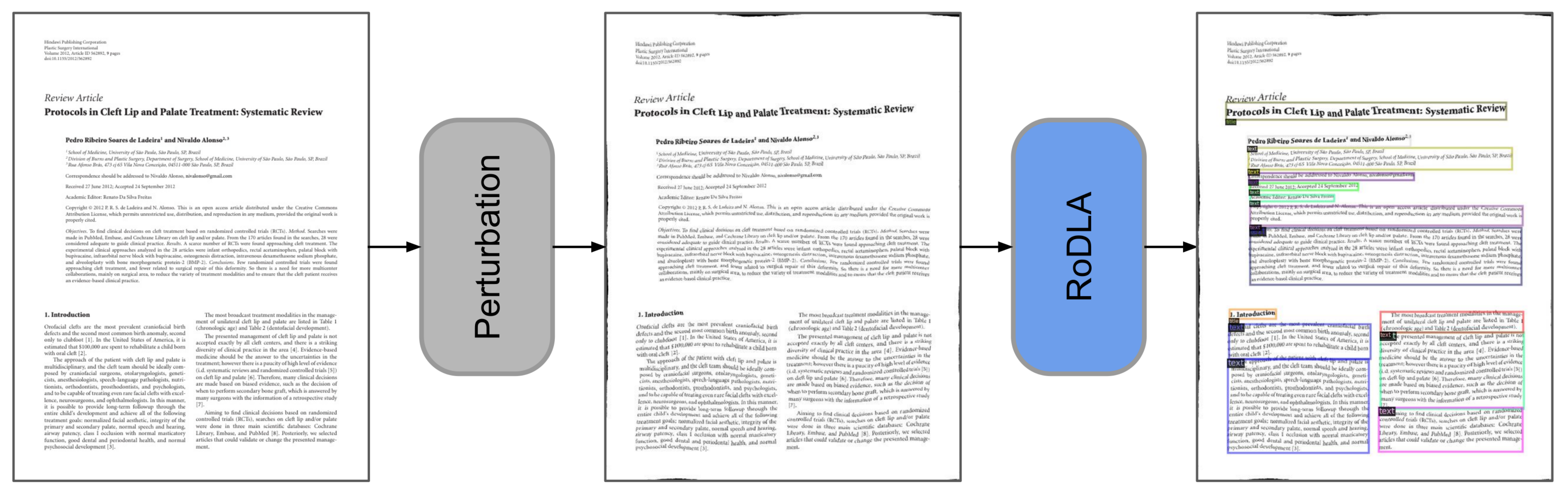}\label{fig:comparison-a}}
 \hspace{30pt}
  \subfloat[The ground truth]{\frame{\includegraphics[width=0.168\textwidth]{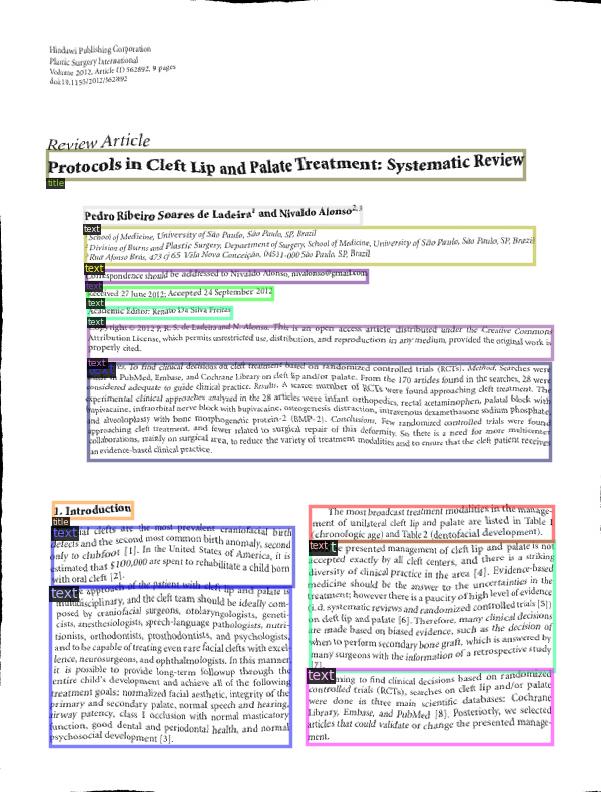}  \label{fig:comparison-c}}\vspace{4pt}}
  \\ \vspace{18pt}
 \subfloat[Two-stage inference pipeline under perturbations, including a document image rectification model (DocTr~\cite{feng2023doctr}) and a DLA model (Faster R-CNN~\cite{ren2017FasterRcnn}).]{\includegraphics[width=\textwidth]{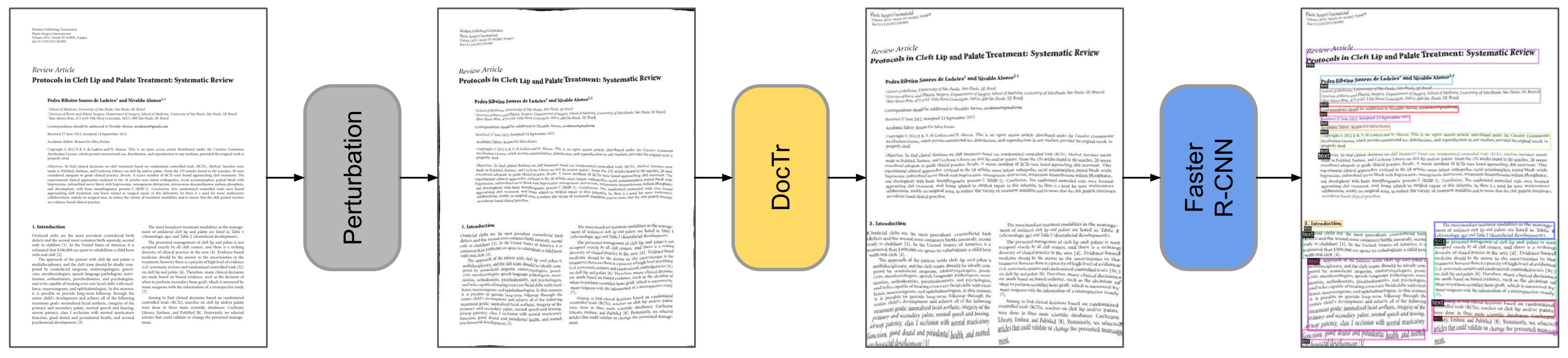}\label{fig:comparison-b}}
 \caption{\textbf{Pipeline comparison} between (a) our one-stage inference (RoDLA) and (c) two-stage inference (first rectification, then DLA). Compared to (b) the ground truth, our RoDLA can obtain better DLA results.} 
\label{fig:pipeline_comparison}
\end{figure*}

Fig.~\ref{fig:pipeline_comparison} visualizes the two different pipelines, with Fig.~\ref{fig:comparison-a} showing the inference flow of our RoDLA model and Fig.~\ref{fig:comparison-b} depicting the combined use of DocTr~\cite{feng2023doctr} for image rectification followed by Faster R-CNN~\cite{ren2017FasterRcnn} inference. Notably, while DocTr~\cite{feng2023doctr} quickly rectifies spatial perturbations (\textbf{0.45s} per image), it is slow in recovering documents with illumination disturbances, typically requiring \textbf{8.2s} to rectify a single image. Furthermore, as evidenced in Fig.~\ref{fig:pipeline_comparison}, it is apparent that for spatial perturbations, the DocTr~\cite{feng2023doctr} fails to adequately restore the original document images. Instead, it introduces additional spatial distortions, leading to erroneous predictions by Faster R-CNN~\cite{ren2017FasterRcnn}.

\subsection{Result Comparison}
To elucidate the distinctions between these two inference pipelines in greater detail, Table~\ref{tab:result-comparison} showcases a comparative analysis of the inference outcomes for our RoDLA model and Faster R-CNN~\cite{ren2017FasterRcnn} on three perturbation datasets. Specifically, it contrasts our model's performance against the combined DocTr~\cite{feng2023doctr} and Faster R-CNN~\cite{ren2017FasterRcnn} approach for four types of perturbations: \textbf{Rotation}, \textbf{Warping}, \textbf{Keystoning}, and \textbf{Illumination}. From Table~\ref{tab:result-comparison}, DocTr~\cite{feng2023doctr} with Faster R-CNN~\cite{ren2017FasterRcnn} Shows a significant drop in performance across all perturbations at the PubLayNet-P, DocLayNet-P, and M$^6$Doc datasets. This suggests DocTr~\cite{feng2023doctr}, while rectifying images, may not be effectively handling these perturbations. Considering the time required for document image rectification, our model RoDLA not only demonstrates exceptional robustness but also exhibits high efficiency.

\begin{table*}
\centering
\caption{Result comparison between two-stage and one-stage pipelines. The \textbf{P-Avg$\uparrow$} are evaluated on\textbf{ PubLayNet-P}, \textbf{DocLayNet-P}, and \textbf{M$^6$Doc-P} datasets. `-' means no document image rectification model has been implemented. Here, \textbf{P-Avg} only refers to the result for four types of perturbations: Rotation, Warping, Keystoning, and Illumination.}
\vskip -2ex
\label{tab:result-comparison}
\setlength{\tabcolsep}{6pt}
\resizebox{\linewidth}{!}{
\begin{tabu}{l|l|c|ccc|ccc|ccc|ccc|c}
\toprule[1.5pt]
\multirow{2}{*}{\textbf{Rectification Model}}&\multirow{2}{*}{\textbf{DLA Model}} & \multirow{2}{*}{\textbf{Clean}} & \multicolumn{3}{c|}{Rotation} & \multicolumn{3}{c|}{Warping} & \multicolumn{3}{c|}{Keystoning} & \multicolumn{3}{c|}{Illumination} &\multirow{2}{*}{\textbf{P-Avg$\uparrow$}}\\
& && L1 & L2 & L3 & L1 & L2 & L3 & L1 & L2 & L3 & L1 & L2 & L3  \\
\midrule[1pt] \midrule[1pt]
\multicolumn{16}{c}{PubLayNet-P}\\
\midrule[1pt]
 DocTr~\cite{feng2023doctr} &Faster R-CNN~\cite{ren2017FasterRcnn} &90.2& 37.5&50.6& 49.6&18.9&19.8&20.0&20.6&27.1&31.4&74.1&72.7&72.1&41.2\\
 -&Faster R-CNN~\cite{ren2017FasterRcnn} &90.2& 67.9&44.1&20.6&79.7&75.2&71.0&80.0&74.1&68.8&81.9&81.3&81.1&68.8\\
\rowcolor[gray]{.9} - &RoDLA (Ours)& 96.0 &71.9&19.9&02.9&89.0&80.4&68.5&88.4&81.2&72.1&92.0&91.4&91.5&\textbf{70.8}
\\
\midrule[1pt]
\multicolumn{16}{c}{DocLayNet-P}\\
\midrule[1pt]
 DocTr~\cite{feng2023doctr} &Faster R-CNN~\cite{ren2017FasterRcnn} & 73.4&10.9&21.0&20.9&04.5 & 04.2&04.0 &04.8&06.5&07.8&62.1&61.4&60.8&22.4\\
 -&Faster R-CNN~\cite{ren2017FasterRcnn} &73.4& 37.6 & 11.0 & 01.4 & 65.2 & 60.4 & 58.8 & 62.1 & 52.7 & 46.3 &70.6 & 70.5 & 70.1 & 50.6\\
\rowcolor[gray]{.9} - &RoDLA (Ours)&80.5 & 49.6 & 17.8 & 04.3 & 72.6 & 64.2 & 59.4 & 73.2 & 65.8 & 59.1 & 80.3 & 80.2 & 80.0 & \textbf{58.9}
\\
\midrule[1pt]
\multicolumn{16}{c}{M$^6$Doc-P}\\
\midrule[1pt]
 DocTr~\cite{feng2023doctr} &Faster R-CNN~\cite{ren2017FasterRcnn} &62.0&12.8&25.1&24.2&06.2&07.0&07.0&07.0&08.7&09.1&53.7&52.6&52.0&22.1\\
  -&Faster R-CNN~\cite{ren2017FasterRcnn} &62.0& 44.6 & 24.4 & 06.2 & 60.1 & 58.4 & 57.8 & 56.6 & 50.5 & 48.8 &59.3 & 58.1 & 56.6 &48.5\\
\rowcolor[gray]{.9} - &RoDLA (Ours)&70.0& 58.4 & 40.9 & 23.7 & 68.2 & 66.0 & 64.0 & 66.3 & 63.4 & 60.9 & 67.5 & 68.2 & 67.8 &\textbf{59.6}
\\
\bottomrule[1.5pt]

\end{tabu}
}
\end{table*}

\section{Visualization Results}
\begin{figure*}[!ht]
 \scriptsize
 \centering
 \subfloat{\frame{\includegraphics[width=0.38\columnwidth]{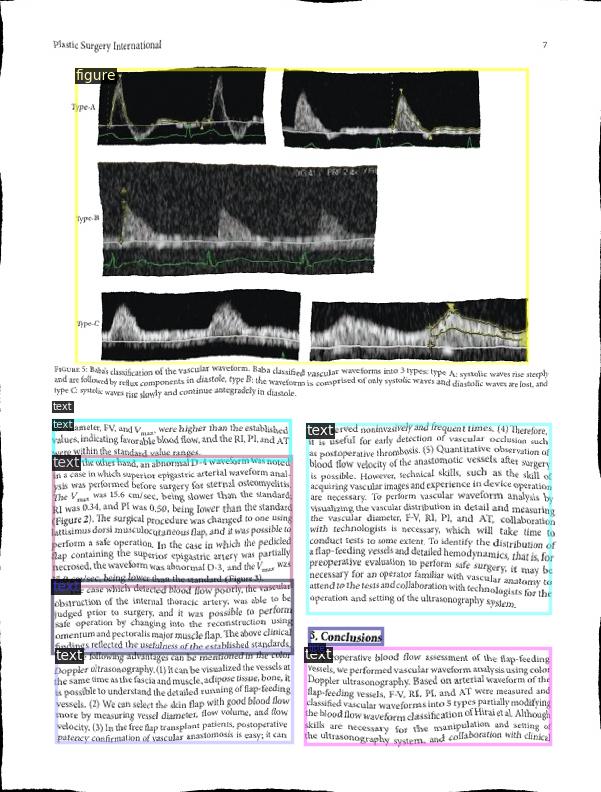}}}\hspace{5pt}
 \subfloat{\frame{\includegraphics[width=0.38\columnwidth]{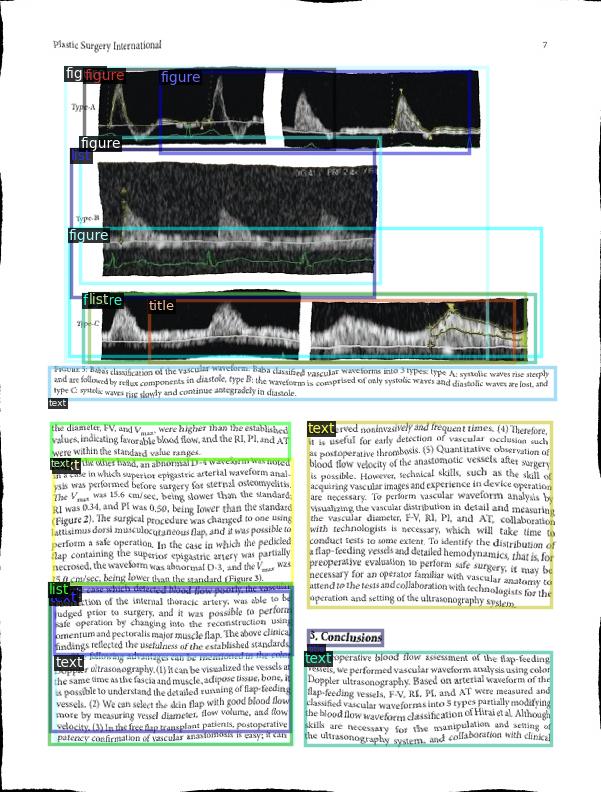}}}\hspace{5pt}
 \subfloat{\frame{\includegraphics[width=0.38\columnwidth]{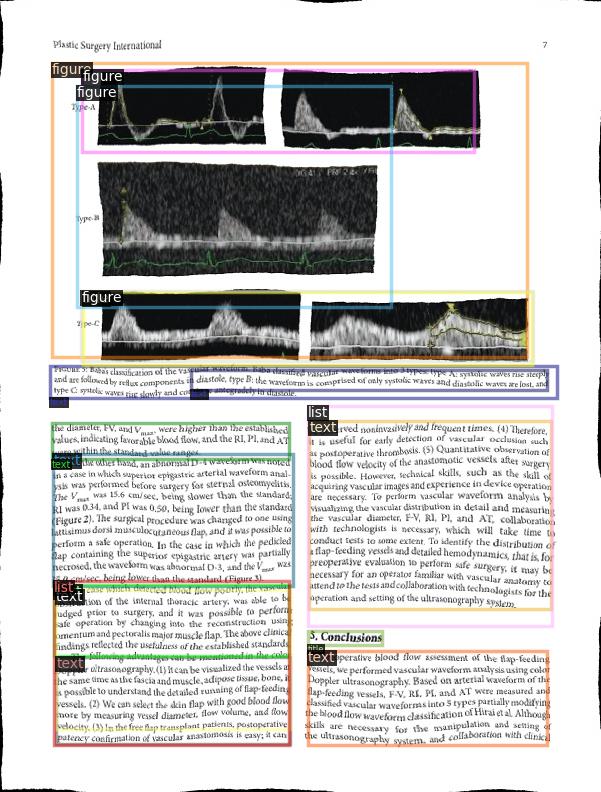}}}\hspace{5pt}
 \subfloat{\frame{\includegraphics[width=0.38\columnwidth]{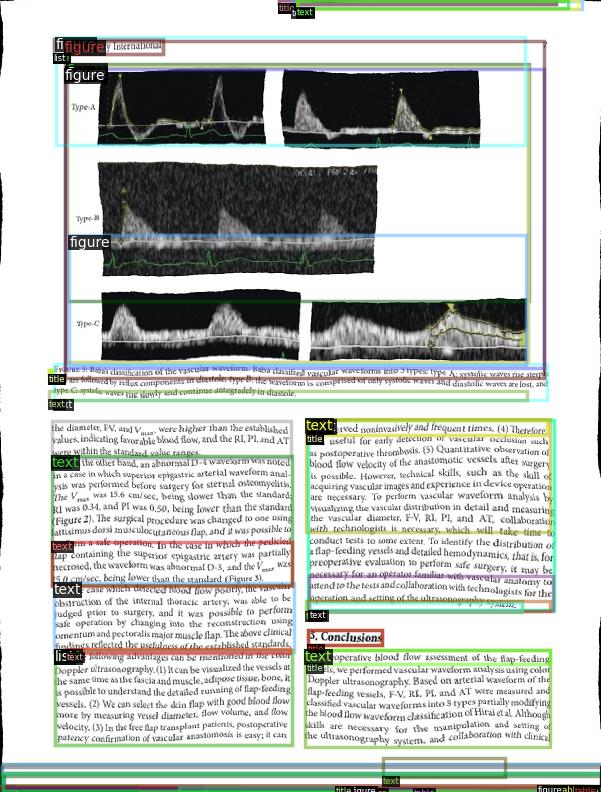}}}\hspace{5pt}
 \subfloat{\frame{\includegraphics[width=0.38\columnwidth]{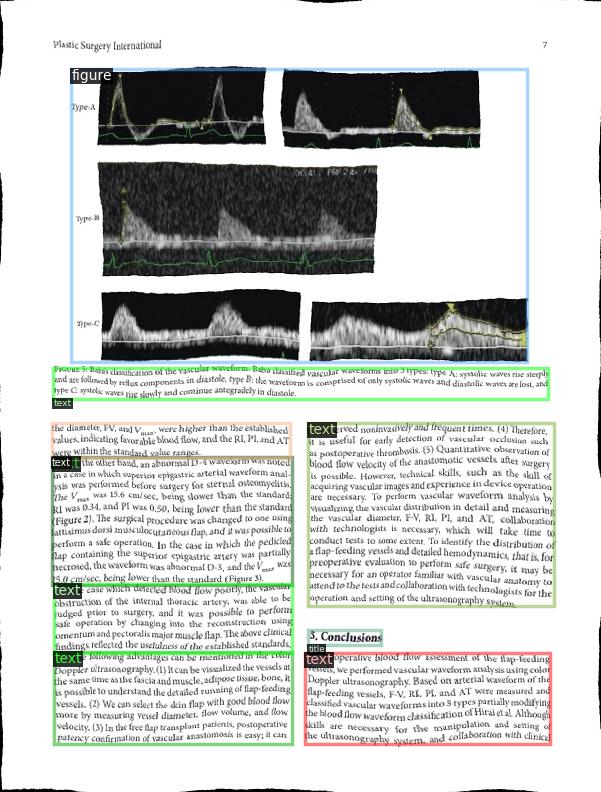}}}\hspace{5pt}
 \\ \vspace{5pt}
  \subfloat{\frame{\includegraphics[width=0.38\columnwidth]{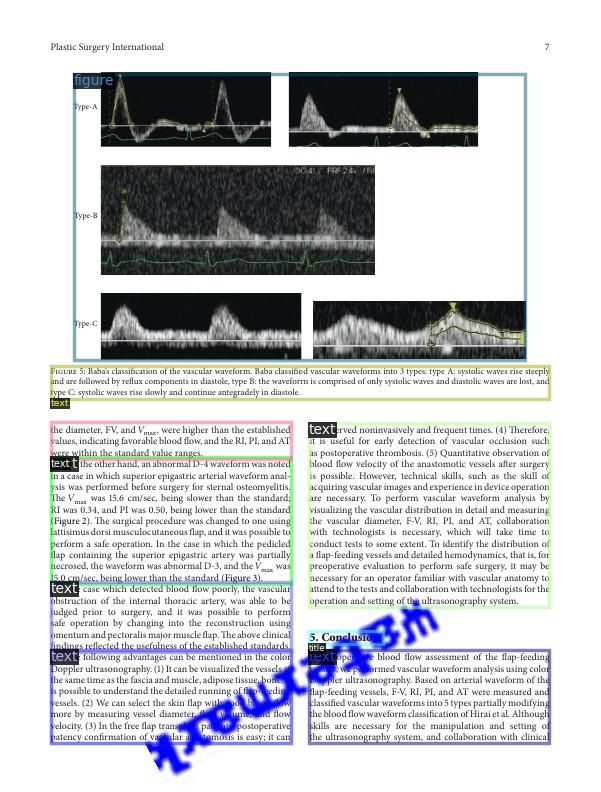}}}\hspace{5pt}
 \subfloat{\frame{\includegraphics[width=0.38\columnwidth]{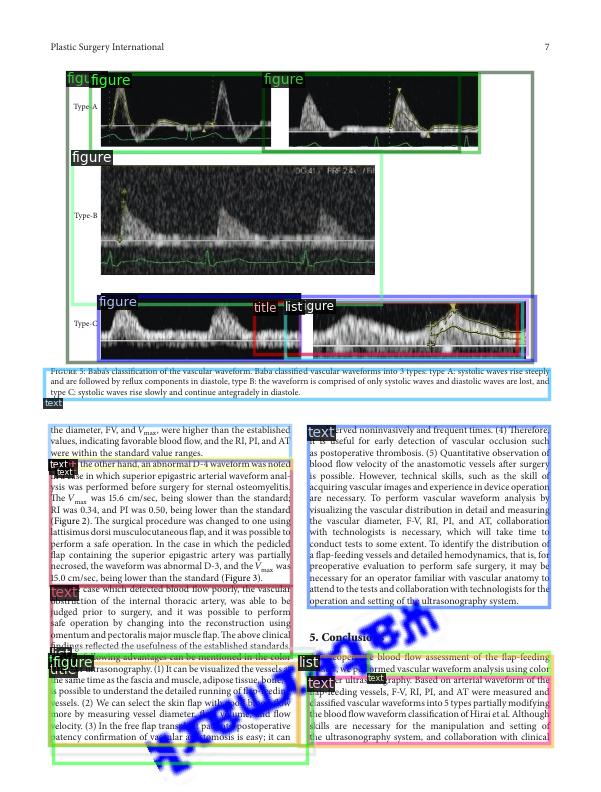}}}\hspace{5pt}
 \subfloat{\frame{\includegraphics[width=0.38\columnwidth]{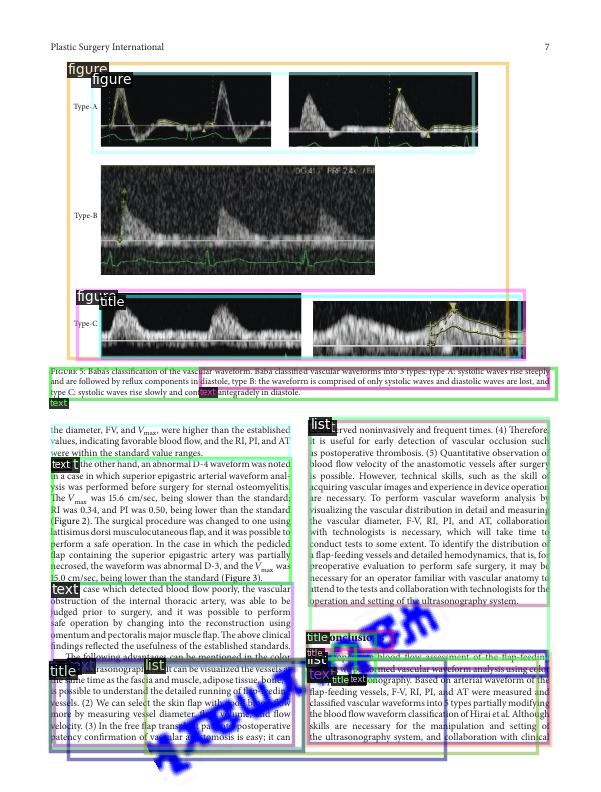}}}\hspace{5pt}
 \subfloat{\frame{\includegraphics[width=0.38\columnwidth]{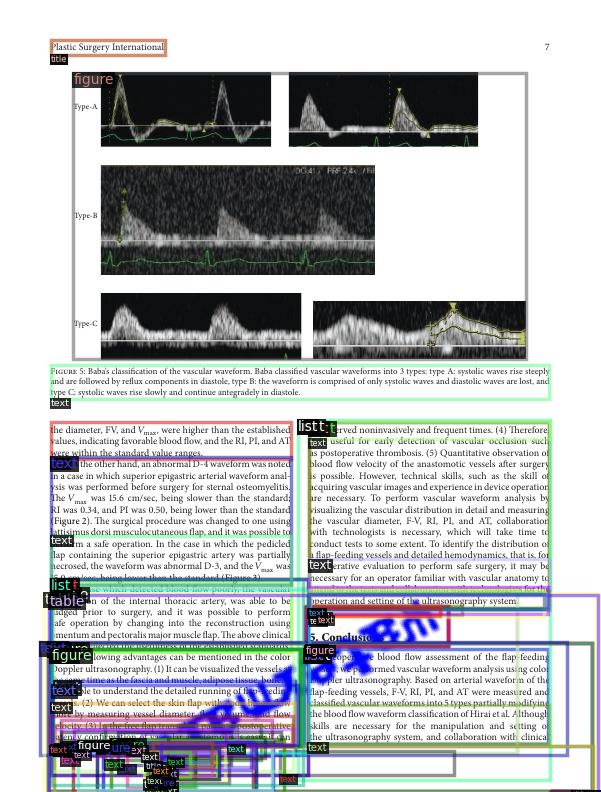}}}\hspace{5pt}
 \subfloat{\frame{\includegraphics[width=0.38\columnwidth]{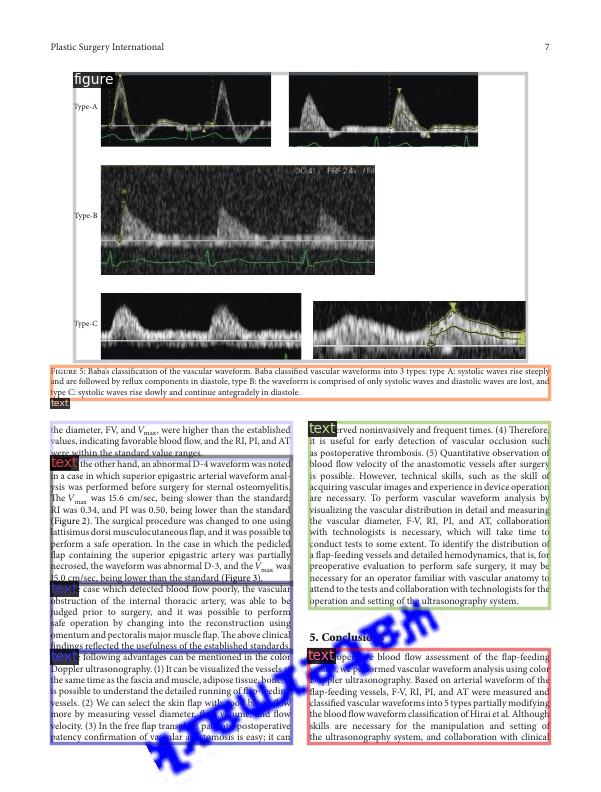}}}\hspace{5pt}
 \\ \vspace{5pt}
  \subfloat{\frame{\includegraphics[width=0.38\columnwidth]{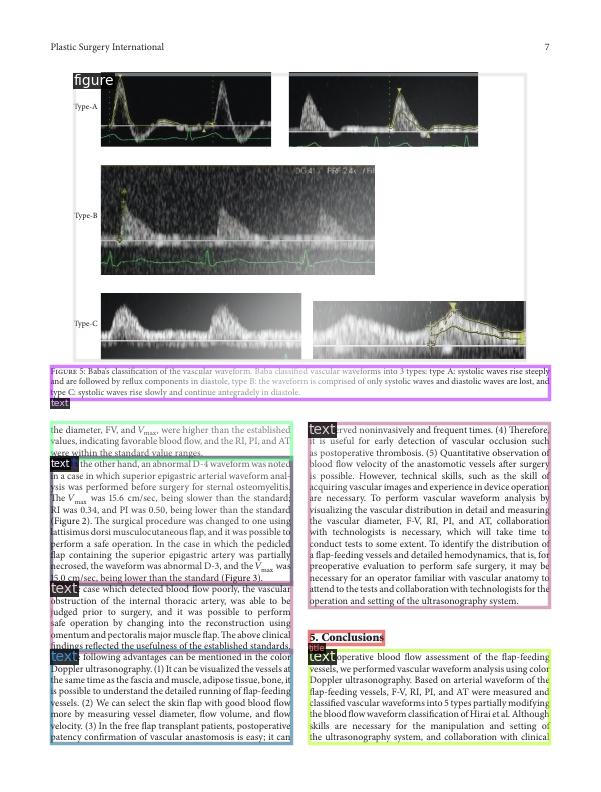}}}\hspace{5pt}
 \subfloat{\frame{\includegraphics[width=0.38\columnwidth]{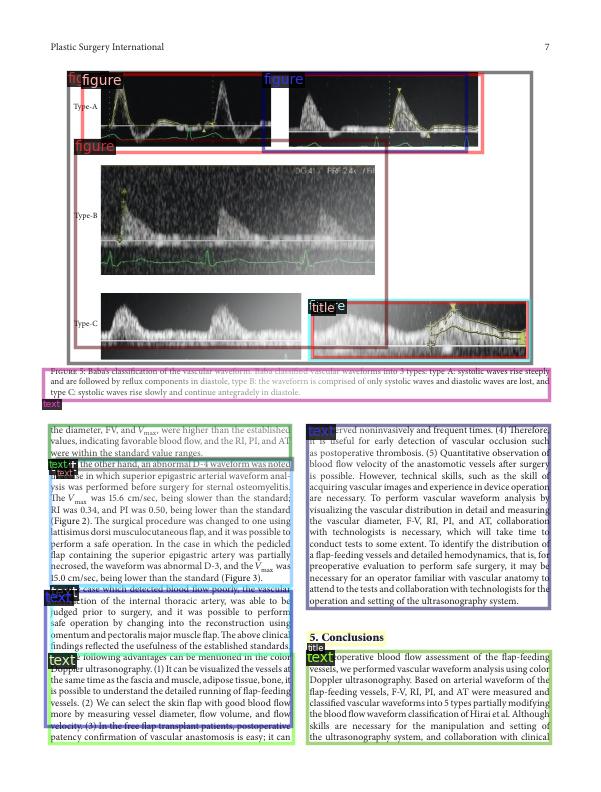}}}\hspace{5pt}
 \subfloat{\frame{\includegraphics[width=0.38\columnwidth]{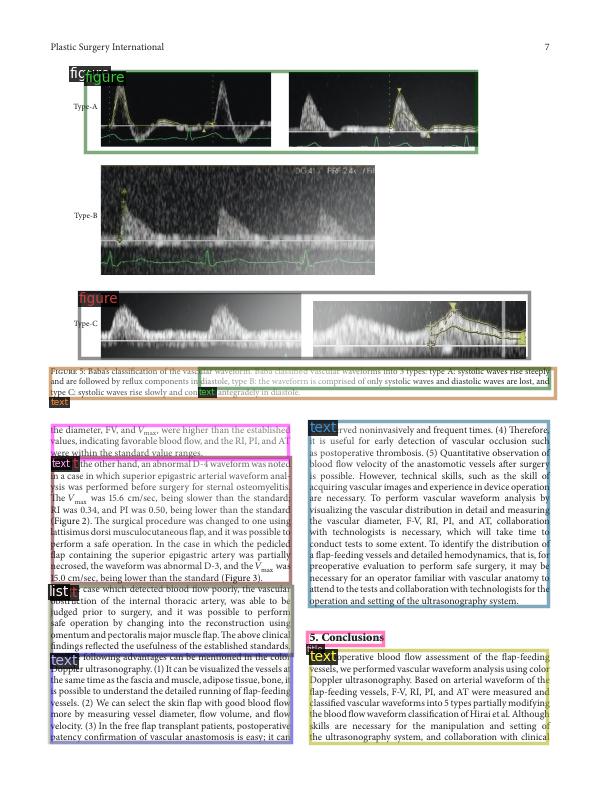}}}\hspace{5pt}
 \subfloat{\frame{\includegraphics[width=0.38\columnwidth]{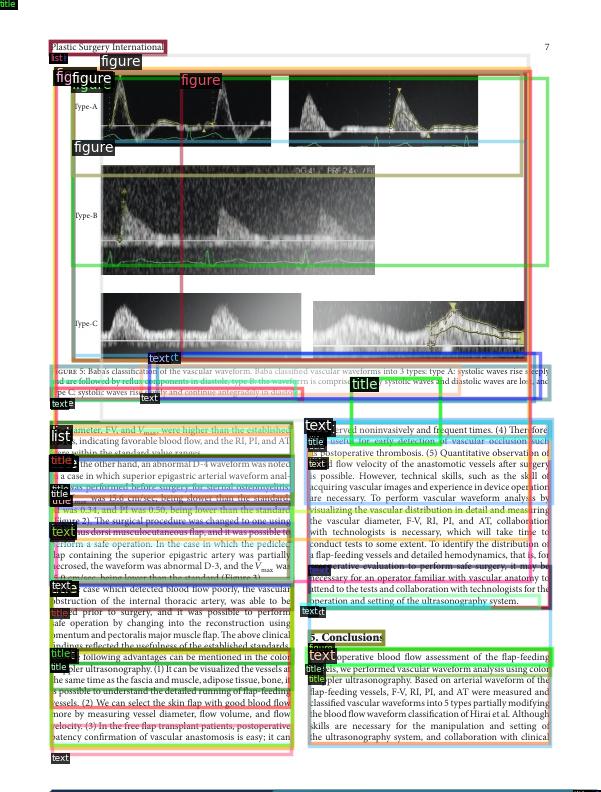}}}\hspace{5pt}
 \subfloat{\frame{\includegraphics[width=0.38\columnwidth]{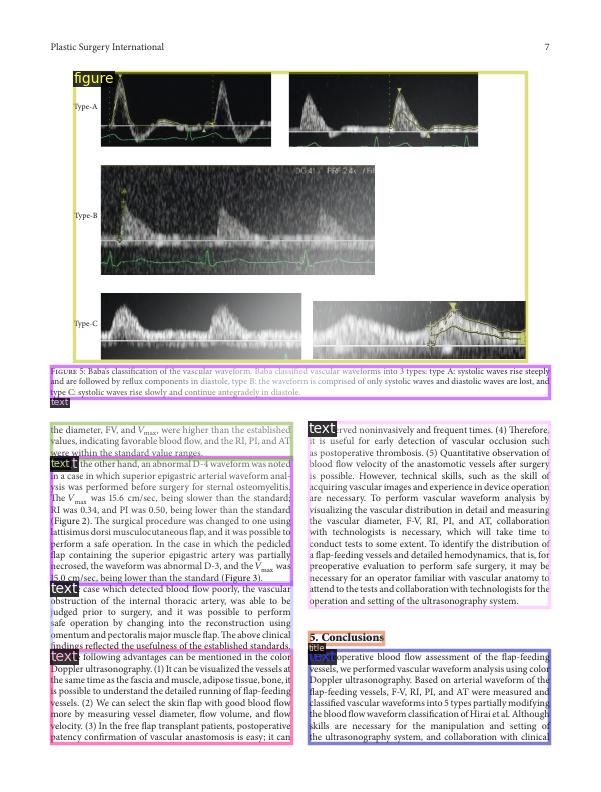}}}\hspace{5pt}
 \\ \vspace{5pt}
  \subfloat{\frame{\includegraphics[width=0.38\columnwidth]{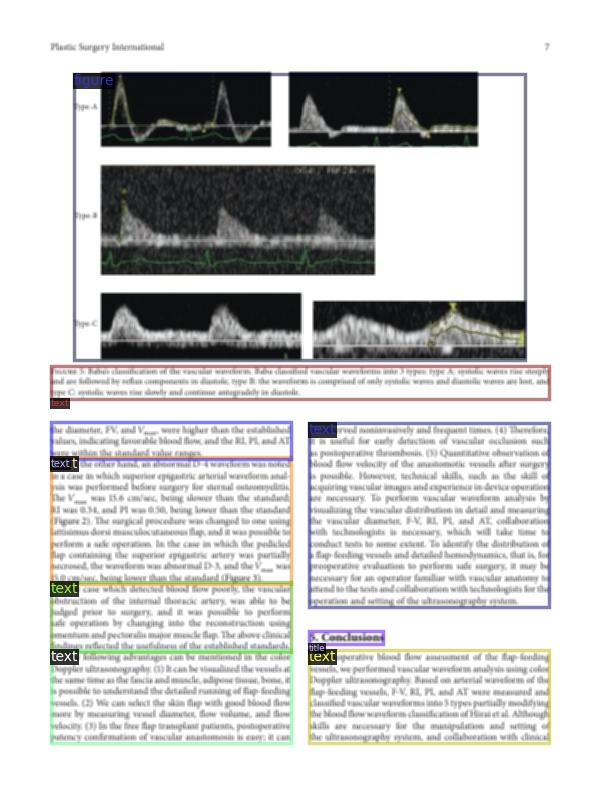}}}\hspace{5pt}
 \subfloat{\frame{\includegraphics[width=0.38\columnwidth]{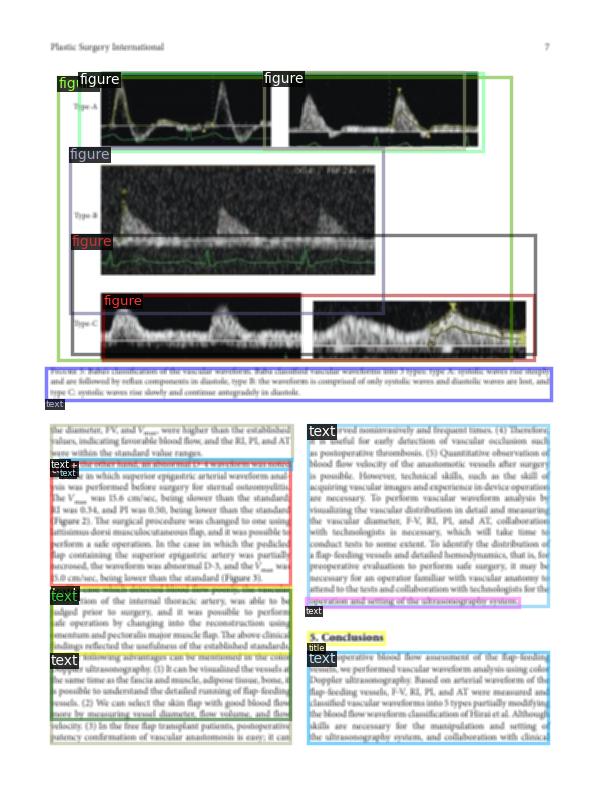}}}\hspace{5pt}
 \subfloat{\frame{\includegraphics[width=0.38\columnwidth]{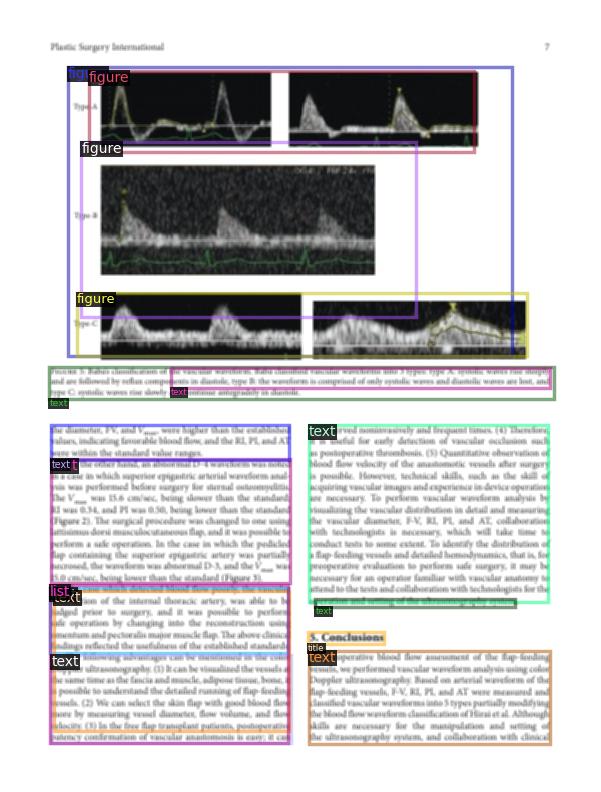}}}\hspace{5pt}
 \subfloat{\frame{\includegraphics[width=0.38\columnwidth]{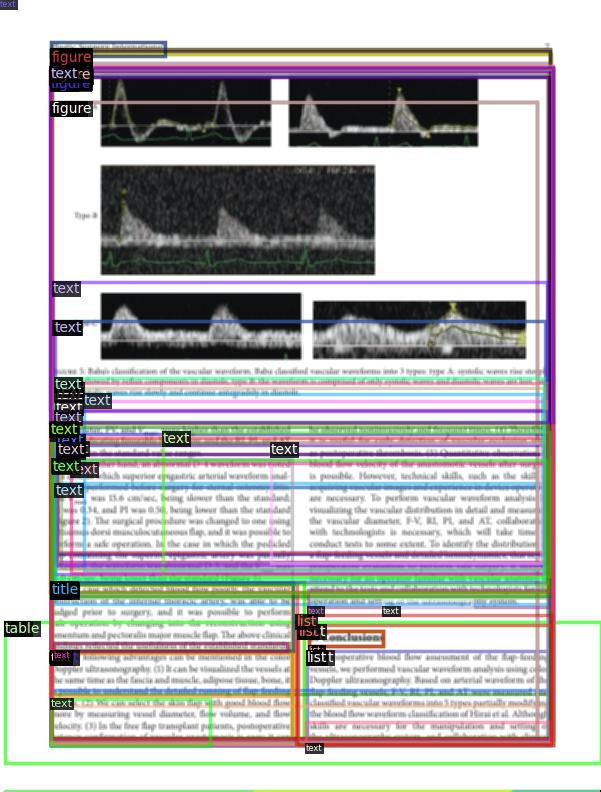}}}\hspace{5pt}
 \subfloat{\frame{\includegraphics[width=0.38\columnwidth]{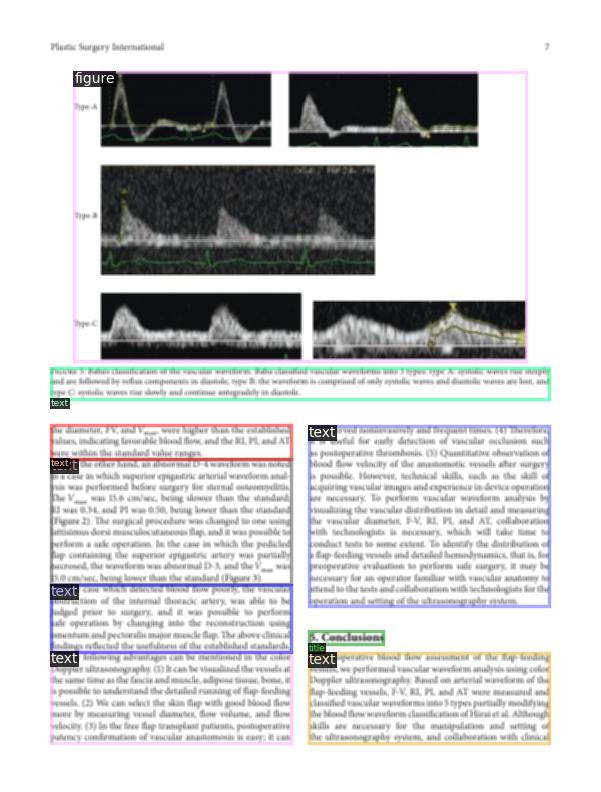}}}\hspace{5pt}
 \\ \vspace{5pt}
   \subfloat{\frame{\includegraphics[width=0.38\columnwidth]{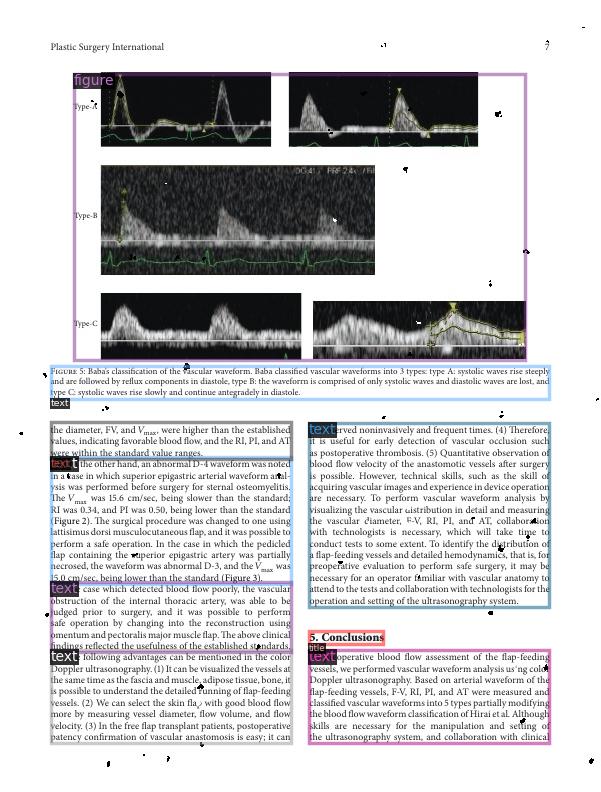}}}\hspace{5pt}
 \subfloat{\frame{\includegraphics[width=0.38\columnwidth]{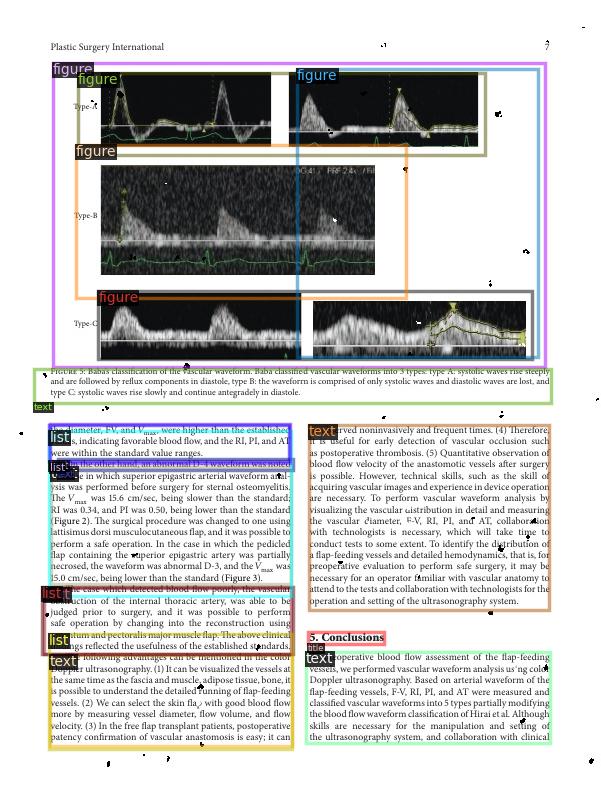}}}\hspace{5pt}
 \subfloat{\frame{\includegraphics[width=0.38\columnwidth]{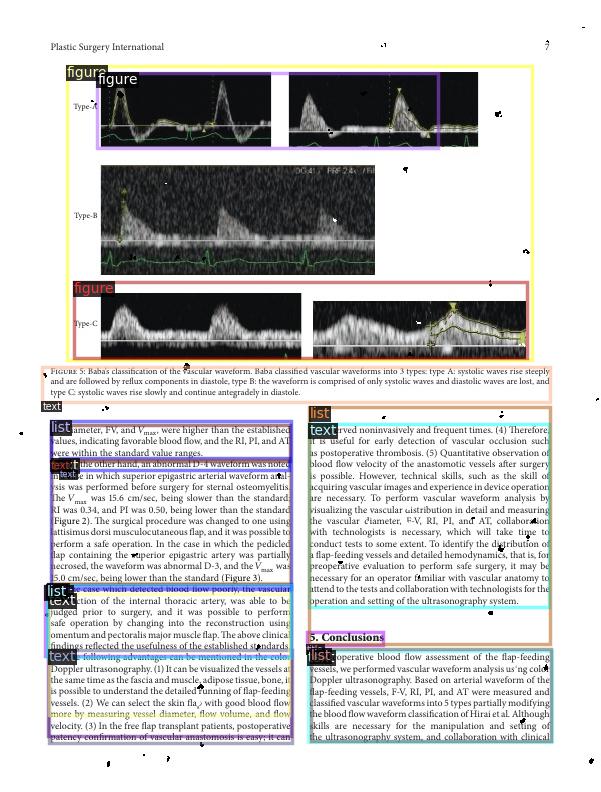}}}\hspace{5pt}
 \subfloat{\frame{\includegraphics[width=0.38\columnwidth]{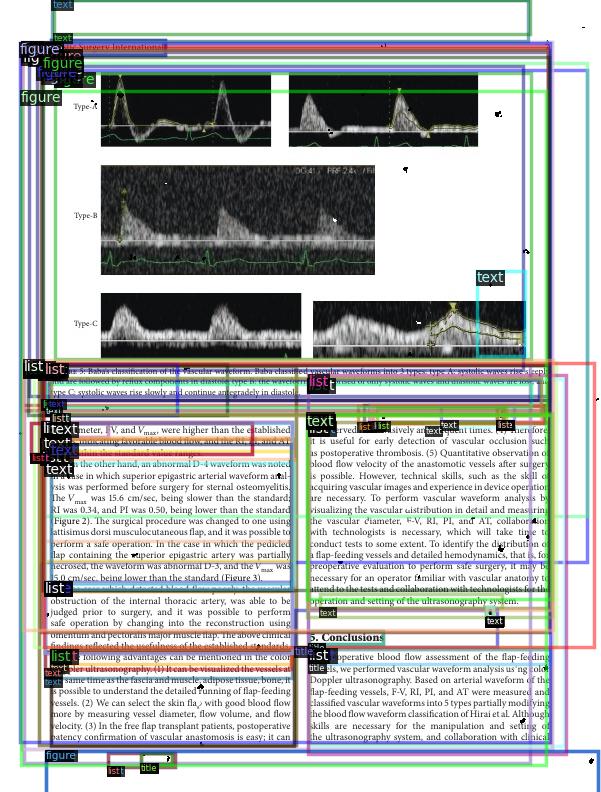}}}\hspace{5pt}
 \subfloat{\frame{\includegraphics[width=0.38\columnwidth]{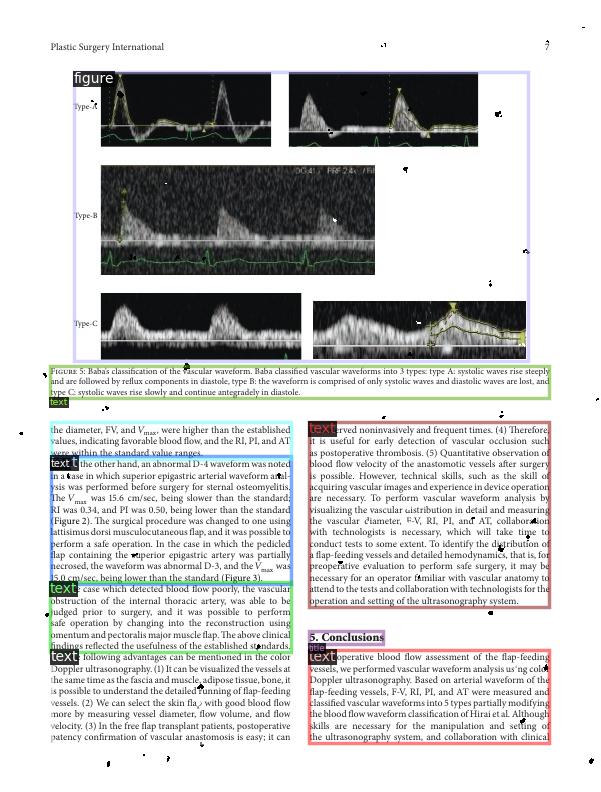}}}\hspace{5pt}
 \\ 
 \vskip -2ex 
 \caption{Visualizations on PubLayNet-P. From top to bottom: \textbf{Warping}, \textbf{Watermark}, \textbf{Illumination}, \textbf{Defocus}, and \textbf{Speckle}. From left to right: \textbf{the ground truth}, predictions from, \ie, \textbf{Faster R-CNN}~\cite{ren2017FasterRcnn}, \textbf{Mask R-CNN}~\cite{He_2017_MaskRcnn}, \textbf{SwinDocSegmenter}~\cite{banerjee2023swindocsegmenter}, and our \textbf{RoDLA}. 
 }
\label{fig:visual_publaynet}
\end{figure*}
Beyond providing detailed performance data, to facilitate a more intuitive understanding, this section includes visualizations of selected model predictions on our robustness benchmark (\eg, the PubLayNet-P dataset) for one document image perturbation in each of the following categories: Warping from Spatial, Watermark from Content, Illumination from Inconsistency, Defocus from Blur, and Speckle from Noise. We further compare the visualization of predictions from Faster R-CNN~\cite{ren2017FasterRcnn}, Mask R-CNN~\cite{He_2017_MaskRcnn}, and SwinDocSegmenter~\cite{banerjee2023swindocsegmenter}. Observing the contrasts in Fig.~\ref{fig:visual_publaynet}, it becomes evident that our RoDLA model exhibits strong robustness against various perturbations, consistently yielding better predictions that align closely with the ground truth. In contrast, the other three models, \ie, Faster R-CNN~\cite{ren2017FasterRcnn}, Mask R-CNN~\cite{He_2017_MaskRcnn}, and SwinDocSegmenter~\cite{banerjee2023swindocsegmenter}, demonstrate varying degrees of erroneous predictions when confronted with different perturbations. Notably, SwinDocSegmenter displays particularly pronounced inaccuracies under the influence of these perturbations. The qualitative analysis proves the effectiveness of our proposed RoDLA in enhancing the robustness of DLA. 

\section{Discussion}
\subsection{Limitations and Future Works} 
Our benchmark, designed for testing the robustness of Document Layout Analysis (DLA) models, currently simulates only a subset of perturbations commonly encountered in document images. It does not account for content tampering or content replacement, which could significantly impact model robustness. Additionally, our robustness benchmark is limited to only three severity levels. This granularity might be too coarse, and further subdivision into more nuanced levels could reveal subtler variations in DLA model robustness across different intensities of perturbations. In addition to our current robustness benchmark, we have only separately identified potential perturbations and assessed their impacts individually. We have not yet combined multiple perturbations on a single image with a certain probability, which would more accurately reflect real-world scenarios. A comprehensive evaluation of these combined perturbations is necessary for a more realistic assessment of the robustness of DLA. Moreover, in the robustness evaluation metric, we have incorporated only two image quality assessment methods and have used the performance of the Faster R-CNN~\cite{ren2017FasterRcnn} as the reference for calculating the Mean Perturbation Effect (mPE). To enhance the objectivity of mPE and more accurately reflect the impact of perturbations on document images, it would be beneficial to include a broader range of image quality assessment methods and performances from various models.

\subsection{Societal Impacts}  Our robustness benchmark and RoDLA model for document layout analysis bear significant implications. While demonstrating a strong ability to withstand various perturbations and achieving impressive robustness metrics, the current evaluation of our model is confined to three document layout datasets. This step is critical to ensuring its applicability in diverse practical scenarios, such as digitizing historical documents, streamlining administrative processes, or enhancing accessibility for visually impaired individuals. As we transition from controlled datasets to varied real-life environments, continuous model refinement is necessary to address any unforeseen challenges, ensuring the model's relevance and effectiveness. Ethical considerations, particularly data integrity and impartiality, are paramount in avoiding biases and incorrect predictions that could have significant societal consequences. While our model demonstrates technical robustness, its deployment in real-world settings requires careful consideration of its broader societal implications to ensure it contributes positively and responsibly.

\section{Acknowledgments}
This work was supported in part by Helmholtz Association of German Research Centers, in part by the Ministry of Science, Research and the Arts of Baden-Württemberg (MWK) through the Cooperative Graduate School Accessibility through AI-based Assistive Technology (KATE) under Grant BW6-03, and in part by BMBF through a fellowship within the IFI programme of DAAD. This work was partially performed on the HoreKa supercomputer funded by the MWK and by the Federal Ministry of Education and Research, partially on the HAICORE@KIT partition supported by the Helmholtz Association Initiative and Networking Fund, and partially on bwForCluster Helix supported by the state of Baden-Württemberg through bwHPC and the German Research Foundation (DFG) through grant INST 35/1597-1 FUGG.

\end{document}